%% file: main.tex
\documentclass{article}
\usepackage{iclr2027_conference,times}
\iclrfinalcopy
\input{math_commands.tex}

\usepackage{hyperref}
\usepackage{url}
\usepackage{amsmath,amssymb}
\usepackage{booktabs}
\usepackage{multirow}
\usepackage{graphicx}
\usepackage{xcolor}
\usepackage{enumitem}
\usepackage{caption}
\usepackage{subcaption}
\usepackage{wrapfig}
\usepackage{longtable}
\usepackage{pdflscape}
\usepackage{tabularx}
\usepackage{adjustbox}
\usepackage{makecell}
\usepackage{pgfplots}
\pgfplotsset{compat=1.18}
\usepgfplotslibrary{groupplots,fillbetween}
\usetikzlibrary{patterns,arrows.meta}

\newcommand{\think}{\texttt{</think>}}

\definecolor{cFourB}{HTML}{2A78D6}
\definecolor{cOneSevenB}{HTML}{EB6834}
\definecolor{cZeroSixB}{HTML}{1BAF7A}
\definecolor{cTeacher}{HTML}{0B0B0B}
\definecolor{cNeutral}{HTML}{BDBCB6}
\definecolor{cGrid}{HTML}{E6E5E0}
\definecolor{cMuted}{HTML}{52514E}
\pgfplotsset{paperstyle/.style={
  tick label style={font=\scriptsize, color=cMuted}, label style={font=\footnotesize},
  legend style={font=\scriptsize, draw=none, fill=none}, title style={font=\footnotesize},
  grid=major, grid style={cGrid}, axis line style={cMuted!60}, tick style={cMuted!60}}}
\newcommand{\sizedot}[1]{\tikz[baseline=-0.6ex]\fill[#1] (0,0) circle (2.2pt);}

\newcommand{\vsdot}[3]{\tikz[baseline=-0.65ex, x=1.1cm]{\draw[cGrid, line width=.9pt] (0,0) -- (0.65,0);
  \draw[cMuted, line width=.7pt] (#3,-0.9ex) -- (#3,0.9ex); \fill[#1] (#2,0) circle (2.1pt);}}

\title{Solving Without Stopping:\\On-Policy Distillation at Small Scale}

\author{Hongyang (Kevin) Li$^{1}$ \quad Yiming Zhu$^{2}$ \quad Xiao Li$^{2}$ \\
\textbf{Caesar Wu$^{1}$ \quad Said Mammar$^{3}$ \quad Pascal Bouvry$^{1}$}}

\begin{document}

\maketitle

\input{sections/00_abstract}

\input{sections/01_intro}

\input{sections/02_setup}

\input{sections/03_ceiling}

\input{sections/04_stopping}

\input{sections/05_unfinished}

\input{sections/06_related}

\input{sections/07_conclusion}

\input{sections/11_statements}

\bibliography{refs}
\bibliographystyle{iclr2027_conference}

\newpage
\appendix
\input{appendix/A_setup}

\input{appendix/B_scoring}

\input{appendix/C_ceiling}

\input{appendix/D_stopping}

\input{appendix/E_unfinished}

\input{appendix/F_notes}

\end{document}

%% file: math_commands.tex
\usepackage{amsmath,amsfonts,bm}

\def\eqref#1{equation~\ref{#1}}

\def\1{\bm{1}}

\def\vs{{\bm{s}}}

\DeclareMathAlphabet{\mathsfit}{\encodingdefault}{\sfdefault}{m}{sl}
\SetMathAlphabet{\mathsfit}{bold}{\encodingdefault}{\sfdefault}{bx}{n}



%% file: sections/00_abstract.tex
\begin{abstract}
On-policy distillation, where a student learns from a stronger teacher's feedback on its own outputs, is a common way
to pass reasoning to smaller models. We analyze what it transfers at small scale, distilling Qwen3-8B into Qwen3 4B,
1.7B and 0.6B students, in thinking mode (reason at length, then end the reasoning and answer) and, for comparison, in
non-thinking mode (no separate reasoning phase). Long reasoning needs two abilities, solving a problem and knowing when
it is solved, and we find that distillation transfers the first, but in thinking mode not the second. Solving improves at every size, up to two ceilings, which we measure
comprehensively across both modes and all student sizes: a student's single attempt never exceeds what it could already reach in many
attempts before training, and the smaller the student, the further it stays below the teacher. Stopping is where the
modes part. In non-thinking mode every student keeps stopping; in thinking mode students stop ending their reasoning
early in training, and the smaller the student, the less of this ability survives: the teacher signals a stop almost
only where a student already ends its reasoning, so distillation teaches no new stops; it only keeps the student's
existing stops that land on a right answer, and a weak student has few such stops. The smallest students often reach the right value
but do not commit to it: they either rarely mark it or mark it and write past it. Together, these results describe
how small students behave under on-policy distillation, and a diagnostic that separates
answer marking, correctness and stopping.
\end{abstract}

%% file: sections/01_intro.tex
\section{Introduction}\label{sec:intro}

\input{figures/fig_overview}

{\renewcommand{\thefootnote}{}\footnotetext{\hspace{-1.8em}$^{1}$University of Luxembourg \quad $^{2}$Seafill Open-Source Community \quad $^{3}$Universit\'e Paris-Saclay}
\footnotetext{\hspace{-1.8em}Emails: \texttt{\{hongyang.li, caesar.wu, pascal.bouvry\}@uni.lu}, \texttt{xiao.li@seafill.com}, \texttt{z13655249157@gmail.com}, \texttt{said.mammar@univ-evry.fr}}}
On-policy distillation trains a student on its own outputs: the student writes a response, a stronger teacher scores
every token of it, and the student moves toward the teacher's distribution
\citep{ross2011dagger,agarwal2024gkd,gu2024minillm,tml2025opd}. It has become a standard way to give small models the
reasoning of larger ones, including from teachers that think at length before they answer. Small models are where this
matters most: they are the ones deployed where compute is scarce, and their reasoning comes almost entirely from
distillation. Yet what a small student actually takes from a thinking teacher is not well understood. Reasoning at length rests on two abilities, solving the problem and recognizing that it is solved, and prior work has
studied the second as a problem of its own, from overthinking
\citep{chen2024overthinking} to knowing when to stop \citep{zhang2025probing,yang2026deer}. We
ask whether on-policy distillation passes both to a small student. End-point scores alone cannot answer this: a student
that stops too rarely looks like one that cannot solve. We therefore follow solving and stopping together.

We distill Qwen3-8B \citep{qwen3} into Qwen3 4B, 1.7B and 0.6B students, so that teacher and students all sit at small
scale. We compare two modes. In thinking mode the model reasons at length before it answers; non-thinking mode, which
answers directly, serves as a reference. In thinking mode, before distillation, each student first learns the thinking
format by supervised fine-tuning, and we use two sources for that data: short external solutions, or the teacher's own
long solutions. Controls vary the rollout budget, start from the officially post-trained model and repeat runs. We
follow every student both at evaluation and through training, and diagnose solving and stopping together, separating
whether a student answers, whether its answer is right, and whether it stops.

The student acquires the first ability and, the smaller it is, less of the second (Figure~\ref{fig:overview}). First, solving has a ceiling (Section~\ref{sec:ceiling}). A distilled student's single-sample accuracy rises toward the
accuracy it could already reach in many samples before training, but never past it (Figure~\ref{fig:overview}a): one
distilled attempt is worth a few attempts of the starting student. The gain comes from problems the student could
already solve sometimes, it arrives early in training, and the set of problems a student can reach at
all barely grows. The share of the teacher's accuracy that a student reaches falls with its size. When the student
gives an answer, that answer is somewhat more often right after distillation at every size. Second, small students do not learn to stop
(Section~\ref{sec:stopping}). Early in thinking-mode distillation
from the external-SFT start, every student stops closing its reasoning, in the same step in which its responses fill
the length budget. What follows works like a filter. Along a response that runs on, the teacher gives almost no signal
to stop, and distillation keeps the stops that land on a correct answer and removes the others. Before training, the
three students stop at similar rates that do not follow their size; the difference is where they stop. When the 4B stops, its answer is right about half the time; when the 0.6B
stops, almost never. The filter therefore leaves the 4B with some of its stops and the 0.6B with almost none. The small
student has not forgotten how to stop; it was never taught where to stop, and the stops it had were the wrong ones. Third, we look inside the samples that do not stop
(Section~\ref{sec:unfinished}). Before distillation, the 0.6B often
gives up with a wrong answer; after it, it often keeps reasoning until it is cut off at the length limit
and writes no answer at all. Samples with only wrong answers shrink sharply, and samples with no answer grow just as
sharply. They are often not unsolved: the correct value appears
in them early, several times more often than by chance, and the student keeps writing past it. From the teacher's own
solutions, the small students do mark answers, but overwrite correct ones as they keep writing. Either way, they reach
the right value but do not commit to it.

Our contributions are the following. \textbf{A comprehensive analysis} of on-policy distillation at small scale,
across three student sizes, thinking and non-thinking mode, and two sources of fine-tuning data, followed both at
evaluation and through training. \textbf{Two findings about what transfers}: solving rises only up to two ceilings, the
student's own reach and the teacher's, systematically measured here for on-policy distillation in both thinking and
non-thinking mode and across student sizes; and in thinking mode distillation filters stops rather than teaching them,
which leaves the smallest students almost unable to stop. \textbf{A diagnostic} that separates whether a student
answers, whether its answer is right, and whether it stops, so that a change in score can be traced to one of them; the
main comparisons carry paired intervals.

%% file: figures/fig_overview.tex
\begin{figure}[t]
\centering
\begin{tikzpicture}
\node[font=\footnotesize] at (1.40,1.20) {(a) how far solving rises};
\node[font=\scriptsize, text=cMuted] at (0.00,0.78) {4B};
\node[font=\scriptsize, text=cMuted] at (1.40,0.78) {1.7B};
\node[font=\scriptsize, text=cMuted] at (2.80,0.78) {0.6B};
\node[font=\scriptsize, text=cMuted, anchor=east, align=right] at (-0.68,0.00) {non-\\thinking};
\fill[cNeutral!35] (0.000,0.000) circle (0.515cm);
\draw[cMuted, densely dashed, line width=.6pt] (0.000,0.000) circle (0.515cm);
\fill[cFourB] (0.000,0.000) circle (0.444cm);
\draw[cTeacher, line width=.9pt] (0.000,0.000) circle (0.463cm);
\fill[cNeutral!35] (1.400,0.000) circle (0.482cm);
\draw[cMuted, densely dashed, line width=.6pt] (1.400,0.000) circle (0.482cm);
\fill[cOneSevenB] (1.400,0.000) circle (0.398cm);
\draw[cTeacher, line width=.9pt] (1.400,0.000) circle (0.463cm);
\fill[cNeutral!35] (2.800,0.000) circle (0.429cm);
\draw[cMuted, densely dashed, line width=.6pt] (2.800,0.000) circle (0.429cm);
\fill[cZeroSixB] (2.800,0.000) circle (0.344cm);
\draw[cTeacher, line width=.9pt] (2.800,0.000) circle (0.463cm);
\node[font=\scriptsize, text=cMuted, anchor=east, align=right] at (-0.68,-1.45) {external-\\SFT};
\fill[cNeutral!35] (0.000,-1.450) circle (0.526cm);
\draw[cMuted, densely dashed, line width=.6pt] (0.000,-1.450) circle (0.526cm);
\fill[cFourB] (0.000,-1.450) circle (0.484cm);
\draw[cTeacher, line width=.9pt] (0.000,-1.450) circle (0.555cm);
\fill[cNeutral!35] (1.400,-1.450) circle (0.489cm);
\draw[cMuted, densely dashed, line width=.6pt] (1.400,-1.450) circle (0.489cm);
\fill[cOneSevenB] (1.400,-1.450) circle (0.376cm);
\draw[cTeacher, line width=.9pt] (1.400,-1.450) circle (0.555cm);
\fill[cNeutral!35] (2.800,-1.450) circle (0.461cm);
\draw[cMuted, densely dashed, line width=.6pt] (2.800,-1.450) circle (0.461cm);
\fill[cZeroSixB] (2.800,-1.450) circle (0.325cm);
\draw[cTeacher, line width=.9pt] (2.800,-1.450) circle (0.555cm);
\node[font=\scriptsize, text=cMuted, anchor=east, align=right] at (-0.68,-2.90) {teacher-\\SFT};
\fill[cNeutral!35] (0.000,-2.900) circle (0.519cm);
\draw[cMuted, densely dashed, line width=.6pt] (0.000,-2.900) circle (0.519cm);
\fill[cFourB] (0.000,-2.900) circle (0.483cm);
\draw[cTeacher, line width=.9pt] (0.000,-2.900) circle (0.555cm);
\fill[cNeutral!35] (1.400,-2.900) circle (0.431cm);
\draw[cMuted, densely dashed, line width=.6pt] (1.400,-2.900) circle (0.431cm);
\fill[cOneSevenB] (1.400,-2.900) circle (0.377cm);
\draw[cTeacher, line width=.9pt] (1.400,-2.900) circle (0.555cm);
\fill[cNeutral!35] (2.800,-2.900) circle (0.379cm);
\draw[cMuted, densely dashed, line width=.6pt] (2.800,-2.900) circle (0.379cm);
\fill[cZeroSixB] (2.800,-2.900) circle (0.307cm);
\draw[cTeacher, line width=.9pt] (2.800,-2.900) circle (0.555cm);
\fill[cFourB] (-0.9,-3.85) circle (.09); \node[font=\tiny, anchor=west] at (-0.78,-3.85) {after: $p@1$};
\fill[cNeutral!35] (0.85,-3.85) circle (.09); \draw[cMuted, densely dashed] (0.85,-3.85) circle (.09); \node[font=\tiny, anchor=west] at (0.97,-3.85) {before: $p@n$};
\draw[cTeacher, line width=.9pt] (-0.9,-4.15) circle (.09); \node[font=\tiny, anchor=west] at (-0.78,-4.15) {teacher: $p@1$};
\node[font=\footnotesize] at (7.75,1.20) {(b) what the samples do: answer and stop};
\fill[cFourB] (4.750,-3.500) rectangle (4.970,-1.689);
\fill[cFourB] (4.750,-1.689) rectangle (4.970,-1.684);
\fill[pattern=north east lines, pattern color=white] (4.750,-1.689) rectangle (4.970,-1.684);
\fill[cOneSevenB] (4.750,-1.684) rectangle (4.970,0.003);
\fill[cOneSevenB] (4.750,0.003) rectangle (4.970,0.029);
\fill[pattern=north east lines, pattern color=white] (4.750,0.003) rectangle (4.970,0.029);
\fill[cNeutral!60] (4.750,0.029) rectangle (4.970,0.216);
\fill[cNeutral!60] (4.750,0.216) rectangle (4.970,0.750);
\fill[pattern=north east lines, pattern color=white] (4.750,0.216) rectangle (4.970,0.750);
\fill[cFourB] (5.000,-3.500) rectangle (5.220,-1.348);
\fill[cFourB] (5.000,-1.348) rectangle (5.220,-1.168);
\fill[pattern=north east lines, pattern color=white] (5.000,-1.348) rectangle (5.220,-1.168);
\fill[cOneSevenB] (5.000,-1.168) rectangle (5.220,-0.099);
\fill[cOneSevenB] (5.000,-0.099) rectangle (5.220,0.519);
\fill[pattern=north east lines, pattern color=white] (5.000,-0.099) rectangle (5.220,0.519);
\fill[cNeutral!60] (5.000,0.519) rectangle (5.220,0.750);
\fill[pattern=north east lines, pattern color=white] (5.000,0.519) rectangle (5.220,0.750);
\node[font=\tiny, text=cMuted] at (4.985,-3.680) {4B};
\fill[cFourB] (5.320,-3.500) rectangle (5.540,-2.252);
\fill[cOneSevenB] (5.320,-2.252) rectangle (5.540,-0.312);
\fill[cOneSevenB] (5.320,-0.312) rectangle (5.540,-0.305);
\fill[pattern=north east lines, pattern color=white] (5.320,-0.312) rectangle (5.540,-0.305);
\fill[cNeutral!60] (5.320,-0.305) rectangle (5.540,0.143);
\fill[cNeutral!60] (5.320,0.143) rectangle (5.540,0.750);
\fill[pattern=north east lines, pattern color=white] (5.320,0.143) rectangle (5.540,0.750);
\fill[cFourB] (5.570,-3.500) rectangle (5.790,-1.637);
\fill[cFourB] (5.570,-1.637) rectangle (5.790,-1.632);
\fill[pattern=north east lines, pattern color=white] (5.570,-1.637) rectangle (5.790,-1.632);
\fill[cOneSevenB] (5.570,-1.632) rectangle (5.790,0.154);
\fill[cOneSevenB] (5.570,0.154) rectangle (5.790,0.265);
\fill[pattern=north east lines, pattern color=white] (5.570,0.154) rectangle (5.790,0.265);
\fill[cNeutral!60] (5.570,0.265) rectangle (5.790,0.750);
\fill[pattern=north east lines, pattern color=white] (5.570,0.265) rectangle (5.790,0.750);
\node[font=\tiny, text=cMuted] at (5.555,-3.680) {1.7B};
\fill[cFourB] (5.890,-3.500) rectangle (6.110,-3.301);
\fill[cFourB] (5.890,-3.301) rectangle (6.110,-3.295);
\fill[pattern=north east lines, pattern color=white] (5.890,-3.301) rectangle (6.110,-3.295);
\fill[cOneSevenB] (5.890,-3.295) rectangle (6.110,-2.725);
\fill[cOneSevenB] (5.890,-2.725) rectangle (6.110,-2.676);
\fill[pattern=north east lines, pattern color=white] (5.890,-2.725) rectangle (6.110,-2.676);
\fill[cNeutral!60] (5.890,-2.676) rectangle (6.110,-2.516);
\fill[cNeutral!60] (5.890,-2.516) rectangle (6.110,0.750);
\fill[pattern=north east lines, pattern color=white] (5.890,-2.516) rectangle (6.110,0.750);
\fill[cFourB] (6.140,-3.500) rectangle (6.360,-2.107);
\fill[cFourB] (6.140,-2.107) rectangle (6.360,-2.105);
\fill[pattern=north east lines, pattern color=white] (6.140,-2.107) rectangle (6.360,-2.105);
\fill[cOneSevenB] (6.140,-2.105) rectangle (6.360,-0.093);
\fill[cOneSevenB] (6.140,-0.093) rectangle (6.360,0.014);
\fill[pattern=north east lines, pattern color=white] (6.140,-0.093) rectangle (6.360,0.014);
\fill[cNeutral!60] (6.140,0.014) rectangle (6.360,0.750);
\fill[pattern=north east lines, pattern color=white] (6.140,0.014) rectangle (6.360,0.750);
\node[font=\tiny, text=cMuted] at (6.125,-3.680) {0.6B};
\node[font=\scriptsize, text=cMuted] at (5.555,-3.950) {non-thinking};
\fill[cFourB] (6.810,-3.500) rectangle (7.030,-1.246);
\fill[cFourB] (6.810,-1.246) rectangle (7.030,-1.239);
\fill[pattern=north east lines, pattern color=white] (6.810,-1.246) rectangle (7.030,-1.239);
\fill[cOneSevenB] (6.810,-1.239) rectangle (7.030,-0.405);
\fill[cOneSevenB] (6.810,-0.405) rectangle (7.030,-0.381);
\fill[pattern=north east lines, pattern color=white] (6.810,-0.405) rectangle (7.030,-0.381);
\fill[cNeutral!60] (6.810,-0.381) rectangle (7.030,0.750);
\fill[pattern=north east lines, pattern color=white] (6.810,-0.381) rectangle (7.030,0.750);
\fill[cFourB] (7.060,-3.500) rectangle (7.280,-0.756);
\fill[cFourB] (7.060,-0.756) rectangle (7.280,-0.733);
\fill[pattern=north east lines, pattern color=white] (7.060,-0.756) rectangle (7.280,-0.733);
\fill[cOneSevenB] (7.060,-0.733) rectangle (7.280,-0.436);
\fill[cOneSevenB] (7.060,-0.436) rectangle (7.280,-0.405);
\fill[pattern=north east lines, pattern color=white] (7.060,-0.436) rectangle (7.280,-0.405);
\fill[cNeutral!60] (7.060,-0.405) rectangle (7.280,0.750);
\fill[pattern=north east lines, pattern color=white] (7.060,-0.405) rectangle (7.280,0.750);
\node[font=\tiny, text=cMuted] at (7.045,-3.680) {4B};
\fill[cFourB] (7.380,-3.500) rectangle (7.600,-1.983);
\fill[cFourB] (7.380,-1.983) rectangle (7.600,-1.976);
\fill[pattern=north east lines, pattern color=white] (7.380,-1.983) rectangle (7.600,-1.976);
\fill[cOneSevenB] (7.380,-1.976) rectangle (7.600,-0.888);
\fill[cOneSevenB] (7.380,-0.888) rectangle (7.600,-0.835);
\fill[pattern=north east lines, pattern color=white] (7.380,-0.888) rectangle (7.600,-0.835);
\fill[cNeutral!60] (7.380,-0.835) rectangle (7.600,-0.834);
\fill[cNeutral!60] (7.380,-0.834) rectangle (7.600,0.750);
\fill[pattern=north east lines, pattern color=white] (7.380,-0.834) rectangle (7.600,0.750);
\fill[cFourB] (7.630,-3.500) rectangle (7.850,-1.871);
\fill[cFourB] (7.630,-1.871) rectangle (7.850,-1.833);
\fill[pattern=north east lines, pattern color=white] (7.630,-1.871) rectangle (7.850,-1.833);
\fill[cOneSevenB] (7.630,-1.833) rectangle (7.850,-1.507);
\fill[cOneSevenB] (7.630,-1.507) rectangle (7.850,-1.451);
\fill[pattern=north east lines, pattern color=white] (7.630,-1.507) rectangle (7.850,-1.451);
\fill[cNeutral!60] (7.630,-1.451) rectangle (7.850,0.750);
\fill[pattern=north east lines, pattern color=white] (7.630,-1.451) rectangle (7.850,0.750);
\node[font=\tiny, text=cMuted] at (7.615,-3.680) {1.7B};
\fill[cFourB] (7.950,-3.500) rectangle (8.170,-2.517);
\fill[cFourB] (7.950,-2.517) rectangle (8.170,-2.488);
\fill[pattern=north east lines, pattern color=white] (7.950,-2.517) rectangle (8.170,-2.488);
\fill[cOneSevenB] (7.950,-2.488) rectangle (8.170,-0.889);
\fill[cOneSevenB] (7.950,-0.889) rectangle (8.170,-0.646);
\fill[pattern=north east lines, pattern color=white] (7.950,-0.889) rectangle (8.170,-0.646);
\fill[cNeutral!60] (7.950,-0.646) rectangle (8.170,-0.643);
\fill[cNeutral!60] (7.950,-0.643) rectangle (8.170,0.750);
\fill[pattern=north east lines, pattern color=white] (7.950,-0.643) rectangle (8.170,0.750);
\fill[cFourB] (8.200,-3.500) rectangle (8.420,-2.326);
\fill[cFourB] (8.200,-2.326) rectangle (8.420,-2.250);
\fill[pattern=north east lines, pattern color=white] (8.200,-2.326) rectangle (8.420,-2.250);
\fill[cOneSevenB] (8.200,-2.250) rectangle (8.420,-1.839);
\fill[cOneSevenB] (8.200,-1.839) rectangle (8.420,-1.710);
\fill[pattern=north east lines, pattern color=white] (8.200,-1.839) rectangle (8.420,-1.710);
\fill[cNeutral!60] (8.200,-1.710) rectangle (8.420,0.750);
\fill[pattern=north east lines, pattern color=white] (8.200,-1.710) rectangle (8.420,0.750);
\node[font=\tiny, text=cMuted] at (8.185,-3.680) {0.6B};
\node[font=\scriptsize, text=cMuted] at (7.615,-3.950) {external-SFT};
\fill[cFourB] (8.870,-3.500) rectangle (9.090,-2.133);
\fill[cFourB] (8.870,-2.133) rectangle (9.090,-2.099);
\fill[pattern=north east lines, pattern color=white] (8.870,-2.133) rectangle (9.090,-2.099);
\fill[cOneSevenB] (8.870,-2.099) rectangle (9.090,-1.967);
\fill[cOneSevenB] (8.870,-1.967) rectangle (9.090,-1.943);
\fill[pattern=north east lines, pattern color=white] (8.870,-1.967) rectangle (9.090,-1.943);
\fill[cNeutral!60] (8.870,-1.943) rectangle (9.090,-1.560);
\fill[cNeutral!60] (8.870,-1.560) rectangle (9.090,0.750);
\fill[pattern=north east lines, pattern color=white] (8.870,-1.560) rectangle (9.090,0.750);
\fill[cFourB] (9.120,-3.500) rectangle (9.340,-0.781);
\fill[cFourB] (9.120,-0.781) rectangle (9.340,-0.741);
\fill[pattern=north east lines, pattern color=white] (9.120,-0.781) rectangle (9.340,-0.741);
\fill[cOneSevenB] (9.120,-0.741) rectangle (9.340,-0.419);
\fill[cOneSevenB] (9.120,-0.419) rectangle (9.340,-0.378);
\fill[pattern=north east lines, pattern color=white] (9.120,-0.419) rectangle (9.340,-0.378);
\fill[cNeutral!60] (9.120,-0.378) rectangle (9.340,0.750);
\fill[pattern=north east lines, pattern color=white] (9.120,-0.378) rectangle (9.340,0.750);
\node[font=\tiny, text=cMuted] at (9.105,-3.680) {4B};
\fill[cFourB] (9.440,-3.500) rectangle (9.660,-3.116);
\fill[cFourB] (9.440,-3.116) rectangle (9.660,-2.955);
\fill[pattern=north east lines, pattern color=white] (9.440,-3.116) rectangle (9.660,-2.955);
\fill[cOneSevenB] (9.440,-2.955) rectangle (9.660,-2.906);
\fill[cOneSevenB] (9.440,-2.906) rectangle (9.660,-2.857);
\fill[pattern=north east lines, pattern color=white] (9.440,-2.906) rectangle (9.660,-2.857);
\fill[cNeutral!60] (9.440,-2.857) rectangle (9.660,-2.182);
\fill[cNeutral!60] (9.440,-2.182) rectangle (9.660,0.750);
\fill[pattern=north east lines, pattern color=white] (9.440,-2.182) rectangle (9.660,0.750);
\fill[cFourB] (9.690,-3.500) rectangle (9.910,-1.825);
\fill[pattern=north east lines, pattern color=white] (9.690,-3.500) rectangle (9.910,-1.825);
\fill[cOneSevenB] (9.690,-1.825) rectangle (9.910,-1.522);
\fill[pattern=north east lines, pattern color=white] (9.690,-1.825) rectangle (9.910,-1.522);
\fill[cNeutral!60] (9.690,-1.522) rectangle (9.910,0.750);
\fill[pattern=north east lines, pattern color=white] (9.690,-1.522) rectangle (9.910,0.750);
\node[font=\tiny, text=cMuted] at (9.675,-3.680) {1.7B};
\fill[cFourB] (10.010,-3.500) rectangle (10.230,-3.342);
\fill[cFourB] (10.010,-3.342) rectangle (10.230,-3.266);
\fill[pattern=north east lines, pattern color=white] (10.010,-3.342) rectangle (10.230,-3.266);
\fill[cOneSevenB] (10.010,-3.266) rectangle (10.230,-3.184);
\fill[cOneSevenB] (10.010,-3.184) rectangle (10.230,-3.122);
\fill[pattern=north east lines, pattern color=white] (10.010,-3.184) rectangle (10.230,-3.122);
\fill[cNeutral!60] (10.010,-3.122) rectangle (10.230,-2.569);
\fill[cNeutral!60] (10.010,-2.569) rectangle (10.230,0.750);
\fill[pattern=north east lines, pattern color=white] (10.010,-2.569) rectangle (10.230,0.750);
\fill[cFourB] (10.260,-3.500) rectangle (10.480,-2.385);
\fill[pattern=north east lines, pattern color=white] (10.260,-3.500) rectangle (10.480,-2.385);
\fill[cOneSevenB] (10.260,-2.385) rectangle (10.480,-2.083);
\fill[pattern=north east lines, pattern color=white] (10.260,-2.385) rectangle (10.480,-2.083);
\fill[cNeutral!60] (10.260,-2.083) rectangle (10.480,0.750);
\fill[pattern=north east lines, pattern color=white] (10.260,-2.083) rectangle (10.480,0.750);
\node[font=\tiny, text=cMuted] at (10.245,-3.680) {0.6B};
\node[font=\scriptsize, text=cMuted] at (9.675,-3.950) {teacher-SFT};
\node[font=\tiny, anchor=west] at (4.75,-4.35) {each pair: before, after};
\fill[cFourB] (7.15,-4.42) rectangle ++(.18,.14); \node[font=\tiny, anchor=west] at (7.35,-4.35) {right};
\fill[cOneSevenB] (8.10,-4.42) rectangle ++(.18,.14); \node[font=\tiny, anchor=west] at (8.30,-4.35) {only wrong};
\fill[cNeutral!60] (9.50,-4.42) rectangle ++(.18,.14); \node[font=\tiny, anchor=west] at (9.70,-4.35) {no answer};
\fill[cMuted] (7.15,-4.72) rectangle ++(.18,.14); \fill[pattern=north east lines, pattern color=white] (7.15,-4.72) rectangle ++(.18,.14); \node[font=\tiny, anchor=west] at (7.35,-4.65) {hatched: never stopped (cut off at the limit)};
\end{tikzpicture}
\caption{\textbf{Overview.} Qwen3-8B distilled into Qwen3 4B, 1.7B and 0.6B, on three routes; three-benchmark
mean, lenient scoring. (a)~Circle areas are proportional to accuracy. After distillation, a student's single-attempt
accuracy (colored disk) stays inside what the start reached in many attempts (dashed), and the smaller the student, the
further inside the teacher's (black ring). (b)~Every sample, before
and after distillation, by its marked answer and by whether it stopped. Without thinking, students keep stopping and
wrong answers turn right; in thinking mode the small students trade finished wrong answers for samples that never stop
and mark nothing.}
\label{fig:overview}
\end{figure}
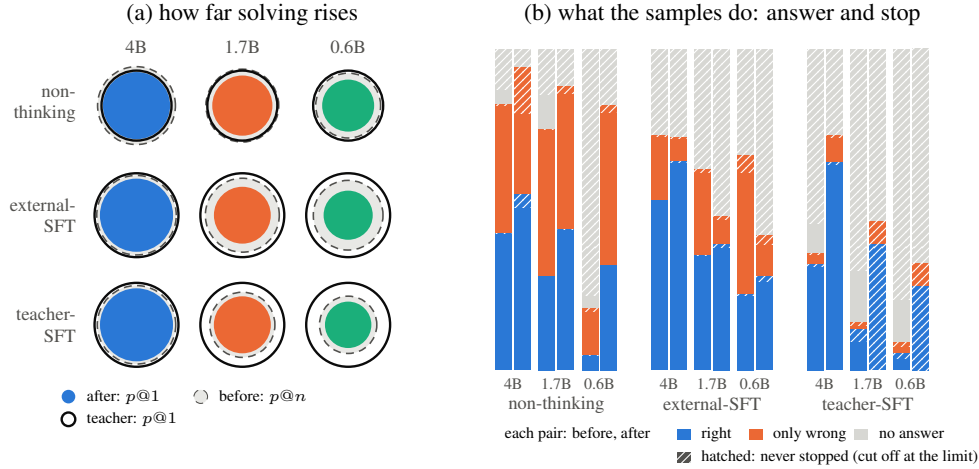

%% file: sections/02_setup.tex
\section{Setup}
\label{sec:setup}

The teacher is Qwen3-8B \citep{qwen3}. In thinking mode it reasons inside \texttt{<think>}\,\ldots\,\think{} and then
answers; in non-thinking mode its chat template pre-fills an empty thinking block and it answers directly. The students
are Qwen3 4B, 1.7B and 0.6B and always run in the teacher's mode. For thinking mode each student first learns the format
by supervised fine-tuning, either on 5,878 short external solutions (the \emph{external-SFT start}) or on 1,524 of the teacher's own correct solutions (the \emph{teacher-SFT start}), and is then distilled for 142 steps. As a reference,
the pretrained-only students are distilled from the non-thinking teacher for 212 steps (the \emph{non-thinking route});
Appendix~\ref{app:setup} lists every run. Distillation samples one response $y$ per prompt $x$, the teacher scores the
same tokens, and each token receives a clipped one-sample estimate of the reverse KL divergence as its advantage in a
PPO-style update, with no task reward, following \citet{tml2025opd} (see also \citealp{agarwal2024gkd,gu2024minillm}):
\begin{equation}
\label{eq:opd}
A_t = -\operatorname{clip}\!\big(\log \pi_{\mathrm{old}}(y_t \mid x, y_{<t}) - \log \pi_T(y_t \mid x, y_{<t}),\; -1,\; 1\big).
\end{equation}
The token \think{} is scored like any other, so the only pressure to stop comes from how likely the teacher is to stop
at the point the student has reached.

Training uses 13,597 DAPO-Math-17k problems \citep{yu2025dapo}, 128 per step, and a training response may use at most
7,168 tokens (the \emph{rollout budget}). We evaluate on AIME 2026, AMC23 and GSM8K-200 with 48, 48 and 32 samples per
problem at temperature 0.6, and a limit of 30,720 tokens (8,192 for GSM8K-200); during training we validate every 20
steps on AIME 2025 and AMC23. Distillation converges early in both modes (Appendix Figure~\ref{fig:valk}): after the first steps
neither validation p@1 nor p@8 changes clearly on any run. A \emph{marked answer} is a value the model itself presents as its answer, by
writing it in \verb|\boxed{}| or on a line that begins with \texttt{Answer:} or \texttt{The final answer is}; it may
appear anywhere in a sample, inside the reasoning or after it, and a sample may mark several. A sample is correct under
\emph{lenient} scoring, our main score, if any of its marked answers is correct. A value that only appears in the
working, for instance as an intermediate result such as ``so $x = 12$'', is not a marked answer and does not count, even
if it happens to equal the correct answer. p@1 is
the average over samples and p@n the fraction of problems solved in at least one of the $n$ samples. Within a benchmark,
lenient p@1 is exactly the \emph{marked-answer rate}, the fraction of samples with a marked answer, times the accuracy of
the samples that have one. We report two behaviors beside the score, not inside it: the \emph{closure rate}, the
fraction of samples that close their reasoning with \think{}, and \emph{truncation}, the fraction that reach the token
limit. Appendix~\ref{app:scoring} defines two stricter scorings; GPQA-Diamond results for the non-thinking route are in
Appendix~\ref{app:block1}. All intervals are
paired 95\% bootstrap intervals over problems (Appendix~\ref{app:scoring}). When we compare accuracy given a marked answer between two
models, we use only the problems both answered.

%% file: sections/03_ceiling.tex
\section{The ceiling of on-policy distillation}
\label{sec:ceiling}

On-policy distillation raises every student's mean single-attempt accuracy, but only up to two ceilings, and no
benchmark breaks the first (Figure~\ref{fig:overview}a; per benchmark in Figures~\ref{fig:ownceil} and~\ref{fig:teachceil}; values in Table~\ref{tab:results}). The first
ceiling is the student itself. In all 27 combinations of route, size and benchmark, the distilled lenient p@1 stays below
the p@n of the student it started from, the fraction of problems that student solved at least once in all the samples we
drew. Over the three benchmarks the distilled p@1 reaches between 0.50 and 0.87 of the starting p@n, with every paired difference below
zero (Appendix Table~\ref{tab:app-ceiling}). The second ceiling is the teacher in the student's mode
and the smaller the student, the farther below it the student stays. Over the three
benchmarks the distilled p@1 reaches 0.922, 0.739 and 0.551 of the non-thinking teacher's from 4B to 0.6B, and 0.760,
0.458 and 0.343 of the thinking teacher's from the external-SFT start. The gap is widest on the hardest benchmark: on
AIME 2026 the 1.7B and 0.6B distilled in thinking mode reach a tenth of the teacher's p@1 or less. In coverage the
students come closer, and on GSM8K-200 the distilled p@n comes close to the teacher's at every size. The thinking teacher
is also much stronger, 0.857 against 0.595 in p@1, so a smaller share of it is not a lower score.

\input{figures/fig_ownceil}

\input{figures/fig_teachceil}

\input{tables/tab_results}

\input{figures/fig_coverage}

The top of the range itself barely moves (Figure~\ref{fig:coverage}). The three-benchmark p@n changes by about 0.1 or less for every student
distilled from the non-thinking teacher or from the external-SFT start, and grows a little more from the
teacher-SFT start, by $+0.134$ at 4B and $+0.103$ at 1.7B, as those students learn to write their answers down. The clear expansions are the 4B in thinking mode on AIME 2026, the hardest benchmark, where p@n rises from 0.400 to
0.700, and to 0.733 from the teacher-SFT start, and the pretrained-only 0.6B on GSM8K-200, from 0.745 to 0.945, which
learns to write its answer down. At 1.7B and 0.6B from the external-SFT start, p@n ends slightly
lower, by 0.036 and 0.070.

\input{figures/fig_bins}

The gain comes from problems the student could already solve (Figure~\ref{fig:bins}). Problems the start never solved
in its $n$ samples account for 0.23 to 0.60 of the three-benchmark mean, yet they supply at most an eighth of the p@1
gain: after distillation they are solved in between 0.002 and 0.183 of attempts, and from the external-SFT start in at
most 0.046. At 4B from the external-SFT start, the problems solved in fewer than half of the samples rise from 0.204 to
0.496; at 1.7B and 0.6B the same group rises much less, from 0.200 to 0.268 and from 0.192 to 0.301, and more problems
leave the reachable set than enter it (11 against 3, and 22 against 9; Appendix Table~\ref{tab:app-bins}).

What limits the small students in thinking mode is not the quality of their answers but their number. On the
non-thinking route the students keep stopping (Figure~\ref{fig:closure}), and lenient p@1 rises at every size, by $+0.121$, $+0.146$ and, clearly, $+0.280$. In thinking
mode from the external-SFT start, answers are right more often at every size: on the problems both the starting and
the distilled student answered, by $+0.190$, $+0.106$ and $+0.115$, all with intervals above zero (Appendix
Table~\ref{tab:app-decomp}). The 4B marks answers as often as before ($-0.006$, within noise), so its lenient p@1 rises by
$+0.119$. The 1.7B and 0.6B mark answers much less often ($-0.145$ and $-0.251$), and their lenient p@1 rises only by
$+0.034$ and $+0.056$. Their missing answers belong to samples that never stopped, which the next two sections explain.

%% file: figures/fig_ownceil.tex
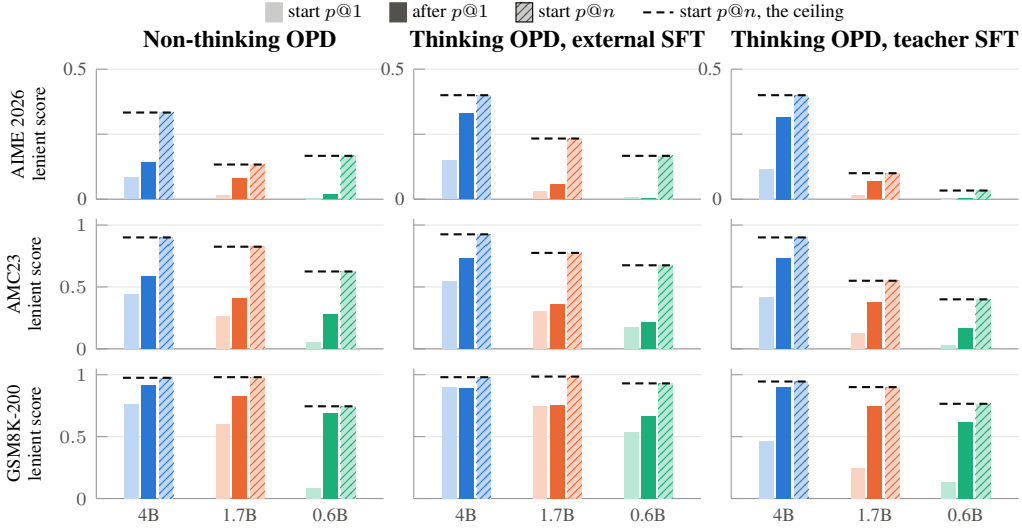
\begin{figure}[t]
\centering
\begin{tikzpicture}
\begin{groupplot}[group style={group size=3 by 3, horizontal sep=0.40cm, vertical sep=0.26cm,
    xticklabels at=edge bottom, yticklabels at=edge left, ylabels at=edge left, group name=own},
  paperstyle, scale only axis, width=3.80cm, height=1.72cm, xmajorgrids=false,
  xmin=0.42, xmax=3.58, ymin=0, ymax=1.05, enlargelimits=false, clip=false,
  xtick={1,2,3}, xticklabels={4B,1.7B,0.6B}, xtick style={draw=none}, ytick={0,0.5,1},
  axis x line*=bottom, axis y line*=left, title style={yshift=-3pt},
  ylabel style={align=center, font=\scriptsize}]
\nextgroupplot[title={\textbf{Non-thinking OPD}}, ylabel={AIME 2026\\lenient score}, ymax=0.5, ytick={0,0.25,0.5}, yticklabels={0,,0.5}]
\fill[cFourB!30] (axis cs:0.733,0) rectangle (axis cs:0.888,0.0826);
\fill[cFourB] (axis cs:0.922,0) rectangle (axis cs:1.077,0.1410);
\fill[cFourB!30] (axis cs:1.113,0) rectangle (axis cs:1.267,0.3333);
\fill[pattern=north east lines, pattern color=cFourB] (axis cs:1.113,0) rectangle (axis cs:1.267,0.3333);
\draw[cTeacher, line width=.8pt, densely dashed] (axis cs:0.713,0.3333) -- (axis cs:1.267,0.3333);
\fill[cOneSevenB!30] (axis cs:1.733,0) rectangle (axis cs:1.887,0.0139);
\fill[cOneSevenB] (axis cs:1.923,0) rectangle (axis cs:2.078,0.0792);
\fill[cOneSevenB!30] (axis cs:2.112,0) rectangle (axis cs:2.268,0.1333);
\fill[pattern=north east lines, pattern color=cOneSevenB] (axis cs:2.112,0) rectangle (axis cs:2.268,0.1333);
\draw[cTeacher, line width=.8pt, densely dashed] (axis cs:1.713,0.1333) -- (axis cs:2.268,0.1333);
\fill[cZeroSixB!30] (axis cs:2.732,0) rectangle (axis cs:2.888,0.0042);
\fill[cZeroSixB] (axis cs:2.922,0) rectangle (axis cs:3.078,0.0181);
\fill[cZeroSixB!30] (axis cs:3.112,0) rectangle (axis cs:3.268,0.1667);
\fill[pattern=north east lines, pattern color=cZeroSixB] (axis cs:3.112,0) rectangle (axis cs:3.268,0.1667);
\draw[cTeacher, line width=.8pt, densely dashed] (axis cs:2.712,0.1667) -- (axis cs:3.268,0.1667);
\nextgroupplot[title={\textbf{Thinking OPD, external SFT}}, ymax=0.5, ytick={0,0.25,0.5}, yticklabels={0,,0.5}, legend style={at={(0.5,1)}, anchor=south, yshift=14pt, legend columns=4, /tikz/every even column/.append style={column sep=0.8em}}]
\addlegendimage{legend image code/.code={\fill[cMuted!30] (0cm,-0.8ex) rectangle (0.22cm,0.8ex);}}
\addlegendentry{start $p@1$}
\addlegendimage{legend image code/.code={\fill[cMuted] (0cm,-0.8ex) rectangle (0.22cm,0.8ex);}}
\addlegendentry{after $p@1$}
\addlegendimage{legend image code/.code={\fill[cMuted!30] (0cm,-0.8ex) rectangle (0.22cm,0.8ex); \fill[pattern=north east lines, pattern color=cMuted] (0cm,-0.8ex) rectangle (0.22cm,0.8ex);}}
\addlegendentry{start $p@n$}
\addlegendimage{legend image code/.code={\draw[cTeacher, line width=.8pt, densely dashed] (0cm,0cm) -- (0.4cm,0cm);}}
\addlegendentry{start $p@n$, the ceiling}
\fill[cFourB!30] (axis cs:0.733,0) rectangle (axis cs:0.888,0.1479);
\fill[cFourB] (axis cs:0.922,0) rectangle (axis cs:1.077,0.3292);
\fill[cFourB!30] (axis cs:1.113,0) rectangle (axis cs:1.267,0.4000);
\fill[pattern=north east lines, pattern color=cFourB] (axis cs:1.113,0) rectangle (axis cs:1.267,0.4000);
\draw[cTeacher, line width=.8pt, densely dashed] (axis cs:0.713,0.4000) -- (axis cs:1.267,0.4000);
\fill[cOneSevenB!30] (axis cs:1.733,0) rectangle (axis cs:1.887,0.0292);
\fill[cOneSevenB] (axis cs:1.923,0) rectangle (axis cs:2.078,0.0569);
\fill[cOneSevenB!30] (axis cs:2.112,0) rectangle (axis cs:2.268,0.2333);
\fill[pattern=north east lines, pattern color=cOneSevenB] (axis cs:2.112,0) rectangle (axis cs:2.268,0.2333);
\draw[cTeacher, line width=.8pt, densely dashed] (axis cs:1.713,0.2333) -- (axis cs:2.268,0.2333);
\fill[cZeroSixB!30] (axis cs:2.732,0) rectangle (axis cs:2.888,0.0056);
\fill[cZeroSixB] (axis cs:2.922,0) rectangle (axis cs:3.078,0.0035);
\fill[cZeroSixB!30] (axis cs:3.112,0) rectangle (axis cs:3.268,0.1667);
\fill[pattern=north east lines, pattern color=cZeroSixB] (axis cs:3.112,0) rectangle (axis cs:3.268,0.1667);
\draw[cTeacher, line width=.8pt, densely dashed] (axis cs:2.712,0.1667) -- (axis cs:3.268,0.1667);
\nextgroupplot[title={\textbf{Thinking OPD, teacher SFT}}, ymax=0.5, ytick={0,0.25,0.5}, yticklabels={0,,0.5}]
\fill[cFourB!30] (axis cs:0.733,0) rectangle (axis cs:0.888,0.1160);
\fill[cFourB] (axis cs:0.922,0) rectangle (axis cs:1.077,0.3160);
\fill[cFourB!30] (axis cs:1.113,0) rectangle (axis cs:1.267,0.4000);
\fill[pattern=north east lines, pattern color=cFourB] (axis cs:1.113,0) rectangle (axis cs:1.267,0.4000);
\draw[cTeacher, line width=.8pt, densely dashed] (axis cs:0.713,0.4000) -- (axis cs:1.267,0.4000);
\fill[cOneSevenB!30] (axis cs:1.733,0) rectangle (axis cs:1.887,0.0132);
\fill[cOneSevenB] (axis cs:1.923,0) rectangle (axis cs:2.078,0.0674);
\fill[cOneSevenB!30] (axis cs:2.112,0) rectangle (axis cs:2.268,0.1000);
\fill[pattern=north east lines, pattern color=cOneSevenB] (axis cs:2.112,0) rectangle (axis cs:2.268,0.1000);
\draw[cTeacher, line width=.8pt, densely dashed] (axis cs:1.713,0.1000) -- (axis cs:2.268,0.1000);
\fill[cZeroSixB!30] (axis cs:2.732,0) rectangle (axis cs:2.888,0.0014);
\fill[cZeroSixB] (axis cs:2.922,0) rectangle (axis cs:3.078,0.0028);
\fill[cZeroSixB!30] (axis cs:3.112,0) rectangle (axis cs:3.268,0.0333);
\fill[pattern=north east lines, pattern color=cZeroSixB] (axis cs:3.112,0) rectangle (axis cs:3.268,0.0333);
\draw[cTeacher, line width=.8pt, densely dashed] (axis cs:2.712,0.0333) -- (axis cs:3.268,0.0333);
\nextgroupplot[ylabel={AMC23\\lenient score}]
\fill[cFourB!30] (axis cs:0.733,0) rectangle (axis cs:0.888,0.4380);
\fill[cFourB] (axis cs:0.922,0) rectangle (axis cs:1.077,0.5865);
\fill[cFourB!30] (axis cs:1.113,0) rectangle (axis cs:1.267,0.9000);
\fill[pattern=north east lines, pattern color=cFourB] (axis cs:1.113,0) rectangle (axis cs:1.267,0.9000);
\draw[cTeacher, line width=.8pt, densely dashed] (axis cs:0.713,0.9000) -- (axis cs:1.267,0.9000);
\fill[cOneSevenB!30] (axis cs:1.733,0) rectangle (axis cs:1.887,0.2661);
\fill[cOneSevenB] (axis cs:1.923,0) rectangle (axis cs:2.078,0.4099);
\fill[cOneSevenB!30] (axis cs:2.112,0) rectangle (axis cs:2.268,0.8250);
\fill[pattern=north east lines, pattern color=cOneSevenB] (axis cs:2.112,0) rectangle (axis cs:2.268,0.8250);
\draw[cTeacher, line width=.8pt, densely dashed] (axis cs:1.713,0.8250) -- (axis cs:2.268,0.8250);
\fill[cZeroSixB!30] (axis cs:2.732,0) rectangle (axis cs:2.888,0.0531);
\fill[cZeroSixB] (axis cs:2.922,0) rectangle (axis cs:3.078,0.2776);
\fill[cZeroSixB!30] (axis cs:3.112,0) rectangle (axis cs:3.268,0.6250);
\fill[pattern=north east lines, pattern color=cZeroSixB] (axis cs:3.112,0) rectangle (axis cs:3.268,0.6250);
\draw[cTeacher, line width=.8pt, densely dashed] (axis cs:2.712,0.6250) -- (axis cs:3.268,0.6250);
\nextgroupplot[]
\fill[cFourB!30] (axis cs:0.733,0) rectangle (axis cs:0.888,0.5484);
\fill[cFourB] (axis cs:0.922,0) rectangle (axis cs:1.077,0.7276);
\fill[cFourB!30] (axis cs:1.113,0) rectangle (axis cs:1.267,0.9250);
\fill[pattern=north east lines, pattern color=cFourB] (axis cs:1.113,0) rectangle (axis cs:1.267,0.9250);
\draw[cTeacher, line width=.8pt, densely dashed] (axis cs:0.713,0.9250) -- (axis cs:1.267,0.9250);
\fill[cOneSevenB!30] (axis cs:1.733,0) rectangle (axis cs:1.887,0.2995);
\fill[cOneSevenB] (axis cs:1.923,0) rectangle (axis cs:2.078,0.3625);
\fill[cOneSevenB!30] (axis cs:2.112,0) rectangle (axis cs:2.268,0.7750);
\fill[pattern=north east lines, pattern color=cOneSevenB] (axis cs:2.112,0) rectangle (axis cs:2.268,0.7750);
\draw[cTeacher, line width=.8pt, densely dashed] (axis cs:1.713,0.7750) -- (axis cs:2.268,0.7750);
\fill[cZeroSixB!30] (axis cs:2.732,0) rectangle (axis cs:2.888,0.1729);
\fill[cZeroSixB] (axis cs:2.922,0) rectangle (axis cs:3.078,0.2120);
\fill[cZeroSixB!30] (axis cs:3.112,0) rectangle (axis cs:3.268,0.6750);
\fill[pattern=north east lines, pattern color=cZeroSixB] (axis cs:3.112,0) rectangle (axis cs:3.268,0.6750);
\draw[cTeacher, line width=.8pt, densely dashed] (axis cs:2.712,0.6750) -- (axis cs:3.268,0.6750);
\nextgroupplot[]
\fill[cFourB!30] (axis cs:0.733,0) rectangle (axis cs:0.888,0.4120);
\fill[cFourB] (axis cs:0.922,0) rectangle (axis cs:1.077,0.7281);
\fill[cFourB!30] (axis cs:1.113,0) rectangle (axis cs:1.267,0.9000);
\fill[pattern=north east lines, pattern color=cFourB] (axis cs:1.113,0) rectangle (axis cs:1.267,0.9000);
\draw[cTeacher, line width=.8pt, densely dashed] (axis cs:0.713,0.9000) -- (axis cs:1.267,0.9000);
\fill[cOneSevenB!30] (axis cs:1.733,0) rectangle (axis cs:1.887,0.1286);
\fill[cOneSevenB] (axis cs:1.923,0) rectangle (axis cs:2.078,0.3719);
\fill[cOneSevenB!30] (axis cs:2.112,0) rectangle (axis cs:2.268,0.5500);
\fill[pattern=north east lines, pattern color=cOneSevenB] (axis cs:2.112,0) rectangle (axis cs:2.268,0.5500);
\draw[cTeacher, line width=.8pt, densely dashed] (axis cs:1.713,0.5500) -- (axis cs:2.268,0.5500);
\fill[cZeroSixB!30] (axis cs:2.732,0) rectangle (axis cs:2.888,0.0286);
\fill[cZeroSixB] (axis cs:2.922,0) rectangle (axis cs:3.078,0.1646);
\fill[cZeroSixB!30] (axis cs:3.112,0) rectangle (axis cs:3.268,0.4000);
\fill[pattern=north east lines, pattern color=cZeroSixB] (axis cs:3.112,0) rectangle (axis cs:3.268,0.4000);
\draw[cTeacher, line width=.8pt, densely dashed] (axis cs:2.712,0.4000) -- (axis cs:3.268,0.4000);
\nextgroupplot[ylabel={GSM8K-200\\lenient score}]
\fill[cFourB!30] (axis cs:0.733,0) rectangle (axis cs:0.888,0.7609);
\fill[cFourB] (axis cs:0.922,0) rectangle (axis cs:1.077,0.9187);
\fill[cFourB!30] (axis cs:1.113,0) rectangle (axis cs:1.267,0.9750);
\fill[pattern=north east lines, pattern color=cFourB] (axis cs:1.113,0) rectangle (axis cs:1.267,0.9750);
\draw[cTeacher, line width=.8pt, densely dashed] (axis cs:0.713,0.9750) -- (axis cs:1.267,0.9750);
\fill[cOneSevenB!30] (axis cs:1.733,0) rectangle (axis cs:1.887,0.6009);
\fill[cOneSevenB] (axis cs:1.923,0) rectangle (axis cs:2.078,0.8297);
\fill[cOneSevenB!30] (axis cs:2.112,0) rectangle (axis cs:2.268,0.9800);
\fill[pattern=north east lines, pattern color=cOneSevenB] (axis cs:2.112,0) rectangle (axis cs:2.268,0.9800);
\draw[cTeacher, line width=.8pt, densely dashed] (axis cs:1.713,0.9800) -- (axis cs:2.268,0.9800);
\fill[cZeroSixB!30] (axis cs:2.732,0) rectangle (axis cs:2.888,0.0873);
\fill[cZeroSixB] (axis cs:2.922,0) rectangle (axis cs:3.078,0.6887);
\fill[cZeroSixB!30] (axis cs:3.112,0) rectangle (axis cs:3.268,0.7450);
\fill[pattern=north east lines, pattern color=cZeroSixB] (axis cs:3.112,0) rectangle (axis cs:3.268,0.7450);
\draw[cTeacher, line width=.8pt, densely dashed] (axis cs:2.712,0.7450) -- (axis cs:3.268,0.7450);
\nextgroupplot[]
\fill[cFourB!30] (axis cs:0.733,0) rectangle (axis cs:0.888,0.8992);
\fill[cFourB] (axis cs:0.922,0) rectangle (axis cs:1.077,0.8964);
\fill[cFourB!30] (axis cs:1.113,0) rectangle (axis cs:1.267,0.9800);
\fill[pattern=north east lines, pattern color=cFourB] (axis cs:1.113,0) rectangle (axis cs:1.267,0.9800);
\draw[cTeacher, line width=.8pt, densely dashed] (axis cs:0.713,0.9800) -- (axis cs:1.267,0.9800);
\fill[cOneSevenB!30] (axis cs:1.733,0) rectangle (axis cs:1.887,0.7467);
\fill[cOneSevenB] (axis cs:1.923,0) rectangle (axis cs:2.078,0.7572);
\fill[cOneSevenB!30] (axis cs:2.112,0) rectangle (axis cs:2.268,0.9850);
\fill[pattern=north east lines, pattern color=cOneSevenB] (axis cs:2.112,0) rectangle (axis cs:2.268,0.9850);
\draw[cTeacher, line width=.8pt, densely dashed] (axis cs:1.713,0.9850) -- (axis cs:2.268,0.9850);
\fill[cZeroSixB!30] (axis cs:2.732,0) rectangle (axis cs:2.888,0.5356);
\fill[cZeroSixB] (axis cs:2.922,0) rectangle (axis cs:3.078,0.6669);
\fill[cZeroSixB!30] (axis cs:3.112,0) rectangle (axis cs:3.268,0.9300);
\fill[pattern=north east lines, pattern color=cZeroSixB] (axis cs:3.112,0) rectangle (axis cs:3.268,0.9300);
\draw[cTeacher, line width=.8pt, densely dashed] (axis cs:2.712,0.9300) -- (axis cs:3.268,0.9300);
\nextgroupplot[]
\fill[cFourB!30] (axis cs:0.733,0) rectangle (axis cs:0.888,0.4611);
\fill[cFourB] (axis cs:0.922,0) rectangle (axis cs:1.077,0.9030);
\fill[cFourB!30] (axis cs:1.113,0) rectangle (axis cs:1.267,0.9450);
\fill[pattern=north east lines, pattern color=cFourB] (axis cs:1.113,0) rectangle (axis cs:1.267,0.9450);
\draw[cTeacher, line width=.8pt, densely dashed] (axis cs:0.713,0.9450) -- (axis cs:1.267,0.9450);
\fill[cOneSevenB!30] (axis cs:1.733,0) rectangle (axis cs:1.887,0.2427);
\fill[cOneSevenB] (axis cs:1.923,0) rectangle (axis cs:2.078,0.7430);
\fill[cOneSevenB!30] (axis cs:2.112,0) rectangle (axis cs:2.268,0.9000);
\fill[pattern=north east lines, pattern color=cOneSevenB] (axis cs:2.112,0) rectangle (axis cs:2.268,0.9000);
\draw[cTeacher, line width=.8pt, densely dashed] (axis cs:1.713,0.9000) -- (axis cs:2.268,0.9000);
\fill[cZeroSixB!30] (axis cs:2.732,0) rectangle (axis cs:2.888,0.1353);
\fill[cZeroSixB] (axis cs:2.922,0) rectangle (axis cs:3.078,0.6197);
\fill[cZeroSixB!30] (axis cs:3.112,0) rectangle (axis cs:3.268,0.7650);
\fill[pattern=north east lines, pattern color=cZeroSixB] (axis cs:3.112,0) rectangle (axis cs:3.268,0.7650);
\draw[cTeacher, line width=.8pt, densely dashed] (axis cs:2.712,0.7650) -- (axis cs:3.268,0.7650);
\end{groupplot}
\end{tikzpicture}
\caption{\textbf{After distillation, a single attempt never reaches what the student could reach in many
attempts before.} Lenient scoring on each benchmark; $p@n$ over every sample we drew ($n=48$ on AIME 2026 and AMC23, 32
on GSM8K-200). In every one of the 27 groups the distilled $p@1$ (dark solid bar) stays below the dashed line, the
starting student's $p@n$ (hatched). How $p@n$ itself changes: Figure~\ref{fig:coverage}. Values:
Table~\ref{tab:app-results}.}
\label{fig:ownceil}
\end{figure}

%% file: figures/fig_teachceil.tex
\begin{figure}[t]
\centering
\begin{tikzpicture}
\begin{groupplot}[group style={group size=3 by 3, horizontal sep=0.40cm, vertical sep=0.26cm,
    xticklabels at=edge bottom, yticklabels at=edge left, ylabels at=edge left, group name=teach},
  paperstyle, scale only axis, width=3.80cm, height=1.72cm, xmajorgrids=false,
  xmin=0.42, xmax=3.58, ymin=0, ymax=1.36, enlargelimits=false, clip=false,
  xtick={1,2,3}, xticklabels={4B,1.7B,0.6B}, xtick style={draw=none}, ytick={0,0.5,1},
  axis x line*=bottom, axis y line*=left, title style={yshift=-3pt},
  ylabel style={align=center, font=\scriptsize}, yticklabels={0,50,100}]
\nextgroupplot[title={\textbf{Non-thinking OPD}\\[-1pt]\scriptsize vs.\ the non-thinking teacher}, title style={align=center, yshift=-3pt}, ylabel={AIME 2026\\\% of teacher}]
\node[font=\tiny, text=cMuted, anchor=north west, inner sep=1pt] at (axis cs:0.45,1.36) {teacher: $p@1$ 0.163, $p@n$ 0.667};
\fill[cFourB] (axis cs:0.700,0) rectangle (axis cs:0.980,0.8677);
\node[font=\tiny, text=cTeacher, anchor=south, inner sep=1pt] at (axis cs:0.840,0.8677) {87};
\fill[cFourB] (axis cs:1.020,0) rectangle (axis cs:1.300,0.7000);
\fill[pattern=north east lines, pattern color=white] (axis cs:1.020,0) rectangle (axis cs:1.300,0.7000);
\node[font=\tiny, text=cTeacher, anchor=south, inner sep=1pt] at (axis cs:1.160,0.7000) {70};
\fill[cOneSevenB] (axis cs:1.700,0) rectangle (axis cs:1.980,0.4874);
\node[font=\tiny, text=cTeacher, anchor=south, inner sep=1pt] at (axis cs:1.840,0.4874) {49};
\fill[cOneSevenB] (axis cs:2.020,0) rectangle (axis cs:2.300,0.4500);
\fill[pattern=north east lines, pattern color=white] (axis cs:2.020,0) rectangle (axis cs:2.300,0.4500);
\node[font=\tiny, text=cTeacher, anchor=south, inner sep=1pt] at (axis cs:2.160,0.4500) {45};
\fill[cZeroSixB] (axis cs:2.700,0) rectangle (axis cs:2.980,0.1114);
\node[font=\tiny, text=cTeacher, anchor=south, inner sep=1pt] at (axis cs:2.840,0.1114) {11};
\fill[cZeroSixB] (axis cs:3.020,0) rectangle (axis cs:3.300,0.1999);
\fill[pattern=north east lines, pattern color=white] (axis cs:3.020,0) rectangle (axis cs:3.300,0.1999);
\node[font=\tiny, text=cTeacher, anchor=south, inner sep=1pt] at (axis cs:3.160,0.1999) {20};
\nextgroupplot[title={\textbf{Thinking OPD, external SFT}\\[-1pt]\scriptsize vs.\ the thinking teacher}, title style={align=center, yshift=-3pt}, legend style={at={(0.5,1)}, anchor=south, yshift=22pt, legend columns=2, /tikz/every even column/.append style={column sep=0.8em}}]
\addlegendimage{legend image code/.code={\fill[cMuted] (0cm,-0.8ex) rectangle (0.22cm,0.8ex);}}
\addlegendentry{distilled $p@1$ / teacher $p@1$}
\addlegendimage{legend image code/.code={\fill[cMuted] (0cm,-0.8ex) rectangle (0.22cm,0.8ex); \fill[pattern=north east lines, pattern color=white] (0cm,-0.8ex) rectangle (0.22cm,0.8ex);}}
\addlegendentry{distilled $p@n$ / teacher $p@n$ (100 = the teacher)}
\node[font=\tiny, text=cMuted, anchor=north west, inner sep=1pt] at (axis cs:0.45,1.36) {teacher: $p@1$ 0.674, $p@n$ 0.900};
\fill[cFourB] (axis cs:0.700,0) rectangle (axis cs:0.980,0.4882);
\node[font=\tiny, text=cTeacher, anchor=south, inner sep=1pt] at (axis cs:0.840,0.4882) {49};
\fill[cFourB] (axis cs:1.020,0) rectangle (axis cs:1.300,0.7778);
\fill[pattern=north east lines, pattern color=white] (axis cs:1.020,0) rectangle (axis cs:1.300,0.7778);
\node[font=\tiny, text=cTeacher, anchor=south, inner sep=1pt] at (axis cs:1.160,0.7778) {78};
\fill[cOneSevenB] (axis cs:1.700,0) rectangle (axis cs:1.980,0.0844);
\node[font=\tiny, text=cTeacher, anchor=south, inner sep=1pt] at (axis cs:1.840,0.0844) {8};
\fill[cOneSevenB] (axis cs:2.020,0) rectangle (axis cs:2.300,0.2222);
\fill[pattern=north east lines, pattern color=white] (axis cs:2.020,0) rectangle (axis cs:2.300,0.2222);
\node[font=\tiny, text=cTeacher, anchor=south, inner sep=1pt] at (axis cs:2.160,0.2222) {22};
\fill[cZeroSixB] (axis cs:2.700,0) rectangle (axis cs:2.980,0.0052);
\node[font=\tiny, text=cTeacher, anchor=south, inner sep=1pt] at (axis cs:2.840,0.0052) {1};
\fill[cZeroSixB] (axis cs:3.020,0) rectangle (axis cs:3.300,0.0741);
\fill[pattern=north east lines, pattern color=white] (axis cs:3.020,0) rectangle (axis cs:3.300,0.0741);
\node[font=\tiny, text=cTeacher, anchor=south, inner sep=1pt] at (axis cs:3.160,0.0741) {7};
\nextgroupplot[title={\textbf{Thinking OPD, teacher SFT}\\[-1pt]\scriptsize vs.\ the thinking teacher}, title style={align=center, yshift=-3pt}]
\node[font=\tiny, text=cMuted, anchor=north west, inner sep=1pt] at (axis cs:0.45,1.36) {teacher: $p@1$ 0.674, $p@n$ 0.900};
\fill[cFourB] (axis cs:0.700,0) rectangle (axis cs:0.980,0.4686);
\node[font=\tiny, text=cTeacher, anchor=south, inner sep=1pt] at (axis cs:0.840,0.4686) {47};
\fill[cFourB] (axis cs:1.020,0) rectangle (axis cs:1.300,0.8148);
\fill[pattern=north east lines, pattern color=white] (axis cs:1.020,0) rectangle (axis cs:1.300,0.8148);
\node[font=\tiny, text=cTeacher, anchor=south, inner sep=1pt] at (axis cs:1.160,0.8148) {81};
\fill[cOneSevenB] (axis cs:1.700,0) rectangle (axis cs:1.980,0.1000);
\node[font=\tiny, text=cTeacher, anchor=south, inner sep=1pt] at (axis cs:1.840,0.1000) {10};
\fill[cOneSevenB] (axis cs:2.020,0) rectangle (axis cs:2.300,0.2592);
\fill[pattern=north east lines, pattern color=white] (axis cs:2.020,0) rectangle (axis cs:2.300,0.2592);
\node[font=\tiny, text=cTeacher, anchor=south, inner sep=1pt] at (axis cs:2.160,0.2592) {26};
\fill[cZeroSixB] (axis cs:2.700,0) rectangle (axis cs:2.980,0.0042);
\node[font=\tiny, text=cTeacher, anchor=south, inner sep=1pt] at (axis cs:2.840,0.0042) {0};
\fill[cZeroSixB] (axis cs:3.020,0) rectangle (axis cs:3.300,0.0741);
\fill[pattern=north east lines, pattern color=white] (axis cs:3.020,0) rectangle (axis cs:3.300,0.0741);
\node[font=\tiny, text=cTeacher, anchor=south, inner sep=1pt] at (axis cs:3.160,0.0741) {7};
\nextgroupplot[ylabel={AMC23\\\% of teacher}]
\node[font=\tiny, text=cMuted, anchor=north west, inner sep=1pt] at (axis cs:0.45,1.36) {teacher: $p@1$ 0.694, $p@n$ 0.975};
\fill[cFourB] (axis cs:0.700,0) rectangle (axis cs:0.980,0.8455);
\node[font=\tiny, text=cTeacher, anchor=south, inner sep=1pt] at (axis cs:0.840,0.8455) {85};
\fill[cFourB] (axis cs:1.020,0) rectangle (axis cs:1.300,0.9487);
\fill[pattern=north east lines, pattern color=white] (axis cs:1.020,0) rectangle (axis cs:1.300,0.9487);
\node[font=\tiny, text=cTeacher, anchor=south, inner sep=1pt] at (axis cs:1.160,0.9487) {95};
\fill[cOneSevenB] (axis cs:1.700,0) rectangle (axis cs:1.980,0.5909);
\node[font=\tiny, text=cTeacher, anchor=south, inner sep=1pt] at (axis cs:1.840,0.5909) {59};
\fill[cOneSevenB] (axis cs:2.020,0) rectangle (axis cs:2.300,0.9231);
\fill[pattern=north east lines, pattern color=white] (axis cs:2.020,0) rectangle (axis cs:2.300,0.9231);
\node[font=\tiny, text=cTeacher, anchor=south, inner sep=1pt] at (axis cs:2.160,0.9231) {92};
\fill[cZeroSixB] (axis cs:2.700,0) rectangle (axis cs:2.980,0.4002);
\node[font=\tiny, text=cTeacher, anchor=south, inner sep=1pt] at (axis cs:2.840,0.4002) {40};
\fill[cZeroSixB] (axis cs:3.020,0) rectangle (axis cs:3.300,0.7179);
\fill[pattern=north east lines, pattern color=white] (axis cs:3.020,0) rectangle (axis cs:3.300,0.7179);
\node[font=\tiny, text=cTeacher, anchor=south, inner sep=1pt] at (axis cs:3.160,0.7179) {72};
\nextgroupplot[]
\node[font=\tiny, text=cMuted, anchor=north west, inner sep=1pt] at (axis cs:0.45,1.36) {teacher: $p@1$ 0.960, $p@n$ 1.000};
\fill[cFourB] (axis cs:0.700,0) rectangle (axis cs:0.980,0.7580);
\node[font=\tiny, text=cTeacher, anchor=south, inner sep=1pt] at (axis cs:0.840,0.7580) {76};
\fill[cFourB] (axis cs:1.020,0) rectangle (axis cs:1.300,0.9000);
\fill[pattern=north east lines, pattern color=white] (axis cs:1.020,0) rectangle (axis cs:1.300,0.9000);
\node[font=\tiny, text=cTeacher, anchor=south, inner sep=1pt] at (axis cs:1.160,0.9000) {90};
\fill[cOneSevenB] (axis cs:1.700,0) rectangle (axis cs:1.980,0.3776);
\node[font=\tiny, text=cTeacher, anchor=south, inner sep=1pt] at (axis cs:1.840,0.3776) {38};
\fill[cOneSevenB] (axis cs:2.020,0) rectangle (axis cs:2.300,0.7250);
\fill[pattern=north east lines, pattern color=white] (axis cs:2.020,0) rectangle (axis cs:2.300,0.7250);
\node[font=\tiny, text=cTeacher, anchor=south, inner sep=1pt] at (axis cs:2.160,0.7250) {72};
\fill[cZeroSixB] (axis cs:2.700,0) rectangle (axis cs:2.980,0.2209);
\node[font=\tiny, text=cTeacher, anchor=south, inner sep=1pt] at (axis cs:2.840,0.2209) {22};
\fill[cZeroSixB] (axis cs:3.020,0) rectangle (axis cs:3.300,0.6000);
\fill[pattern=north east lines, pattern color=white] (axis cs:3.020,0) rectangle (axis cs:3.300,0.6000);
\node[font=\tiny, text=cTeacher, anchor=south, inner sep=1pt] at (axis cs:3.160,0.6000) {60};
\nextgroupplot[]
\node[font=\tiny, text=cMuted, anchor=north west, inner sep=1pt] at (axis cs:0.45,1.36) {teacher: $p@1$ 0.960, $p@n$ 1.000};
\fill[cFourB] (axis cs:0.700,0) rectangle (axis cs:0.980,0.7585);
\node[font=\tiny, text=cTeacher, anchor=south, inner sep=1pt] at (axis cs:0.840,0.7585) {76};
\fill[cFourB] (axis cs:1.020,0) rectangle (axis cs:1.300,0.9500);
\fill[pattern=north east lines, pattern color=white] (axis cs:1.020,0) rectangle (axis cs:1.300,0.9500);
\node[font=\tiny, text=cTeacher, anchor=south, inner sep=1pt] at (axis cs:1.160,0.9500) {95};
\fill[cOneSevenB] (axis cs:1.700,0) rectangle (axis cs:1.980,0.3874);
\node[font=\tiny, text=cTeacher, anchor=south, inner sep=1pt] at (axis cs:1.840,0.3874) {39};
\fill[cOneSevenB] (axis cs:2.020,0) rectangle (axis cs:2.300,0.6750);
\fill[pattern=north east lines, pattern color=white] (axis cs:2.020,0) rectangle (axis cs:2.300,0.6750);
\node[font=\tiny, text=cTeacher, anchor=south, inner sep=1pt] at (axis cs:2.160,0.6750) {68};
\fill[cZeroSixB] (axis cs:2.700,0) rectangle (axis cs:2.980,0.1715);
\node[font=\tiny, text=cTeacher, anchor=south, inner sep=1pt] at (axis cs:2.840,0.1715) {17};
\fill[cZeroSixB] (axis cs:3.020,0) rectangle (axis cs:3.300,0.4750);
\fill[pattern=north east lines, pattern color=white] (axis cs:3.020,0) rectangle (axis cs:3.300,0.4750);
\node[font=\tiny, text=cTeacher, anchor=south, inner sep=1pt] at (axis cs:3.160,0.4750) {48};
\nextgroupplot[ylabel={GSM8K-200\\\% of teacher}]
\node[font=\tiny, text=cMuted, anchor=north west, inner sep=1pt] at (axis cs:0.45,1.36) {teacher: $p@1$ 0.929, $p@n$ 0.965};
\fill[cFourB] (axis cs:0.700,0) rectangle (axis cs:0.980,0.9888);
\node[font=\tiny, text=cTeacher, anchor=south, inner sep=1pt] at (axis cs:0.840,0.9888) {99};
\fill[cFourB] (axis cs:1.020,0) rectangle (axis cs:1.300,1.0155);
\fill[pattern=north east lines, pattern color=white] (axis cs:1.020,0) rectangle (axis cs:1.300,1.0155);
\node[font=\tiny, text=cTeacher, anchor=south, inner sep=1pt] at (axis cs:1.160,1.0155) {102};
\fill[cOneSevenB] (axis cs:1.700,0) rectangle (axis cs:1.980,0.8930);
\node[font=\tiny, text=cTeacher, anchor=south, inner sep=1pt] at (axis cs:1.840,0.8930) {89};
\fill[cOneSevenB] (axis cs:2.020,0) rectangle (axis cs:2.300,1.0104);
\fill[pattern=north east lines, pattern color=white] (axis cs:2.020,0) rectangle (axis cs:2.300,1.0104);
\node[font=\tiny, text=cTeacher, anchor=south, inner sep=1pt] at (axis cs:2.160,1.0104) {101};
\fill[cZeroSixB] (axis cs:2.700,0) rectangle (axis cs:2.980,0.7413);
\node[font=\tiny, text=cTeacher, anchor=south, inner sep=1pt] at (axis cs:2.840,0.7413) {74};
\fill[cZeroSixB] (axis cs:3.020,0) rectangle (axis cs:3.300,0.9793);
\fill[pattern=north east lines, pattern color=white] (axis cs:3.020,0) rectangle (axis cs:3.300,0.9793);
\node[font=\tiny, text=cTeacher, anchor=south, inner sep=1pt] at (axis cs:3.160,0.9793) {98};
\nextgroupplot[]
\node[font=\tiny, text=cMuted, anchor=north west, inner sep=1pt] at (axis cs:0.45,1.36) {teacher: $p@1$ 0.937, $p@n$ 0.965};
\fill[cFourB] (axis cs:0.700,0) rectangle (axis cs:0.980,0.9565);
\node[font=\tiny, text=cTeacher, anchor=south, inner sep=1pt] at (axis cs:0.840,0.9565) {96};
\fill[cFourB] (axis cs:1.020,0) rectangle (axis cs:1.300,1.0000);
\fill[pattern=north east lines, pattern color=white] (axis cs:1.020,0) rectangle (axis cs:1.300,1.0000);
\node[font=\tiny, text=cTeacher, anchor=south, inner sep=1pt] at (axis cs:1.160,1.0000) {100};
\fill[cOneSevenB] (axis cs:1.700,0) rectangle (axis cs:1.980,0.8079);
\node[font=\tiny, text=cTeacher, anchor=south, inner sep=1pt] at (axis cs:1.840,0.8079) {81};
\fill[cOneSevenB] (axis cs:2.020,0) rectangle (axis cs:2.300,0.9948);
\fill[pattern=north east lines, pattern color=white] (axis cs:2.020,0) rectangle (axis cs:2.300,0.9948);
\node[font=\tiny, text=cTeacher, anchor=south, inner sep=1pt] at (axis cs:2.160,0.9948) {99};
\fill[cZeroSixB] (axis cs:2.700,0) rectangle (axis cs:2.980,0.7116);
\node[font=\tiny, text=cTeacher, anchor=south, inner sep=1pt] at (axis cs:2.840,0.7116) {71};
\fill[cZeroSixB] (axis cs:3.020,0) rectangle (axis cs:3.300,0.9275);
\fill[pattern=north east lines, pattern color=white] (axis cs:3.020,0) rectangle (axis cs:3.300,0.9275);
\node[font=\tiny, text=cTeacher, anchor=south, inner sep=1pt] at (axis cs:3.160,0.9275) {93};
\nextgroupplot[]
\node[font=\tiny, text=cMuted, anchor=north west, inner sep=1pt] at (axis cs:0.45,1.36) {teacher: $p@1$ 0.937, $p@n$ 0.965};
\fill[cFourB] (axis cs:0.700,0) rectangle (axis cs:0.980,0.9635);
\node[font=\tiny, text=cTeacher, anchor=south, inner sep=1pt] at (axis cs:0.840,0.9635) {96};
\fill[cFourB] (axis cs:1.020,0) rectangle (axis cs:1.300,1.0000);
\fill[pattern=north east lines, pattern color=white] (axis cs:1.020,0) rectangle (axis cs:1.300,1.0000);
\node[font=\tiny, text=cTeacher, anchor=south, inner sep=1pt] at (axis cs:1.160,1.0000) {100};
\fill[cOneSevenB] (axis cs:1.700,0) rectangle (axis cs:1.980,0.7928);
\node[font=\tiny, text=cTeacher, anchor=south, inner sep=1pt] at (axis cs:1.840,0.7928) {79};
\fill[cOneSevenB] (axis cs:2.020,0) rectangle (axis cs:2.300,0.9845);
\fill[pattern=north east lines, pattern color=white] (axis cs:2.020,0) rectangle (axis cs:2.300,0.9845);
\node[font=\tiny, text=cTeacher, anchor=south, inner sep=1pt] at (axis cs:2.160,0.9845) {98};
\fill[cZeroSixB] (axis cs:2.700,0) rectangle (axis cs:2.980,0.6612);
\node[font=\tiny, text=cTeacher, anchor=south, inner sep=1pt] at (axis cs:2.840,0.6612) {66};
\fill[cZeroSixB] (axis cs:3.020,0) rectangle (axis cs:3.300,0.9119);
\fill[pattern=north east lines, pattern color=white] (axis cs:3.020,0) rectangle (axis cs:3.300,0.9119);
\node[font=\tiny, text=cTeacher, anchor=south, inner sep=1pt] at (axis cs:3.160,0.9119) {91};
\end{groupplot}
\end{tikzpicture}
\caption{\textbf{The smaller the student, the farther it stays below the teacher.} The distilled student's
lenient $p@1$ and $p@n$ on each benchmark as a percentage of the teacher's in the student's own mode (100: the
teacher, whose own values are printed in each panel). On GSM8K-200 the distilled $p@n$ comes close to the teacher's at every
size; on AIME 2026 the 1.7B and 0.6B in thinking mode reach a tenth of the teacher's $p@1$ or less.}
\label{fig:teachceil}
\end{figure}
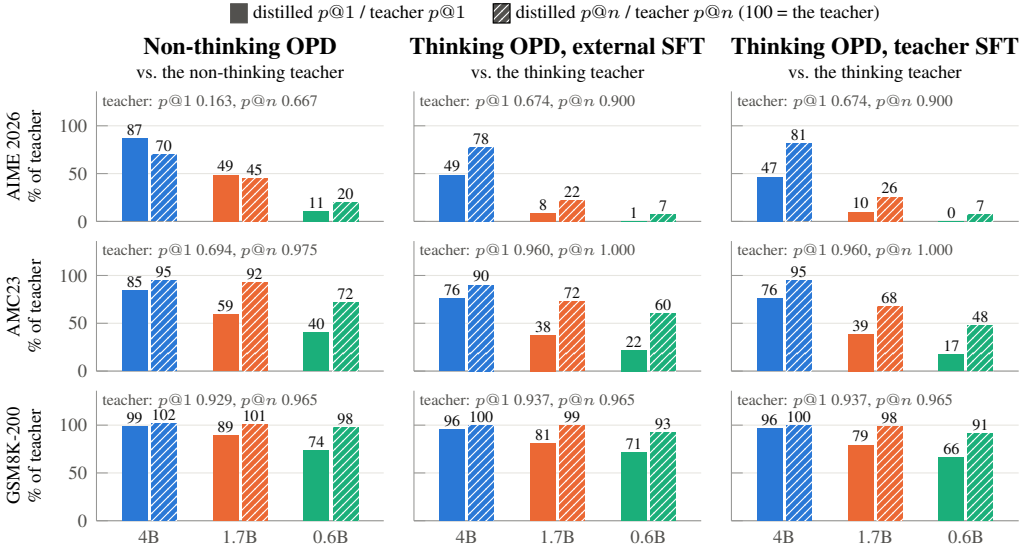

%% file: tables/tab_results.tex
\begin{table}[t]
\centering
\caption{\textbf{Main results, three-benchmark mean.} Lenient scoring; $p@n$ over every sample drawn ($n=48$
on AIME 2026 and AMC23, 32 on GSM8K-200). The distilled $p@1$ stays below the start's $p@n$ in every row. From the
teacher-SFT start, the 1.7B and 0.6B mark many answers per sample; Appendix~\ref{app:scoring} gives their scores without this.
Per benchmark: Table~\ref{tab:app-results}.}
\label{tab:results}
\footnotesize
\setlength{\tabcolsep}{5pt}
\renewcommand{\arraystretch}{0.95}
\begin{tabular}{@{}lcccccc@{}}
\toprule
 & \multicolumn{2}{c}{Start} & \multicolumn{2}{c}{After distillation} & \multicolumn{2}{c}{After $p@1$ as a share of} \\
\cmidrule(lr){2-3}\cmidrule(lr){4-5}\cmidrule(l){6-7}
Student & $p@1$ & $p@n$ & $p@1$ & $p@n$ & start $p@n$ & teacher $p@1$ \\
\midrule
\multicolumn{7}{@{}l}{\textit{Non-thinking OPD}\enspace\textcolor{cMuted}{(teacher: $p@1$ 0.595, $p@n$ 0.869)}} \\
\quad \sizedot{cFourB}\,4B & \textcolor{cMuted}{0.427} & \textcolor{cMuted}{0.736} & 0.549 & 0.791 & 0.75 & 0.922 \\
\quad \sizedot{cOneSevenB}\,1.7B & \textcolor{cMuted}{0.294} & \textcolor{cMuted}{0.646} & 0.440 & 0.725 & 0.68 & 0.739 \\
\quad \sizedot{cZeroSixB}\,0.6B & \textcolor{cMuted}{0.048} & \textcolor{cMuted}{0.512} & 0.328 & 0.593 & 0.64 & 0.551 \\
\addlinespace[2pt]
\multicolumn{7}{@{}l}{\textit{Thinking OPD, external SFT}\enspace\textcolor{cMuted}{(teacher: $p@1$ 0.857, $p@n$ 0.955)}} \\
\quad \sizedot{cFourB}\,4B & \textcolor{cMuted}{0.532} & \textcolor{cMuted}{0.768} & 0.651 & 0.855 & 0.85 & 0.760 \\
\quad \sizedot{cOneSevenB}\,1.7B & \textcolor{cMuted}{0.358} & \textcolor{cMuted}{0.664} & 0.392 & 0.628 & 0.59 & 0.458 \\
\quad \sizedot{cZeroSixB}\,0.6B & \textcolor{cMuted}{0.238} & \textcolor{cMuted}{0.591} & 0.294 & 0.521 & 0.50 & 0.343 \\
\addlinespace[2pt]
\multicolumn{7}{@{}l}{\textit{Thinking OPD, teacher SFT}\enspace\textcolor{cMuted}{(teacher: $p@1$ 0.857, $p@n$ 0.955)}} \\
\quad \sizedot{cFourB}\,4B & \textcolor{cMuted}{0.330} & \textcolor{cMuted}{0.748} & 0.649 & 0.883 & 0.87 & 0.757 \\
\quad \sizedot{cOneSevenB}\,1.7B & \textcolor{cMuted}{0.128} & \textcolor{cMuted}{0.517} & 0.394 & 0.619 & 0.76 & 0.460 \\
\quad \sizedot{cZeroSixB}\,0.6B & \textcolor{cMuted}{0.055} & \textcolor{cMuted}{0.399} & 0.262 & 0.474 & 0.66 & 0.306 \\
\bottomrule
\end{tabular}
\end{table}

%% file: figures/fig_coverage.tex
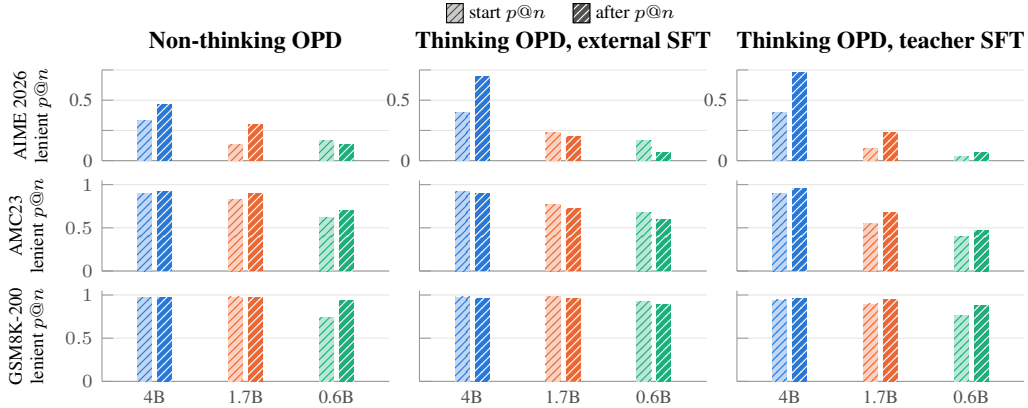
\begin{figure}[t]
\centering
\begin{tikzpicture}
\begin{groupplot}[group style={group size=3 by 3, horizontal sep=0.40cm, vertical sep=0.26cm,
    xticklabels at=edge bottom, yticklabels at=edge left, ylabels at=edge left, group name=cov},
  paperstyle, scale only axis, width=3.80cm, height=1.20cm, xmajorgrids=false,
  xmin=0.42, xmax=3.58, ymin=0, ymax=1.05, enlargelimits=false, clip=false,
  xtick={1,2,3}, xticklabels={4B,1.7B,0.6B}, xtick style={draw=none}, ytick={0,0.5,1},
  axis x line*=bottom, axis y line*=left, title style={yshift=-3pt},
  ylabel style={align=center, font=\scriptsize}]
\nextgroupplot[title={\textbf{Non-thinking OPD}}, ylabel={AIME 2026\\lenient $p@n$}, ymax=0.75, ytick={0,0.25,0.5,0.75}, yticklabels={0,,0.5,}]
\fill[cFourB!30] (axis cs:0.812,0) rectangle (axis cs:0.968,0.3333);
\fill[pattern=north east lines, pattern color=cFourB] (axis cs:0.812,0) rectangle (axis cs:0.968,0.3333);
\fill[cFourB] (axis cs:1.033,0) rectangle (axis cs:1.188,0.4667);
\fill[pattern=north east lines, pattern color=white] (axis cs:1.033,0) rectangle (axis cs:1.188,0.4667);
\fill[cOneSevenB!30] (axis cs:1.812,0) rectangle (axis cs:1.967,0.1333);
\fill[pattern=north east lines, pattern color=cOneSevenB] (axis cs:1.812,0) rectangle (axis cs:1.967,0.1333);
\fill[cOneSevenB] (axis cs:2.032,0) rectangle (axis cs:2.188,0.3000);
\fill[pattern=north east lines, pattern color=white] (axis cs:2.032,0) rectangle (axis cs:2.188,0.3000);
\fill[cZeroSixB!30] (axis cs:2.812,0) rectangle (axis cs:2.968,0.1667);
\fill[pattern=north east lines, pattern color=cZeroSixB] (axis cs:2.812,0) rectangle (axis cs:2.968,0.1667);
\fill[cZeroSixB] (axis cs:3.032,0) rectangle (axis cs:3.188,0.1333);
\fill[pattern=north east lines, pattern color=white] (axis cs:3.032,0) rectangle (axis cs:3.188,0.1333);
\nextgroupplot[title={\textbf{Thinking OPD, external SFT}}, ymax=0.75, ytick={0,0.25,0.5,0.75}, yticklabels={0,,0.5,}, legend style={at={(0.5,1)}, anchor=south, yshift=14pt, legend columns=2, /tikz/every even column/.append style={column sep=0.8em}}]
\addlegendimage{legend image code/.code={\fill[cMuted!30] (0cm,-0.8ex) rectangle (0.22cm,0.8ex); \fill[pattern=north east lines, pattern color=cMuted] (0cm,-0.8ex) rectangle (0.22cm,0.8ex);}}
\addlegendentry{start $p@n$}
\addlegendimage{legend image code/.code={\fill[cMuted] (0cm,-0.8ex) rectangle (0.22cm,0.8ex); \fill[pattern=north east lines, pattern color=white] (0cm,-0.8ex) rectangle (0.22cm,0.8ex);}}
\addlegendentry{after $p@n$}
\fill[cFourB!30] (axis cs:0.812,0) rectangle (axis cs:0.968,0.4000);
\fill[pattern=north east lines, pattern color=cFourB] (axis cs:0.812,0) rectangle (axis cs:0.968,0.4000);
\fill[cFourB] (axis cs:1.033,0) rectangle (axis cs:1.188,0.7000);
\fill[pattern=north east lines, pattern color=white] (axis cs:1.033,0) rectangle (axis cs:1.188,0.7000);
\fill[cOneSevenB!30] (axis cs:1.812,0) rectangle (axis cs:1.967,0.2333);
\fill[pattern=north east lines, pattern color=cOneSevenB] (axis cs:1.812,0) rectangle (axis cs:1.967,0.2333);
\fill[cOneSevenB] (axis cs:2.032,0) rectangle (axis cs:2.188,0.2000);
\fill[pattern=north east lines, pattern color=white] (axis cs:2.032,0) rectangle (axis cs:2.188,0.2000);
\fill[cZeroSixB!30] (axis cs:2.812,0) rectangle (axis cs:2.968,0.1667);
\fill[pattern=north east lines, pattern color=cZeroSixB] (axis cs:2.812,0) rectangle (axis cs:2.968,0.1667);
\fill[cZeroSixB] (axis cs:3.032,0) rectangle (axis cs:3.188,0.0667);
\fill[pattern=north east lines, pattern color=white] (axis cs:3.032,0) rectangle (axis cs:3.188,0.0667);
\nextgroupplot[title={\textbf{Thinking OPD, teacher SFT}}, ymax=0.75, ytick={0,0.25,0.5,0.75}, yticklabels={0,,0.5,}]
\fill[cFourB!30] (axis cs:0.812,0) rectangle (axis cs:0.968,0.4000);
\fill[pattern=north east lines, pattern color=cFourB] (axis cs:0.812,0) rectangle (axis cs:0.968,0.4000);
\fill[cFourB] (axis cs:1.033,0) rectangle (axis cs:1.188,0.7333);
\fill[pattern=north east lines, pattern color=white] (axis cs:1.033,0) rectangle (axis cs:1.188,0.7333);
\fill[cOneSevenB!30] (axis cs:1.812,0) rectangle (axis cs:1.967,0.1000);
\fill[pattern=north east lines, pattern color=cOneSevenB] (axis cs:1.812,0) rectangle (axis cs:1.967,0.1000);
\fill[cOneSevenB] (axis cs:2.032,0) rectangle (axis cs:2.188,0.2333);
\fill[pattern=north east lines, pattern color=white] (axis cs:2.032,0) rectangle (axis cs:2.188,0.2333);
\fill[cZeroSixB!30] (axis cs:2.812,0) rectangle (axis cs:2.968,0.0333);
\fill[pattern=north east lines, pattern color=cZeroSixB] (axis cs:2.812,0) rectangle (axis cs:2.968,0.0333);
\fill[cZeroSixB] (axis cs:3.032,0) rectangle (axis cs:3.188,0.0667);
\fill[pattern=north east lines, pattern color=white] (axis cs:3.032,0) rectangle (axis cs:3.188,0.0667);
\nextgroupplot[ylabel={AMC23\\lenient $p@n$}]
\fill[cFourB!30] (axis cs:0.812,0) rectangle (axis cs:0.968,0.9000);
\fill[pattern=north east lines, pattern color=cFourB] (axis cs:0.812,0) rectangle (axis cs:0.968,0.9000);
\fill[cFourB] (axis cs:1.033,0) rectangle (axis cs:1.188,0.9250);
\fill[pattern=north east lines, pattern color=white] (axis cs:1.033,0) rectangle (axis cs:1.188,0.9250);
\fill[cOneSevenB!30] (axis cs:1.812,0) rectangle (axis cs:1.967,0.8250);
\fill[pattern=north east lines, pattern color=cOneSevenB] (axis cs:1.812,0) rectangle (axis cs:1.967,0.8250);
\fill[cOneSevenB] (axis cs:2.032,0) rectangle (axis cs:2.188,0.9000);
\fill[pattern=north east lines, pattern color=white] (axis cs:2.032,0) rectangle (axis cs:2.188,0.9000);
\fill[cZeroSixB!30] (axis cs:2.812,0) rectangle (axis cs:2.968,0.6250);
\fill[pattern=north east lines, pattern color=cZeroSixB] (axis cs:2.812,0) rectangle (axis cs:2.968,0.6250);
\fill[cZeroSixB] (axis cs:3.032,0) rectangle (axis cs:3.188,0.7000);
\fill[pattern=north east lines, pattern color=white] (axis cs:3.032,0) rectangle (axis cs:3.188,0.7000);
\nextgroupplot[]
\fill[cFourB!30] (axis cs:0.812,0) rectangle (axis cs:0.968,0.9250);
\fill[pattern=north east lines, pattern color=cFourB] (axis cs:0.812,0) rectangle (axis cs:0.968,0.9250);
\fill[cFourB] (axis cs:1.033,0) rectangle (axis cs:1.188,0.9000);
\fill[pattern=north east lines, pattern color=white] (axis cs:1.033,0) rectangle (axis cs:1.188,0.9000);
\fill[cOneSevenB!30] (axis cs:1.812,0) rectangle (axis cs:1.967,0.7750);
\fill[pattern=north east lines, pattern color=cOneSevenB] (axis cs:1.812,0) rectangle (axis cs:1.967,0.7750);
\fill[cOneSevenB] (axis cs:2.032,0) rectangle (axis cs:2.188,0.7250);
\fill[pattern=north east lines, pattern color=white] (axis cs:2.032,0) rectangle (axis cs:2.188,0.7250);
\fill[cZeroSixB!30] (axis cs:2.812,0) rectangle (axis cs:2.968,0.6750);
\fill[pattern=north east lines, pattern color=cZeroSixB] (axis cs:2.812,0) rectangle (axis cs:2.968,0.6750);
\fill[cZeroSixB] (axis cs:3.032,0) rectangle (axis cs:3.188,0.6000);
\fill[pattern=north east lines, pattern color=white] (axis cs:3.032,0) rectangle (axis cs:3.188,0.6000);
\nextgroupplot[]
\fill[cFourB!30] (axis cs:0.812,0) rectangle (axis cs:0.968,0.9000);
\fill[pattern=north east lines, pattern color=cFourB] (axis cs:0.812,0) rectangle (axis cs:0.968,0.9000);
\fill[cFourB] (axis cs:1.033,0) rectangle (axis cs:1.188,0.9500);
\fill[pattern=north east lines, pattern color=white] (axis cs:1.033,0) rectangle (axis cs:1.188,0.9500);
\fill[cOneSevenB!30] (axis cs:1.812,0) rectangle (axis cs:1.967,0.5500);
\fill[pattern=north east lines, pattern color=cOneSevenB] (axis cs:1.812,0) rectangle (axis cs:1.967,0.5500);
\fill[cOneSevenB] (axis cs:2.032,0) rectangle (axis cs:2.188,0.6750);
\fill[pattern=north east lines, pattern color=white] (axis cs:2.032,0) rectangle (axis cs:2.188,0.6750);
\fill[cZeroSixB!30] (axis cs:2.812,0) rectangle (axis cs:2.968,0.4000);
\fill[pattern=north east lines, pattern color=cZeroSixB] (axis cs:2.812,0) rectangle (axis cs:2.968,0.4000);
\fill[cZeroSixB] (axis cs:3.032,0) rectangle (axis cs:3.188,0.4750);
\fill[pattern=north east lines, pattern color=white] (axis cs:3.032,0) rectangle (axis cs:3.188,0.4750);
\nextgroupplot[ylabel={GSM8K-200\\lenient $p@n$}]
\fill[cFourB!30] (axis cs:0.812,0) rectangle (axis cs:0.968,0.9750);
\fill[pattern=north east lines, pattern color=cFourB] (axis cs:0.812,0) rectangle (axis cs:0.968,0.9750);
\fill[cFourB] (axis cs:1.033,0) rectangle (axis cs:1.188,0.9800);
\fill[pattern=north east lines, pattern color=white] (axis cs:1.033,0) rectangle (axis cs:1.188,0.9800);
\fill[cOneSevenB!30] (axis cs:1.812,0) rectangle (axis cs:1.967,0.9800);
\fill[pattern=north east lines, pattern color=cOneSevenB] (axis cs:1.812,0) rectangle (axis cs:1.967,0.9800);
\fill[cOneSevenB] (axis cs:2.032,0) rectangle (axis cs:2.188,0.9750);
\fill[pattern=north east lines, pattern color=white] (axis cs:2.032,0) rectangle (axis cs:2.188,0.9750);
\fill[cZeroSixB!30] (axis cs:2.812,0) rectangle (axis cs:2.968,0.7450);
\fill[pattern=north east lines, pattern color=cZeroSixB] (axis cs:2.812,0) rectangle (axis cs:2.968,0.7450);
\fill[cZeroSixB] (axis cs:3.032,0) rectangle (axis cs:3.188,0.9450);
\fill[pattern=north east lines, pattern color=white] (axis cs:3.032,0) rectangle (axis cs:3.188,0.9450);
\nextgroupplot[]
\fill[cFourB!30] (axis cs:0.812,0) rectangle (axis cs:0.968,0.9800);
\fill[pattern=north east lines, pattern color=cFourB] (axis cs:0.812,0) rectangle (axis cs:0.968,0.9800);
\fill[cFourB] (axis cs:1.033,0) rectangle (axis cs:1.188,0.9650);
\fill[pattern=north east lines, pattern color=white] (axis cs:1.033,0) rectangle (axis cs:1.188,0.9650);
\fill[cOneSevenB!30] (axis cs:1.812,0) rectangle (axis cs:1.967,0.9850);
\fill[pattern=north east lines, pattern color=cOneSevenB] (axis cs:1.812,0) rectangle (axis cs:1.967,0.9850);
\fill[cOneSevenB] (axis cs:2.032,0) rectangle (axis cs:2.188,0.9600);
\fill[pattern=north east lines, pattern color=white] (axis cs:2.032,0) rectangle (axis cs:2.188,0.9600);
\fill[cZeroSixB!30] (axis cs:2.812,0) rectangle (axis cs:2.968,0.9300);
\fill[pattern=north east lines, pattern color=cZeroSixB] (axis cs:2.812,0) rectangle (axis cs:2.968,0.9300);
\fill[cZeroSixB] (axis cs:3.032,0) rectangle (axis cs:3.188,0.8950);
\fill[pattern=north east lines, pattern color=white] (axis cs:3.032,0) rectangle (axis cs:3.188,0.8950);
\nextgroupplot[]
\fill[cFourB!30] (axis cs:0.812,0) rectangle (axis cs:0.968,0.9450);
\fill[pattern=north east lines, pattern color=cFourB] (axis cs:0.812,0) rectangle (axis cs:0.968,0.9450);
\fill[cFourB] (axis cs:1.033,0) rectangle (axis cs:1.188,0.9650);
\fill[pattern=north east lines, pattern color=white] (axis cs:1.033,0) rectangle (axis cs:1.188,0.9650);
\fill[cOneSevenB!30] (axis cs:1.812,0) rectangle (axis cs:1.967,0.9000);
\fill[pattern=north east lines, pattern color=cOneSevenB] (axis cs:1.812,0) rectangle (axis cs:1.967,0.9000);
\fill[cOneSevenB] (axis cs:2.032,0) rectangle (axis cs:2.188,0.9500);
\fill[pattern=north east lines, pattern color=white] (axis cs:2.032,0) rectangle (axis cs:2.188,0.9500);
\fill[cZeroSixB!30] (axis cs:2.812,0) rectangle (axis cs:2.968,0.7650);
\fill[pattern=north east lines, pattern color=cZeroSixB] (axis cs:2.812,0) rectangle (axis cs:2.968,0.7650);
\fill[cZeroSixB] (axis cs:3.032,0) rectangle (axis cs:3.188,0.8800);
\fill[pattern=north east lines, pattern color=white] (axis cs:3.032,0) rectangle (axis cs:3.188,0.8800);
\end{groupplot}
\end{tikzpicture}
\caption{\textbf{What a student can reach at all barely changes.} Lenient $p@n$ on each benchmark before and
after distillation, over every sample drawn. It grows clearly for the 4B in thinking mode on AIME 2026 and for the
pretrained-only 0.6B on GSM8K-200.}
\label{fig:coverage}
\end{figure}

%% file: sections/04_stopping.tex
\section{Why small students do not learn to stop}
\label{sec:stopping}

In thinking mode, distillation filters a student's stops rather than teaching them: it keeps the stops that land on a
correct answer and removes the others, so a student whose stops were rarely right is left almost unable to stop. In every run from the external-SFT start at the 7,168-token budget, the fraction of
training responses that close their reasoning drops sharply at step 6 (Figure~\ref{fig:closure}), in the same step in
which the median response reaches the rollout budget (Appendix Figure~\ref{fig:collapse}): between steps 1--5 and steps
6--10 it falls from 0.617 to 0.211 at 4B, from 0.445 to 0.128 at 1.7B and from 0.583 to 0.113 at 0.6B. Nothing we
varied within thinking-mode training prevents this. Two runs of the same configuration at 4B and at 0.6B collapse at the same
step; the 0.6B collapses under budgets of 7,168 and 16,384 tokens, filling the larger one a step later, and at 1,024
tokens fills the budget from the first step; the students from the teacher-SFT start fill the budget from the first step; and the officially
post-trained Qwen3-0.6B, a much stronger starting point, also drops at step 6 (Appendix~\ref{app:block2}).

\input{figures/fig_closure}

Nothing brings the stops back because of where the teacher asks for one. We had the teacher score the students' first training
responses (Figure~\ref{fig:probe}; numbers in Appendix Table~\ref{tab:probe}). Where a student closed its reasoning, the teacher agrees: its probability of \think{}
at that point is 0.976 at 4B, 0.764 at 1.7B and 0.673 at 0.6B. Along a typical response that ran to the budget, the
teacher's largest probability of stopping at any point is at most about $10^{-7}$: along most such responses the
teacher does not point to a better place to stop; it points nowhere. A stop is therefore reinforced where the student already stops, and
nothing pulls a response that runs on toward a stop.

\input{figures/fig_probe}

\input{figures/fig_filter}

What survives the filter is the correct stops (Figure~\ref{fig:filter}), so the collapse is not the budget simply cutting responses short.
Stretching every response of the officially post-trained 0.6B before distillation by one common factor, chosen to match the observed closure, predicts
that correct stops fall from 0.305 to 0.229, but they stay at 0.311, while most of the wrong stops disappear. The stop
token itself is unchanged: on identical prefixes, the officially post-trained 0.6B ranks \think{} first wherever it had
closed, both before and after 20 steps of distillation, and tens of thousands of ranks deep everywhere else, again in
both (Appendix~\ref{app:probes}). From the external-SFT start, what the filter leaves therefore depends on how many of a
student's stops were right to begin with. Before training the students close their reasoning at rates that do not
follow their size (0.648, 0.459 and 0.611 from 4B to 0.6B), but their \emph{stop reliability}, the fraction of stops that land on a correct answer, is 0.521 at 4B, 0.128
at 1.7B and 0.042 at 0.6B, against 0.885 for the teacher (Figure~\ref{fig:filter}a). Late in training they keep their closure in the same order,
0.226, 0.064 and 0.023, and the stops that remain are more often right where the student can solve: the accuracy of
closed training responses rises at 4B from about one half to at most 0.698 within 30 steps, reaches 0.256 at 1.7B and
stays near zero at 0.6B.

%% file: figures/fig_closure.tex
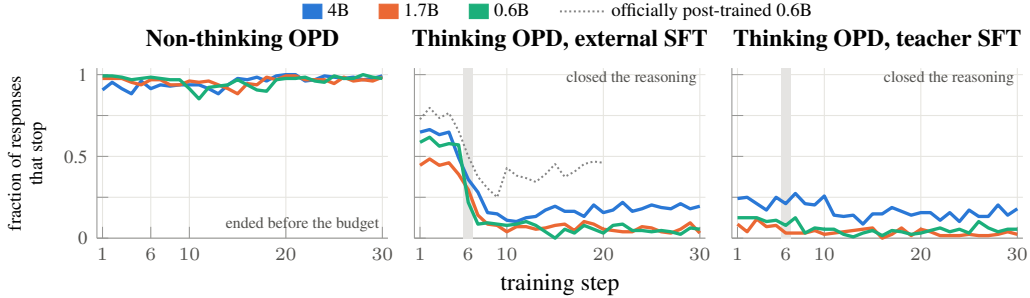
\begin{figure}[t]
\centering
\begin{tikzpicture}
\begin{groupplot}[group style={group size=3 by 1, horizontal sep=0.35cm, yticklabels at=edge left, ylabels at=edge left, group name=cl},
  paperstyle, scale only axis, width=3.85cm, height=2.25cm, xmin=1, xmax=30, ymin=0, ymax=1.04,
  enlarge x limits=0.02, clip=false, xtick={1,6,10,20,30}, ytick={0,0.25,0.5,0.75,1},
  yticklabels={0,,0.5,,1}, axis x line*=bottom, axis y line*=left, title style={yshift=-3pt}]
\nextgroupplot[title={\textbf{Non-thinking OPD}}, ylabel={fraction of responses\\that stop}, ylabel style={align=center, font=\scriptsize}]
\node[font=\tiny, text=cMuted, inner sep=1pt, anchor=south east] at (axis cs:30,0.02) {ended before the budget};
\addplot[cFourB, line width=1.25pt, line join=round, forget plot] coordinates {(1,0.9062) (2,0.9531) (3,0.9141) (4,0.8828) (5,0.9609) (6,0.9141) (7,0.9375) (8,0.9297) (9,0.9375) (10,0.9375) (11,0.9375) (12,0.9141) (13,0.8828) (14,0.9375) (15,0.9766) (16,0.9688) (17,0.9844) (18,0.9609) (19,0.9922) (20,1.0000) (21,1.0000) (22,0.9609) (23,0.9688) (24,0.9922) (25,0.9844) (26,0.9766) (27,0.9844) (28,0.9688) (29,0.9766) (30,0.9922)};
\addplot[cOneSevenB, line width=1.25pt, line join=round, forget plot] coordinates {(1,0.9766) (2,0.9766) (3,0.9766) (4,0.9531) (5,0.9375) (6,0.9688) (7,0.9688) (8,0.9375) (9,0.9375) (10,0.9609) (11,0.9531) (12,0.9609) (13,0.9375) (14,0.9141) (15,0.8828) (16,0.9453) (17,0.9375) (18,0.9844) (19,0.9688) (20,0.9922) (21,0.9922) (22,0.9688) (23,0.9688) (24,0.9688) (25,0.9453) (26,0.9844) (27,0.9609) (28,0.9688) (29,0.9609) (30,0.9844)};
\addplot[cZeroSixB, line width=1.25pt, line join=round, forget plot] coordinates {(1,0.9922) (2,0.9922) (3,0.9844) (4,0.9688) (5,0.9766) (6,0.9844) (7,0.9766) (8,0.9688) (9,0.9688) (10,0.9141) (11,0.8516) (12,0.9219) (13,0.9297) (14,0.9375) (15,0.9688) (16,0.9375) (17,0.9062) (18,0.8984) (19,0.9688) (20,0.9766) (21,0.9766) (22,0.9844) (23,0.9609) (24,0.9531) (25,0.9922) (26,0.9766) (27,0.9766) (28,1.0000) (29,0.9844) (30,0.9766)};
\nextgroupplot[title={\textbf{Thinking OPD, external SFT}}, legend style={at={(0.5,1)}, anchor=south, yshift=14pt, legend columns=4, /tikz/every even column/.append style={column sep=0.9em}}]
\addlegendimage{legend image code/.code={\fill[cFourB] (0cm,-0.6ex) rectangle (0.25cm,0.6ex);}}
\addlegendentry{4B}
\addlegendimage{legend image code/.code={\fill[cOneSevenB] (0cm,-0.6ex) rectangle (0.25cm,0.6ex);}}
\addlegendentry{1.7B}
\addlegendimage{legend image code/.code={\fill[cZeroSixB] (0cm,-0.6ex) rectangle (0.25cm,0.6ex);}}
\addlegendentry{0.6B}
\addlegendimage{line width=.75pt, line join=round, cTeacher!50, densely dotted}
\addlegendentry{officially post-trained 0.6B}
\fill[cNeutral!40] (axis cs:5.5,0) rectangle (axis cs:6.5,1.04);
\node[font=\tiny, text=cMuted, inner sep=1pt, anchor=north east] at (axis cs:30,1.04) {closed the reasoning};
\addplot[cFourB, line width=1.25pt, line join=round, forget plot] coordinates {(1,0.6484) (2,0.6641) (3,0.6328) (4,0.6484) (5,0.4922) (6,0.3594) (7,0.2812) (8,0.1562) (9,0.1484) (10,0.1094) (11,0.1016) (12,0.1250) (13,0.1328) (14,0.1719) (15,0.1953) (16,0.1641) (17,0.1641) (18,0.1328) (19,0.2031) (20,0.1562) (21,0.1719) (22,0.2188) (23,0.1641) (24,0.1797) (25,0.2031) (26,0.1875) (27,0.1797) (28,0.2109) (29,0.1797) (30,0.1953)};
\addplot[cOneSevenB, line width=1.25pt, line join=round, forget plot] coordinates {(1,0.4453) (2,0.4844) (3,0.4453) (4,0.4609) (5,0.3906) (6,0.2969) (7,0.1406) (8,0.0859) (9,0.0781) (10,0.0391) (11,0.0703) (12,0.0703) (13,0.0547) (14,0.0625) (15,0.0781) (16,0.0859) (17,0.0469) (18,0.1016) (19,0.0859) (20,0.0547) (21,0.0469) (22,0.0391) (23,0.0391) (24,0.0703) (25,0.0625) (26,0.0391) (27,0.0312) (28,0.0547) (29,0.0938) (30,0.0312)};
\addplot[cZeroSixB, line width=1.25pt, line join=round, forget plot] coordinates {(1,0.5859) (2,0.6172) (3,0.5625) (4,0.5781) (5,0.5703) (6,0.2188) (7,0.0859) (8,0.0938) (9,0.0859) (10,0.0781) (11,0.0859) (12,0.1016) (13,0.0781) (14,0.0391) (15,0.0000) (16,0.0547) (17,0.0312) (18,0.0781) (19,0.0547) (20,0.0312) (21,0.0781) (22,0.0859) (23,0.0469) (24,0.0469) (25,0.0391) (26,0.0469) (27,0.0391) (28,0.0234) (29,0.0625) (30,0.0547)};
\addplot[line width=.75pt, line join=round, cTeacher!50, densely dotted, forget plot] coordinates {(1,0.7266) (2,0.7969) (3,0.7344) (4,0.7656) (5,0.6562) (6,0.5000) (7,0.3750) (8,0.3047) (9,0.2500) (10,0.4297) (11,0.3828) (12,0.3672) (13,0.3438) (14,0.3906) (15,0.4531) (16,0.3750) (17,0.4062) (18,0.4531) (19,0.4688) (20,0.4609)};
\nextgroupplot[title={\textbf{Thinking OPD, teacher SFT}}]
\fill[cNeutral!40] (axis cs:5.5,0) rectangle (axis cs:6.5,1.04);
\node[font=\tiny, text=cMuted, inner sep=1pt, anchor=north east] at (axis cs:30,1.04) {closed the reasoning};
\addplot[cFourB, line width=1.25pt, line join=round, forget plot] coordinates {(1,0.2422) (2,0.2500) (3,0.2109) (4,0.1719) (5,0.2500) (6,0.2109) (7,0.2734) (8,0.2109) (9,0.2031) (10,0.2578) (11,0.1406) (12,0.1328) (13,0.1406) (14,0.0859) (15,0.1484) (16,0.1484) (17,0.1875) (18,0.1641) (19,0.1406) (20,0.1562) (21,0.1562) (22,0.1094) (23,0.1562) (24,0.1016) (25,0.1719) (26,0.1328) (27,0.1328) (28,0.2031) (29,0.1406) (30,0.1797)};
\addplot[cOneSevenB, line width=1.25pt, line join=round, forget plot] coordinates {(1,0.0859) (2,0.0391) (3,0.1172) (4,0.0703) (5,0.0781) (6,0.0312) (7,0.0312) (8,0.0312) (9,0.0469) (10,0.0234) (11,0.0312) (12,0.0391) (13,0.0469) (14,0.0547) (15,0.0625) (16,0.0000) (17,0.0234) (18,0.0625) (19,0.0156) (20,0.0547) (21,0.0391) (22,0.0156) (23,0.0156) (24,0.0156) (25,0.0234) (26,0.0156) (27,0.0156) (28,0.0234) (29,0.0391) (30,0.0234)};
\addplot[cZeroSixB, line width=1.25pt, line join=round, forget plot] coordinates {(1,0.1250) (2,0.1250) (3,0.1250) (4,0.1016) (5,0.1094) (6,0.0781) (7,0.1250) (8,0.0312) (9,0.0625) (10,0.0547) (11,0.0547) (12,0.0234) (13,0.0078) (14,0.0312) (15,0.0469) (16,0.0156) (17,0.0469) (18,0.0391) (19,0.0234) (20,0.0312) (21,0.0469) (22,0.0625) (23,0.0391) (24,0.0547) (25,0.0312) (26,0.1016) (27,0.0625) (28,0.0391) (29,0.0547) (30,0.0547)};
\end{groupplot}
\path (cl c1r1.south west) -- (cl c3r1.south east) node[midway, below=10pt, font=\footnotesize] {training step};
\end{tikzpicture}
\caption{\textbf{Only thinking-mode students lose their stopping.} Training responses that stop, 7{,}168-token
budget. Non-thinking OPD: almost every response ends before the budget throughout. Thinking OPD from the
external-SFT start: closure collapses at step 6 (gray) at every size; afterwards the 4B keeps some, the 1.7B and 0.6B
almost none (dotted: the officially post-trained 0.6B, 16{,}384 tokens). From the teacher-SFT start the students
close little from the first step, because their responses already fill the budget. Response length, stop reliability
and marked answers: Figure~\ref{fig:collapse}.}
\label{fig:closure}
\end{figure}

%% file: figures/fig_probe.tex
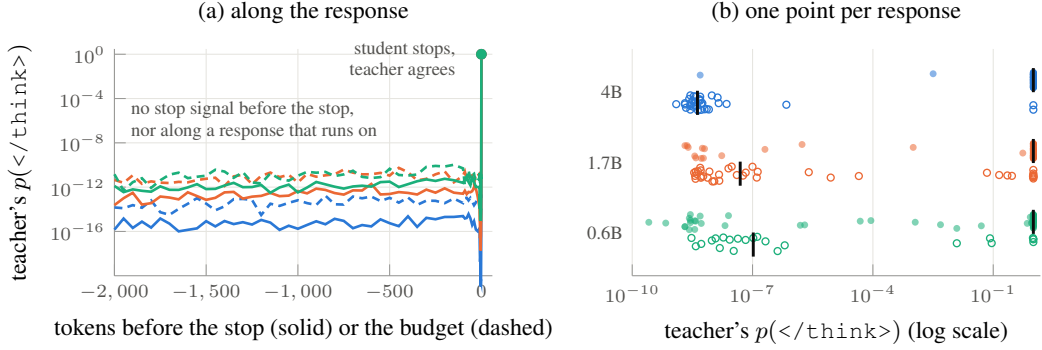
\begin{figure}[t]
\centering
\begin{tikzpicture}
\begin{axis}[paperstyle, scale only axis, width=5.0cm, height=3.0cm, xmin=-2000, xmax=60, ymode=log, log basis y=10, ymin=1e-20, ymax=3, ytick={1e-16,1e-12,1e-8,1e-4,1}, xtick={-2000,-1500,-1000,-500,0}, xticklabel style={/pgf/number format/1000 sep={,}}, axis x line*=bottom, axis y line*=left, clip=false, title={(a) along the response}, xlabel={tokens before the stop (solid) or the budget (dashed)}, ylabel={teacher's $p(\think)$}]
\addplot[cFourB, densely dashed, line width=1.0pt, mark=none] coordinates {(-2000,1.4689e-14) (-1950,1.2457e-14) (-1900,2.3747e-14) (-1850,1.3690e-14) (-1800,6.2748e-14) (-1750,5.4013e-15) (-1700,3.1769e-15) (-1650,1.0844e-14) (-1600,3.1637e-14) (-1550,1.0104e-13) (-1500,3.2539e-14) (-1450,8.0816e-15) (-1400,1.8459e-14) (-1350,8.1941e-14) (-1300,8.5467e-14) (-1250,6.9904e-14) (-1200,3.0346e-15) (-1150,3.1754e-14) (-1100,2.0230e-14) (-1050,3.8397e-14) (-1000,6.5358e-14) (-950,3.6058e-14) (-900,2.4479e-14) (-850,1.7077e-13) (-800,2.8366e-14) (-750,2.6730e-14) (-700,1.4132e-13) (-650,9.8378e-14) (-600,1.4067e-13) (-550,5.5770e-14) (-500,1.0849e-13) (-450,3.9792e-14) (-400,1.1511e-13) (-350,2.4791e-14) (-300,6.0298e-14) (-250,5.8103e-14) (-200,3.5736e-14) (-150,9.3541e-14) (-100,3.6501e-14) (-90,1.0184e-13) (-80,1.0879e-13) (-70,6.2647e-14) (-60,1.0600e-13) (-50,4.9602e-14) (-40,2.3089e-14) (-30,4.0757e-14) (-20,1.6342e-13) (-10,9.5697e-14) (-9,2.2888e-14) (-8,1.1855e-14) (-7,1.5202e-13) (-6,3.9283e-14) (-5,3.2412e-14) (-4,3.7792e-14) (-3,1.8336e-13) (-2,1.0179e-13) (-1,2.3014e-13) (0,1.5171e-13)};
\addplot[cFourB, solid, line width=1.0pt, mark=none] coordinates {(-2000,1.4514e-16) (-1950,1.4632e-15) (-1900,2.0654e-16) (-1850,7.6966e-16) (-1800,2.2156e-16) (-1750,1.4682e-15) (-1700,5.0664e-16) (-1650,1.0137e-16) (-1600,1.3925e-16) (-1550,1.8863e-16) (-1500,8.4840e-16) (-1450,2.8399e-16) (-1400,8.7177e-16) (-1350,4.6121e-16) (-1300,6.8297e-16) (-1250,4.2005e-16) (-1200,1.0985e-15) (-1150,6.2849e-16) (-1100,1.3794e-16) (-1050,3.2832e-16) (-1000,1.7454e-16) (-950,3.8468e-16) (-900,8.2547e-16) (-850,4.4259e-16) (-800,1.3820e-15) (-750,8.9702e-16) (-700,1.9418e-16) (-650,3.7644e-16) (-600,2.8418e-16) (-550,1.2345e-15) (-500,5.5642e-16) (-450,2.9813e-16) (-400,2.3222e-15) (-350,2.5728e-16) (-300,1.6904e-15) (-250,1.2589e-15) (-200,2.0263e-15) (-150,2.0493e-15) (-100,2.4423e-15) (-90,1.4771e-15) (-80,1.6364e-15) (-70,3.6974e-15) (-60,1.5733e-15) (-50,3.8238e-16) (-40,8.3580e-16) (-30,1.4973e-16) (-20,1.0454e-16) (-10,1.9151e-16) (-9,9.0053e-18) (-8,9.1876e-19) (-7,1.8093e-19) (-6,3.7094e-19) (-5,3.1506e-19) (-4,1.0757e-21) (-3,4.3893e-21) (-2,3.6392e-21) (-1,5.4551e-18) (0,1.0000e+00)};
\addplot[only marks, mark=*, mark size=1.8pt, cFourB] coordinates {(0,1.0000e+00)};
\addplot[cOneSevenB, densely dashed, line width=1.0pt, mark=none] coordinates {(-2000,4.5415e-12) (-1950,2.4344e-12) (-1900,1.1679e-11) (-1850,1.9852e-12) (-1800,3.3159e-12) (-1750,2.5533e-12) (-1700,2.3367e-12) (-1650,3.0012e-12) (-1600,2.7114e-12) (-1550,4.3601e-12) (-1500,5.7214e-12) (-1450,7.0534e-12) (-1400,1.2621e-12) (-1350,9.3519e-12) (-1300,1.3110e-11) (-1250,1.8728e-12) (-1200,1.0541e-11) (-1150,2.0592e-11) (-1100,7.9744e-12) (-1050,1.2945e-11) (-1000,2.6461e-12) (-950,1.5805e-11) (-900,8.7096e-12) (-850,1.1671e-11) (-800,1.4514e-11) (-750,3.8521e-11) (-700,2.0361e-11) (-650,1.4272e-11) (-600,1.4548e-11) (-550,1.9409e-11) (-500,1.1071e-11) (-450,5.5132e-11) (-400,2.2187e-11) (-350,4.1295e-12) (-300,5.5963e-12) (-250,1.4398e-11) (-200,2.6687e-11) (-150,2.7164e-12) (-100,1.1133e-11) (-90,2.5445e-11) (-80,5.2204e-12) (-70,7.6648e-12) (-60,1.0184e-11) (-50,1.1397e-11) (-40,1.8793e-12) (-30,1.6967e-11) (-20,1.7717e-11) (-10,8.8491e-12) (-9,1.5410e-11) (-8,1.8210e-11) (-7,1.9002e-11) (-6,9.0511e-12) (-5,4.3501e-12) (-4,2.0170e-12) (-3,8.9702e-12) (-2,1.1452e-11) (-1,9.0970e-12) (0,7.5701e-12)};
\addplot[cOneSevenB, solid, line width=1.0pt, mark=none] coordinates {(-2000,2.3812e-14) (-1950,1.0042e-13) (-1900,3.0882e-13) (-1850,2.6662e-14) (-1800,3.5588e-14) (-1750,1.4194e-13) (-1700,1.1458e-13) (-1650,7.8777e-14) (-1600,3.4143e-13) (-1550,1.5588e-13) (-1500,7.1450e-14) (-1450,1.0794e-13) (-1400,3.3551e-13) (-1350,3.7068e-13) (-1300,9.1833e-14) (-1250,1.3823e-13) (-1200,1.6263e-13) (-1150,2.9971e-13) (-1100,1.5297e-13) (-1050,1.7084e-13) (-1000,4.4689e-13) (-950,2.7397e-13) (-900,7.3131e-13) (-850,5.5950e-13) (-800,2.9228e-13) (-750,2.1140e-13) (-700,5.1109e-13) (-650,4.9659e-13) (-600,1.9395e-12) (-550,2.1807e-13) (-500,2.4010e-13) (-450,3.5703e-13) (-400,3.0068e-13) (-350,2.1687e-13) (-300,5.9951e-13) (-250,3.1572e-13) (-200,7.7696e-13) (-150,3.1652e-13) (-100,2.5351e-13) (-90,4.7457e-13) (-80,2.3153e-13) (-70,3.7025e-13) (-60,3.4324e-13) (-50,4.2865e-13) (-40,9.5830e-13) (-30,7.6173e-13) (-20,1.3747e-13) (-10,1.2996e-14) (-9,9.2491e-16) (-8,6.5058e-16) (-7,2.4446e-17) (-6,5.1761e-17) (-5,2.6990e-16) (-4,1.9147e-18) (-3,2.2967e-17) (-2,4.2238e-17) (-1,1.7612e-15) (0,1.0000e+00)};
\addplot[only marks, mark=*, mark size=1.8pt, cOneSevenB] coordinates {(0,1.0000e+00)};
\addplot[cZeroSixB, densely dashed, line width=1.0pt, mark=none] coordinates {(-2000,1.4508e-11) (-1950,4.7940e-13) (-1900,2.8074e-12) (-1850,6.3504e-12) (-1800,8.5704e-13) (-1750,4.1572e-12) (-1700,1.3468e-11) (-1650,2.0235e-11) (-1600,6.4343e-12) (-1550,8.6596e-12) (-1500,1.0802e-11) (-1450,3.7325e-11) (-1400,2.5562e-11) (-1350,6.9518e-12) (-1300,2.2667e-11) (-1250,2.6297e-12) (-1200,2.1697e-12) (-1150,1.7515e-11) (-1100,5.6260e-12) (-1050,3.8459e-12) (-1000,1.0646e-11) (-950,2.4227e-11) (-900,2.0105e-11) (-850,2.3741e-11) (-800,1.0016e-11) (-750,1.4077e-11) (-700,3.0332e-11) (-650,8.4684e-12) (-600,6.6328e-12) (-550,8.0168e-12) (-500,4.5677e-11) (-450,1.2555e-11) (-400,1.1046e-11) (-350,1.3884e-11) (-300,6.7468e-11) (-250,8.2699e-11) (-200,7.2594e-11) (-150,1.0493e-10) (-100,4.9465e-11) (-90,1.4605e-12) (-80,4.4340e-12) (-70,3.5859e-11) (-60,5.2252e-12) (-50,1.2255e-11) (-40,1.6047e-11) (-30,2.6092e-12) (-20,3.5343e-12) (-10,7.7768e-12) (-9,3.2984e-11) (-8,1.6722e-12) (-7,3.7949e-12) (-6,7.6208e-12) (-5,1.2537e-12) (-4,3.9102e-12) (-3,2.2793e-12) (-2,7.6683e-12) (-1,5.4388e-12) (0,3.8770e-11)};
\addplot[cZeroSixB, solid, line width=1.0pt, mark=none] coordinates {(-2000,1.2756e-12) (-1950,3.5416e-13) (-1900,6.2878e-13) (-1850,4.6591e-13) (-1800,3.8690e-13) (-1750,1.1727e-12) (-1700,5.7478e-13) (-1650,5.6572e-13) (-1600,9.4711e-13) (-1550,1.8531e-12) (-1500,6.6681e-13) (-1450,6.9936e-13) (-1400,1.3948e-12) (-1350,2.4356e-12) (-1300,9.9266e-13) (-1250,1.0207e-12) (-1200,1.5809e-12) (-1150,3.1074e-13) (-1100,9.2449e-13) (-1050,1.6873e-12) (-1000,3.3412e-13) (-950,9.1918e-13) (-900,2.9329e-12) (-850,1.7857e-12) (-800,1.0975e-12) (-750,7.4029e-13) (-700,1.1487e-12) (-650,1.3471e-12) (-600,1.1099e-12) (-550,2.8392e-12) (-500,3.1420e-12) (-450,1.0198e-12) (-400,1.8531e-12) (-350,3.4017e-12) (-300,5.0026e-12) (-250,3.6375e-12) (-200,5.6911e-12) (-150,4.1831e-12) (-100,2.4429e-12) (-90,4.8284e-12) (-80,9.2790e-12) (-70,2.8927e-12) (-60,1.4866e-12) (-50,7.0583e-12) (-40,2.9730e-12) (-30,1.0423e-12) (-20,1.0299e-12) (-10,1.0247e-12) (-9,1.1910e-14) (-8,2.3895e-13) (-7,4.7315e-15) (-6,9.9197e-16) (-5,4.8820e-14) (-4,8.8389e-16) (-3,1.0917e-14) (-2,9.0323e-15) (-1,1.8281e-13) (0,1.0000e+00)};
\addplot[only marks, mark=*, mark size=1.8pt, cZeroSixB] coordinates {(0,1.0000e+00)};
\node[font=\scriptsize, text=cMuted, anchor=east, align=right] at (axis cs:-90,0.3) {student stops,\\ teacher agrees};
\node[font=\scriptsize, text=cMuted, anchor=west, align=left] at (axis cs:-1960,1e-6) {no stop signal before the stop,\\ nor along a response that runs on};
\end{axis}
\begin{axis}[at={(6.85cm,0)}, paperstyle, scale only axis, width=5.4cm, height=3.0cm, xmode=log, log basis x=10, xmin=1e-10, xmax=1.5, ymin=0.4, ymax=3.6, ytick={1,2,3}, yticklabels={0.6B,1.7B,4B}, ytick style={draw=none}, ymajorgrids=false, xtick={1e-10,1e-7,1e-4,1e-1}, axis x line*=bottom, axis y line*=left, y axis line style={draw=none}, clip=false, title={(b) one point per response}, xlabel={teacher's $p(\think)$ (log scale)}]
\fill[cZeroSixB, fill opacity=.55] (axis cs:9.9991e-01,1.236) circle (1.3pt);
\fill[cZeroSixB, fill opacity=.55] (axis cs:1.0000e+00,1.217) circle (1.3pt);
\fill[cZeroSixB, fill opacity=.55] (axis cs:6.8493e-10,1.143) circle (1.3pt);
\fill[cZeroSixB, fill opacity=.55] (axis cs:1.0000e+00,1.107) circle (1.3pt);
\fill[cZeroSixB, fill opacity=.55] (axis cs:1.0000e+00,1.162) circle (1.3pt);
\fill[cZeroSixB, fill opacity=.55] (axis cs:7.0420e-01,1.139) circle (1.3pt);
\fill[cZeroSixB, fill opacity=.55] (axis cs:1.0000e+00,1.222) circle (1.3pt);
\fill[cZeroSixB, fill opacity=.55] (axis cs:1.9947e-06,1.117) circle (1.3pt);
\fill[cZeroSixB, fill opacity=.55] (axis cs:1.0000e+00,1.155) circle (1.3pt);
\fill[cZeroSixB, fill opacity=.55] (axis cs:8.9237e-05,1.178) circle (1.3pt);
\fill[cZeroSixB, fill opacity=.55] (axis cs:1.0000e+00,1.250) circle (1.3pt);
\fill[cZeroSixB, fill opacity=.55] (axis cs:1.0000e+00,1.161) circle (1.3pt);
\fill[cZeroSixB, fill opacity=.55] (axis cs:1.5497e-06,1.112) circle (1.3pt);
\fill[cZeroSixB, fill opacity=.55] (axis cs:9.9972e-01,1.216) circle (1.3pt);
\fill[cZeroSixB, fill opacity=.55] (axis cs:1.0000e+00,1.186) circle (1.3pt);
\fill[cZeroSixB, fill opacity=.55] (axis cs:9.9739e-01,1.105) circle (1.3pt);
\fill[cZeroSixB, fill opacity=.55] (axis cs:1.0000e+00,1.250) circle (1.3pt);
\fill[cZeroSixB, fill opacity=.55] (axis cs:1.0000e+00,1.266) circle (1.3pt);
\fill[cZeroSixB, fill opacity=.55] (axis cs:1.0000e+00,1.228) circle (1.3pt);
\fill[cZeroSixB, fill opacity=.55] (axis cs:2.5108e-08,1.248) circle (1.3pt);
\fill[cZeroSixB, fill opacity=.55] (axis cs:1.0000e+00,1.118) circle (1.3pt);
\fill[cZeroSixB, fill opacity=.55] (axis cs:2.6465e-09,1.211) circle (1.3pt);
\fill[cZeroSixB, fill opacity=.55] (axis cs:1.0000e+00,1.248) circle (1.3pt);
\fill[cZeroSixB, fill opacity=.55] (axis cs:1.0000e+00,1.200) circle (1.3pt);
\fill[cZeroSixB, fill opacity=.55] (axis cs:2.5914e-10,1.154) circle (1.3pt);
\fill[cZeroSixB, fill opacity=.55] (axis cs:1.0000e+00,1.072) circle (1.3pt);
\fill[cZeroSixB, fill opacity=.55] (axis cs:8.9315e-01,1.146) circle (1.3pt);
\fill[cZeroSixB, fill opacity=.55] (axis cs:9.9998e-01,1.184) circle (1.3pt);
\fill[cZeroSixB, fill opacity=.55] (axis cs:1.0000e+00,1.251) circle (1.3pt);
\fill[cZeroSixB, fill opacity=.55] (axis cs:9.9997e-01,1.263) circle (1.3pt);
\fill[cZeroSixB, fill opacity=.55] (axis cs:1.0000e+00,1.155) circle (1.3pt);
\fill[cZeroSixB, fill opacity=.55] (axis cs:9.9999e-01,1.240) circle (1.3pt);
\fill[cZeroSixB, fill opacity=.55] (axis cs:2.0576e-08,1.107) circle (1.3pt);
\fill[cZeroSixB, fill opacity=.55] (axis cs:8.3914e-01,1.227) circle (1.3pt);
\fill[cZeroSixB, fill opacity=.55] (axis cs:3.8507e-09,1.171) circle (1.3pt);
\fill[cZeroSixB, fill opacity=.55] (axis cs:3.6172e-09,1.053) circle (1.3pt);
\fill[cZeroSixB, fill opacity=.55] (axis cs:1.0000e+00,1.208) circle (1.3pt);
\fill[cZeroSixB, fill opacity=.55] (axis cs:5.2630e-09,1.138) circle (1.3pt);
\fill[cZeroSixB, fill opacity=.55] (axis cs:9.9886e-01,1.231) circle (1.3pt);
\fill[cZeroSixB, fill opacity=.55] (axis cs:1.0000e+00,1.197) circle (1.3pt);
\fill[cZeroSixB, fill opacity=.55] (axis cs:2.6466e-09,1.050) circle (1.3pt);
\fill[cZeroSixB, fill opacity=.55] (axis cs:1.0000e+00,1.159) circle (1.3pt);
\fill[cZeroSixB, fill opacity=.55] (axis cs:9.9999e-01,1.241) circle (1.3pt);
\fill[cZeroSixB, fill opacity=.55] (axis cs:5.0480e-02,1.104) circle (1.3pt);
\fill[cZeroSixB, fill opacity=.55] (axis cs:1.0000e+00,1.122) circle (1.3pt);
\fill[cZeroSixB, fill opacity=.55] (axis cs:4.9443e-09,1.242) circle (1.3pt);
\fill[cZeroSixB, fill opacity=.55] (axis cs:2.0611e-09,1.092) circle (1.3pt);
\fill[cZeroSixB, fill opacity=.55] (axis cs:3.6173e-09,1.175) circle (1.3pt);
\fill[cZeroSixB, fill opacity=.55] (axis cs:1.0000e+00,1.102) circle (1.3pt);
\fill[cZeroSixB, fill opacity=.55] (axis cs:2.8172e-09,1.263) circle (1.3pt);
\fill[cZeroSixB, fill opacity=.55] (axis cs:1.0000e+00,1.227) circle (1.3pt);
\fill[cZeroSixB, fill opacity=.55] (axis cs:2.8171e-09,1.149) circle (1.3pt);
\fill[cZeroSixB, fill opacity=.55] (axis cs:1.2353e-02,1.068) circle (1.3pt);
\fill[cZeroSixB, fill opacity=.55] (axis cs:4.8450e-03,1.120) circle (1.3pt);
\fill[cZeroSixB, fill opacity=.55] (axis cs:9.9956e-01,1.162) circle (1.3pt);
\fill[cZeroSixB, fill opacity=.55] (axis cs:9.9927e-01,1.255) circle (1.3pt);
\fill[cZeroSixB, fill opacity=.55] (axis cs:9.9858e-01,1.074) circle (1.3pt);
\fill[cZeroSixB, fill opacity=.55] (axis cs:1.0000e+00,1.171) circle (1.3pt);
\fill[cZeroSixB, fill opacity=.55] (axis cs:1.0000e+00,1.205) circle (1.3pt);
\fill[cZeroSixB, fill opacity=.55] (axis cs:4.8355e-05,1.170) circle (1.3pt);
\fill[cZeroSixB, fill opacity=.55] (axis cs:1.0000e+00,1.229) circle (1.3pt);
\fill[cZeroSixB, fill opacity=.55] (axis cs:9.7346e-01,1.169) circle (1.3pt);
\fill[cZeroSixB, fill opacity=.55] (axis cs:1.0000e+00,1.262) circle (1.3pt);
\fill[cZeroSixB, fill opacity=.55] (axis cs:1.0000e+00,1.183) circle (1.3pt);
\fill[cZeroSixB, fill opacity=.55] (axis cs:9.9992e-01,1.179) circle (1.3pt);
\fill[cZeroSixB, fill opacity=.55] (axis cs:1.0000e+00,1.148) circle (1.3pt);
\fill[cZeroSixB, fill opacity=.55] (axis cs:2.8171e-09,1.181) circle (1.3pt);
\fill[cZeroSixB, fill opacity=.55] (axis cs:9.7743e-01,1.135) circle (1.3pt);
\fill[cZeroSixB, fill opacity=.55] (axis cs:1.0000e+00,1.177) circle (1.3pt);
\fill[cZeroSixB, fill opacity=.55] (axis cs:9.9999e-01,1.114) circle (1.3pt);
\fill[cZeroSixB, fill opacity=.55] (axis cs:2.0611e-09,1.092) circle (1.3pt);
\fill[cZeroSixB, fill opacity=.55] (axis cs:1.0000e+00,1.091) circle (1.3pt);
\fill[cZeroSixB, fill opacity=.55] (axis cs:1.0000e+00,1.185) circle (1.3pt);
\fill[cZeroSixB, fill opacity=.55] (axis cs:1.0000e+00,1.194) circle (1.3pt);
\fill[cZeroSixB, fill opacity=.55] (axis cs:1.1360e-03,1.155) circle (1.3pt);
\draw[cTeacher, line width=1.1pt] (axis cs:9.9997e-01,0.990) -- (axis cs:9.9997e-01,1.330);
\draw[cZeroSixB, line width=.5pt] (axis cs:3.3904e-08,0.750) circle (1.3pt);
\draw[cZeroSixB, line width=.5pt] (axis cs:9.6577e-09,0.897) circle (1.3pt);
\draw[cZeroSixB, line width=.5pt] (axis cs:4.3087e-08,0.923) circle (1.3pt);
\draw[cZeroSixB, line width=.5pt] (axis cs:2.9989e-09,0.933) circle (1.3pt);
\draw[cZeroSixB, line width=.5pt] (axis cs:9.9739e-01,0.915) circle (1.3pt);
\draw[cZeroSixB, line width=.5pt] (axis cs:8.3070e-02,0.928) circle (1.3pt);
\draw[cZeroSixB, line width=.5pt] (axis cs:9.7665e-01,0.933) circle (1.3pt);
\draw[cZeroSixB, line width=.5pt] (axis cs:5.2632e-09,0.849) circle (1.3pt);
\draw[cZeroSixB, line width=.5pt] (axis cs:3.8507e-09,0.816) circle (1.3pt);
\draw[cZeroSixB, line width=.5pt] (axis cs:2.1011e-07,0.885) circle (1.3pt);
\draw[cZeroSixB, line width=.5pt] (axis cs:1.5253e-08,0.791) circle (1.3pt);
\draw[cZeroSixB, line width=.5pt] (axis cs:2.6959e-08,0.909) circle (1.3pt);
\draw[cZeroSixB, line width=.5pt] (axis cs:9.9999e-01,0.917) circle (1.3pt);
\draw[cZeroSixB, line width=.5pt] (axis cs:1.1979e-08,0.927) circle (1.3pt);
\draw[cZeroSixB, line width=.5pt] (axis cs:9.1474e-02,0.860) circle (1.3pt);
\draw[cZeroSixB, line width=.5pt] (axis cs:1.8706e-08,0.939) circle (1.3pt);
\draw[cZeroSixB, line width=.5pt] (axis cs:1.2353e-02,0.858) circle (1.3pt);
\draw[cZeroSixB, line width=.5pt] (axis cs:6.3347e-07,0.829) circle (1.3pt);
\draw[cZeroSixB, line width=.5pt] (axis cs:1.0000e+00,0.875) circle (1.3pt);
\draw[cZeroSixB, line width=.5pt] (axis cs:1.3089e-07,0.949) circle (1.3pt);
\draw[cZeroSixB, line width=.5pt] (axis cs:1.0422e-07,0.932) circle (1.3pt);
\draw[cZeroSixB, line width=.5pt] (axis cs:6.7123e-08,0.905) circle (1.3pt);
\draw[cZeroSixB, line width=.5pt] (axis cs:4.2110e-07,0.748) circle (1.3pt);
\draw[cTeacher, line width=1.1pt] (axis cs:1.0422e-07,0.670) -- (axis cs:1.0422e-07,1.010);
\fill[cOneSevenB, fill opacity=.55] (axis cs:1.0000e+00,2.185) circle (1.3pt);
\fill[cOneSevenB, fill opacity=.55] (axis cs:1.0000e+00,2.157) circle (1.3pt);
\fill[cOneSevenB, fill opacity=.55] (axis cs:9.9983e-01,2.189) circle (1.3pt);
\fill[cOneSevenB, fill opacity=.55] (axis cs:2.3355e-09,2.236) circle (1.3pt);
\fill[cOneSevenB, fill opacity=.55] (axis cs:1.0000e+00,2.103) circle (1.3pt);
\fill[cOneSevenB, fill opacity=.55] (axis cs:2.4862e-09,2.211) circle (1.3pt);
\fill[cOneSevenB, fill opacity=.55] (axis cs:1.0000e+00,2.076) circle (1.3pt);
\fill[cOneSevenB, fill opacity=.55] (axis cs:1.0000e+00,2.099) circle (1.3pt);
\fill[cOneSevenB, fill opacity=.55] (axis cs:1.0000e+00,2.225) circle (1.3pt);
\fill[cOneSevenB, fill opacity=.55] (axis cs:1.0000e+00,2.123) circle (1.3pt);
\fill[cOneSevenB, fill opacity=.55] (axis cs:1.0000e+00,2.230) circle (1.3pt);
\fill[cOneSevenB, fill opacity=.55] (axis cs:1.0000e+00,2.072) circle (1.3pt);
\fill[cOneSevenB, fill opacity=.55] (axis cs:1.0000e+00,2.082) circle (1.3pt);
\fill[cOneSevenB, fill opacity=.55] (axis cs:1.0000e+00,2.203) circle (1.3pt);
\fill[cOneSevenB, fill opacity=.55] (axis cs:1.0000e+00,2.060) circle (1.3pt);
\fill[cOneSevenB, fill opacity=.55] (axis cs:1.0000e+00,2.176) circle (1.3pt);
\fill[cOneSevenB, fill opacity=.55] (axis cs:1.0000e+00,2.250) circle (1.3pt);
\fill[cOneSevenB, fill opacity=.55] (axis cs:1.0000e+00,2.168) circle (1.3pt);
\fill[cOneSevenB, fill opacity=.55] (axis cs:1.0000e+00,2.200) circle (1.3pt);
\fill[cOneSevenB, fill opacity=.55] (axis cs:5.2632e-09,2.056) circle (1.3pt);
\fill[cOneSevenB, fill opacity=.55] (axis cs:1.0000e+00,2.190) circle (1.3pt);
\fill[cOneSevenB, fill opacity=.55] (axis cs:2.0368e-07,2.183) circle (1.3pt);
\fill[cOneSevenB, fill opacity=.55] (axis cs:1.0000e+00,2.177) circle (1.3pt);
\fill[cOneSevenB, fill opacity=.55] (axis cs:1.0000e+00,2.136) circle (1.3pt);
\fill[cOneSevenB, fill opacity=.55] (axis cs:5.4629e-01,2.131) circle (1.3pt);
\fill[cOneSevenB, fill opacity=.55] (axis cs:1.0000e+00,2.266) circle (1.3pt);
\fill[cOneSevenB, fill opacity=.55] (axis cs:1.0000e+00,2.058) circle (1.3pt);
\fill[cOneSevenB, fill opacity=.55] (axis cs:9.9790e-01,2.055) circle (1.3pt);
\fill[cOneSevenB, fill opacity=.55] (axis cs:1.0000e+00,2.261) circle (1.3pt);
\fill[cOneSevenB, fill opacity=.55] (axis cs:1.0000e+00,2.091) circle (1.3pt);
\fill[cOneSevenB, fill opacity=.55] (axis cs:1.0000e+00,2.077) circle (1.3pt);
\fill[cOneSevenB, fill opacity=.55] (axis cs:1.0000e+00,2.096) circle (1.3pt);
\fill[cOneSevenB, fill opacity=.55] (axis cs:1.0000e+00,2.226) circle (1.3pt);
\fill[cOneSevenB, fill opacity=.55] (axis cs:1.0000e+00,2.256) circle (1.3pt);
\fill[cOneSevenB, fill opacity=.55] (axis cs:1.0000e+00,2.055) circle (1.3pt);
\fill[cOneSevenB, fill opacity=.55] (axis cs:1.0000e+00,2.144) circle (1.3pt);
\fill[cOneSevenB, fill opacity=.55] (axis cs:1.0000e+00,2.072) circle (1.3pt);
\fill[cOneSevenB, fill opacity=.55] (axis cs:1.0000e+00,2.107) circle (1.3pt);
\fill[cOneSevenB, fill opacity=.55] (axis cs:1.0000e+00,2.099) circle (1.3pt);
\fill[cOneSevenB, fill opacity=.55] (axis cs:6.3477e-09,2.192) circle (1.3pt);
\fill[cOneSevenB, fill opacity=.55] (axis cs:1.0000e+00,2.127) circle (1.3pt);
\fill[cOneSevenB, fill opacity=.55] (axis cs:3.6173e-09,2.090) circle (1.3pt);
\fill[cOneSevenB, fill opacity=.55] (axis cs:9.9998e-01,2.161) circle (1.3pt);
\fill[cOneSevenB, fill opacity=.55] (axis cs:5.9638e-09,2.059) circle (1.3pt);
\fill[cOneSevenB, fill opacity=.55] (axis cs:1.0000e+00,2.072) circle (1.3pt);
\fill[cOneSevenB, fill opacity=.55] (axis cs:1.0000e+00,2.267) circle (1.3pt);
\fill[cOneSevenB, fill opacity=.55] (axis cs:1.7048e-08,2.094) circle (1.3pt);
\fill[cOneSevenB, fill opacity=.55] (axis cs:1.0000e+00,2.129) circle (1.3pt);
\fill[cOneSevenB, fill opacity=.55] (axis cs:1.0110e-03,2.211) circle (1.3pt);
\fill[cOneSevenB, fill opacity=.55] (axis cs:1.0000e+00,2.234) circle (1.3pt);
\fill[cOneSevenB, fill opacity=.55] (axis cs:1.7135e-06,2.252) circle (1.3pt);
\fill[cOneSevenB, fill opacity=.55] (axis cs:1.0000e+00,2.087) circle (1.3pt);
\fill[cOneSevenB, fill opacity=.55] (axis cs:1.0000e+00,2.198) circle (1.3pt);
\fill[cOneSevenB, fill opacity=.55] (axis cs:1.0000e+00,2.263) circle (1.3pt);
\fill[cOneSevenB, fill opacity=.55] (axis cs:3.8507e-09,2.063) circle (1.3pt);
\fill[cOneSevenB, fill opacity=.55] (axis cs:5.6026e-09,2.199) circle (1.3pt);
\fill[cOneSevenB, fill opacity=.55] (axis cs:2.0611e-09,2.236) circle (1.3pt);
\draw[cTeacher, line width=1.1pt] (axis cs:1.0000e+00,1.990) -- (axis cs:1.0000e+00,2.330);
\draw[cOneSevenB, line width=.5pt] (axis cs:1.0000e+00,1.805) circle (1.3pt);
\draw[cOneSevenB, line width=.5pt] (axis cs:8.8808e-06,1.785) circle (1.3pt);
\draw[cOneSevenB, line width=.5pt] (axis cs:2.5194e-06,1.861) circle (1.3pt);
\draw[cOneSevenB, line width=.5pt] (axis cs:1.4591e-08,1.827) circle (1.3pt);
\draw[cOneSevenB, line width=.5pt] (axis cs:3.7856e-09,1.768) circle (1.3pt);
\draw[cOneSevenB, line width=.5pt] (axis cs:4.0989e-09,1.834) circle (1.3pt);
\draw[cOneSevenB, line width=.5pt] (axis cs:1.0000e+00,1.820) circle (1.3pt);
\draw[cOneSevenB, line width=.5pt] (axis cs:3.4995e-09,1.855) circle (1.3pt);
\draw[cOneSevenB, line width=.5pt] (axis cs:1.0000e+00,1.842) circle (1.3pt);
\draw[cOneSevenB, line width=.5pt] (axis cs:5.2632e-09,1.799) circle (1.3pt);
\draw[cOneSevenB, line width=.5pt] (axis cs:3.7020e-09,1.809) circle (1.3pt);
\draw[cOneSevenB, line width=.5pt] (axis cs:2.5108e-08,1.914) circle (1.3pt);
\draw[cOneSevenB, line width=.5pt] (axis cs:1.2740e-07,1.785) circle (1.3pt);
\draw[cOneSevenB, line width=.5pt] (axis cs:6.9784e-02,1.853) circle (1.3pt);
\draw[cOneSevenB, line width=.5pt] (axis cs:1.1108e-08,1.733) circle (1.3pt);
\draw[cOneSevenB, line width=.5pt] (axis cs:9.9906e-01,1.893) circle (1.3pt);
\draw[cOneSevenB, line width=.5pt] (axis cs:5.9087e-09,1.804) circle (1.3pt);
\draw[cOneSevenB, line width=.5pt] (axis cs:1.4804e-08,1.740) circle (1.3pt);
\draw[cOneSevenB, line width=.5pt] (axis cs:1.0000e+00,1.792) circle (1.3pt);
\draw[cOneSevenB, line width=.5pt] (axis cs:5.1678e-09,1.783) circle (1.3pt);
\draw[cOneSevenB, line width=.5pt] (axis cs:3.0175e-08,1.940) circle (1.3pt);
\draw[cOneSevenB, line width=.5pt] (axis cs:4.4618e-05,1.807) circle (1.3pt);
\draw[cOneSevenB, line width=.5pt] (axis cs:1.0000e+00,1.793) circle (1.3pt);
\draw[cOneSevenB, line width=.5pt] (axis cs:2.9251e-01,1.809) circle (1.3pt);
\draw[cOneSevenB, line width=.5pt] (axis cs:9.9540e-08,1.938) circle (1.3pt);
\draw[cOneSevenB, line width=.5pt] (axis cs:3.8506e-09,1.869) circle (1.3pt);
\draw[cOneSevenB, line width=.5pt] (axis cs:1.3426e-07,1.867) circle (1.3pt);
\draw[cOneSevenB, line width=.5pt] (axis cs:5.1025e-09,1.887) circle (1.3pt);
\draw[cOneSevenB, line width=.5pt] (axis cs:2.2515e-01,1.815) circle (1.3pt);
\draw[cOneSevenB, line width=.5pt] (axis cs:4.0555e-08,1.821) circle (1.3pt);
\draw[cOneSevenB, line width=.5pt] (axis cs:4.3634e-09,1.873) circle (1.3pt);
\draw[cOneSevenB, line width=.5pt] (axis cs:1.0007e-08,1.730) circle (1.3pt);
\draw[cOneSevenB, line width=.5pt] (axis cs:9.9960e-01,1.772) circle (1.3pt);
\draw[cOneSevenB, line width=.5pt] (axis cs:9.8103e-01,1.804) circle (1.3pt);
\draw[cOneSevenB, line width=.5pt] (axis cs:7.5785e-09,1.783) circle (1.3pt);
\draw[cOneSevenB, line width=.5pt] (axis cs:8.3938e-09,1.870) circle (1.3pt);
\draw[cOneSevenB, line width=.5pt] (axis cs:5.7852e-08,1.813) circle (1.3pt);
\draw[cOneSevenB, line width=.5pt] (axis cs:4.9444e-09,1.923) circle (1.3pt);
\draw[cOneSevenB, line width=.5pt] (axis cs:6.7465e-08,1.855) circle (1.3pt);
\draw[cOneSevenB, line width=.5pt] (axis cs:1.4080e-01,1.821) circle (1.3pt);
\draw[cTeacher, line width=1.1pt] (axis cs:4.9203e-08,1.670) -- (axis cs:4.9203e-08,2.010);
\fill[cFourB, fill opacity=.55] (axis cs:1.0000e+00,3.138) circle (1.3pt);
\fill[cFourB, fill opacity=.55] (axis cs:1.0000e+00,3.204) circle (1.3pt);
\fill[cFourB, fill opacity=.55] (axis cs:1.0000e+00,3.142) circle (1.3pt);
\fill[cFourB, fill opacity=.55] (axis cs:1.0000e+00,3.196) circle (1.3pt);
\fill[cFourB, fill opacity=.55] (axis cs:1.0000e+00,3.060) circle (1.3pt);
\fill[cFourB, fill opacity=.55] (axis cs:1.0000e+00,3.148) circle (1.3pt);
\fill[cFourB, fill opacity=.55] (axis cs:1.0000e+00,3.107) circle (1.3pt);
\fill[cFourB, fill opacity=.55] (axis cs:1.0000e+00,3.085) circle (1.3pt);
\fill[cFourB, fill opacity=.55] (axis cs:1.0000e+00,3.166) circle (1.3pt);
\fill[cFourB, fill opacity=.55] (axis cs:1.0000e+00,3.157) circle (1.3pt);
\fill[cFourB, fill opacity=.55] (axis cs:1.0000e+00,3.174) circle (1.3pt);
\fill[cFourB, fill opacity=.55] (axis cs:1.0000e+00,3.216) circle (1.3pt);
\fill[cFourB, fill opacity=.55] (axis cs:1.0000e+00,3.244) circle (1.3pt);
\fill[cFourB, fill opacity=.55] (axis cs:1.0000e+00,3.159) circle (1.3pt);
\fill[cFourB, fill opacity=.55] (axis cs:1.0000e+00,3.119) circle (1.3pt);
\fill[cFourB, fill opacity=.55] (axis cs:1.0000e+00,3.153) circle (1.3pt);
\fill[cFourB, fill opacity=.55] (axis cs:1.0000e+00,3.228) circle (1.3pt);
\fill[cFourB, fill opacity=.55] (axis cs:1.0000e+00,3.243) circle (1.3pt);
\fill[cFourB, fill opacity=.55] (axis cs:1.0000e+00,3.229) circle (1.3pt);
\fill[cFourB, fill opacity=.55] (axis cs:1.0000e+00,3.091) circle (1.3pt);
\fill[cFourB, fill opacity=.55] (axis cs:1.0000e+00,3.270) circle (1.3pt);
\fill[cFourB, fill opacity=.55] (axis cs:1.0000e+00,3.189) circle (1.3pt);
\fill[cFourB, fill opacity=.55] (axis cs:1.0000e+00,3.068) circle (1.3pt);
\fill[cFourB, fill opacity=.55] (axis cs:1.0000e+00,3.210) circle (1.3pt);
\fill[cFourB, fill opacity=.55] (axis cs:1.0000e+00,3.267) circle (1.3pt);
\fill[cFourB, fill opacity=.55] (axis cs:1.0000e+00,3.138) circle (1.3pt);
\fill[cFourB, fill opacity=.55] (axis cs:1.0000e+00,3.199) circle (1.3pt);
\fill[cFourB, fill opacity=.55] (axis cs:1.0000e+00,3.120) circle (1.3pt);
\fill[cFourB, fill opacity=.55] (axis cs:1.0000e+00,3.097) circle (1.3pt);
\fill[cFourB, fill opacity=.55] (axis cs:9.9973e-01,3.208) circle (1.3pt);
\fill[cFourB, fill opacity=.55] (axis cs:1.0000e+00,3.051) circle (1.3pt);
\fill[cFourB, fill opacity=.55] (axis cs:4.9443e-09,3.231) circle (1.3pt);
\fill[cFourB, fill opacity=.55] (axis cs:1.0000e+00,3.166) circle (1.3pt);
\fill[cFourB, fill opacity=.55] (axis cs:1.0000e+00,3.072) circle (1.3pt);
\fill[cFourB, fill opacity=.55] (axis cs:1.0000e+00,3.076) circle (1.3pt);
\fill[cFourB, fill opacity=.55] (axis cs:1.0000e+00,3.193) circle (1.3pt);
\fill[cFourB, fill opacity=.55] (axis cs:1.0000e+00,3.242) circle (1.3pt);
\fill[cFourB, fill opacity=.55] (axis cs:1.0000e+00,3.112) circle (1.3pt);
\fill[cFourB, fill opacity=.55] (axis cs:1.0000e+00,3.265) circle (1.3pt);
\fill[cFourB, fill opacity=.55] (axis cs:1.0000e+00,3.072) circle (1.3pt);
\fill[cFourB, fill opacity=.55] (axis cs:1.0000e+00,3.238) circle (1.3pt);
\fill[cFourB, fill opacity=.55] (axis cs:1.0000e+00,3.137) circle (1.3pt);
\fill[cFourB, fill opacity=.55] (axis cs:1.0000e+00,3.068) circle (1.3pt);
\fill[cFourB, fill opacity=.55] (axis cs:1.0000e+00,3.110) circle (1.3pt);
\fill[cFourB, fill opacity=.55] (axis cs:1.0000e+00,3.150) circle (1.3pt);
\fill[cFourB, fill opacity=.55] (axis cs:1.0000e+00,3.224) circle (1.3pt);
\fill[cFourB, fill opacity=.55] (axis cs:1.0000e+00,3.239) circle (1.3pt);
\fill[cFourB, fill opacity=.55] (axis cs:1.0000e+00,3.079) circle (1.3pt);
\fill[cFourB, fill opacity=.55] (axis cs:9.8994e-01,3.165) circle (1.3pt);
\fill[cFourB, fill opacity=.55] (axis cs:1.0000e+00,3.193) circle (1.3pt);
\fill[cFourB, fill opacity=.55] (axis cs:1.0000e+00,3.126) circle (1.3pt);
\fill[cFourB, fill opacity=.55] (axis cs:9.8780e-01,3.242) circle (1.3pt);
\fill[cFourB, fill opacity=.55] (axis cs:1.0000e+00,3.111) circle (1.3pt);
\fill[cFourB, fill opacity=.55] (axis cs:1.0000e+00,3.054) circle (1.3pt);
\fill[cFourB, fill opacity=.55] (axis cs:1.0000e+00,3.059) circle (1.3pt);
\fill[cFourB, fill opacity=.55] (axis cs:1.0000e+00,3.200) circle (1.3pt);
\fill[cFourB, fill opacity=.55] (axis cs:1.0000e+00,3.173) circle (1.3pt);
\fill[cFourB, fill opacity=.55] (axis cs:1.0000e+00,3.258) circle (1.3pt);
\fill[cFourB, fill opacity=.55] (axis cs:1.0000e+00,3.256) circle (1.3pt);
\fill[cFourB, fill opacity=.55] (axis cs:3.1756e-03,3.250) circle (1.3pt);
\fill[cFourB, fill opacity=.55] (axis cs:1.0000e+00,3.059) circle (1.3pt);
\fill[cFourB, fill opacity=.55] (axis cs:1.0000e+00,3.215) circle (1.3pt);
\fill[cFourB, fill opacity=.55] (axis cs:1.0000e+00,3.204) circle (1.3pt);
\fill[cFourB, fill opacity=.55] (axis cs:1.0000e+00,3.194) circle (1.3pt);
\fill[cFourB, fill opacity=.55] (axis cs:1.0000e+00,3.207) circle (1.3pt);
\fill[cFourB, fill opacity=.55] (axis cs:1.0000e+00,3.249) circle (1.3pt);
\fill[cFourB, fill opacity=.55] (axis cs:1.0000e+00,3.191) circle (1.3pt);
\fill[cFourB, fill opacity=.55] (axis cs:1.0000e+00,3.132) circle (1.3pt);
\fill[cFourB, fill opacity=.55] (axis cs:1.0000e+00,3.168) circle (1.3pt);
\fill[cFourB, fill opacity=.55] (axis cs:1.0000e+00,3.096) circle (1.3pt);
\fill[cFourB, fill opacity=.55] (axis cs:1.0000e+00,3.179) circle (1.3pt);
\fill[cFourB, fill opacity=.55] (axis cs:1.0000e+00,3.052) circle (1.3pt);
\fill[cFourB, fill opacity=.55] (axis cs:1.0000e+00,3.083) circle (1.3pt);
\fill[cFourB, fill opacity=.55] (axis cs:1.0000e+00,3.123) circle (1.3pt);
\fill[cFourB, fill opacity=.55] (axis cs:1.0000e+00,3.224) circle (1.3pt);
\fill[cFourB, fill opacity=.55] (axis cs:1.0000e+00,3.208) circle (1.3pt);
\fill[cFourB, fill opacity=.55] (axis cs:1.0000e+00,3.124) circle (1.3pt);
\fill[cFourB, fill opacity=.55] (axis cs:1.0000e+00,3.187) circle (1.3pt);
\fill[cFourB, fill opacity=.55] (axis cs:1.0000e+00,3.059) circle (1.3pt);
\fill[cFourB, fill opacity=.55] (axis cs:1.0000e+00,3.086) circle (1.3pt);
\fill[cFourB, fill opacity=.55] (axis cs:1.0000e+00,3.266) circle (1.3pt);
\fill[cFourB, fill opacity=.55] (axis cs:1.0000e+00,3.114) circle (1.3pt);
\fill[cFourB, fill opacity=.55] (axis cs:1.0000e+00,3.137) circle (1.3pt);
\draw[cTeacher, line width=1.1pt] (axis cs:1.0000e+00,2.990) -- (axis cs:1.0000e+00,3.330);
\draw[cFourB, line width=.5pt] (axis cs:3.8507e-09,2.851) circle (1.3pt);
\draw[cFourB, line width=.5pt] (axis cs:4.0989e-09,2.795) circle (1.3pt);
\draw[cFourB, line width=.5pt] (axis cs:1.4307e-08,2.835) circle (1.3pt);
\draw[cFourB, line width=.5pt] (axis cs:3.3982e-09,2.783) circle (1.3pt);
\draw[cFourB, line width=.5pt] (axis cs:5.2631e-09,2.741) circle (1.3pt);
\draw[cFourB, line width=.5pt] (axis cs:2.1940e-09,2.770) circle (1.3pt);
\draw[cFourB, line width=.5pt] (axis cs:2.9989e-09,2.845) circle (1.3pt);
\draw[cFourB, line width=.5pt] (axis cs:8.1514e-09,2.746) circle (1.3pt);
\draw[cFourB, line width=.5pt] (axis cs:2.8172e-09,2.819) circle (1.3pt);
\draw[cFourB, line width=.5pt] (axis cs:3.8505e-09,2.802) circle (1.3pt);
\draw[cFourB, line width=.5pt] (axis cs:1.2501e-09,2.821) circle (1.3pt);
\draw[cFourB, line width=.5pt] (axis cs:7.1940e-09,2.752) circle (1.3pt);
\draw[cFourB, line width=.5pt] (axis cs:3.8507e-09,2.930) circle (1.3pt);
\draw[cFourB, line width=.5pt] (axis cs:5.6027e-09,2.834) circle (1.3pt);
\draw[cFourB, line width=.5pt] (axis cs:1.3438e-08,2.915) circle (1.3pt);
\draw[cFourB, line width=.5pt] (axis cs:2.8172e-09,2.945) circle (1.3pt);
\draw[cFourB, line width=.5pt] (axis cs:1.0000e+00,2.806) circle (1.3pt);
\draw[cFourB, line width=.5pt] (axis cs:2.0611e-09,2.835) circle (1.3pt);
\draw[cFourB, line width=.5pt] (axis cs:4.9444e-09,2.884) circle (1.3pt);
\draw[cFourB, line width=.5pt] (axis cs:2.2493e-08,2.824) circle (1.3pt);
\draw[cFourB, line width=.5pt] (axis cs:2.8172e-09,2.796) circle (1.3pt);
\draw[cFourB, line width=.5pt] (axis cs:4.6447e-09,2.892) circle (1.3pt);
\draw[cFourB, line width=.5pt] (axis cs:4.9443e-09,2.927) circle (1.3pt);
\draw[cFourB, line width=.5pt] (axis cs:3.3982e-09,2.932) circle (1.3pt);
\draw[cFourB, line width=.5pt] (axis cs:3.1922e-09,2.868) circle (1.3pt);
\draw[cFourB, line width=.5pt] (axis cs:6.9658e-07,2.813) circle (1.3pt);
\draw[cFourB, line width=.5pt] (axis cs:9.6577e-09,2.944) circle (1.3pt);
\draw[cFourB, line width=.5pt] (axis cs:4.9440e-09,2.871) circle (1.3pt);
\draw[cFourB, line width=.5pt] (axis cs:2.8171e-09,2.744) circle (1.3pt);
\draw[cFourB, line width=.5pt] (axis cs:6.3477e-09,2.749) circle (1.3pt);
\draw[cFourB, line width=.5pt] (axis cs:2.6325e-09,2.895) circle (1.3pt);
\draw[cFourB, line width=.5pt] (axis cs:1.0000e+00,2.743) circle (1.3pt);
\draw[cFourB, line width=.5pt] (axis cs:4.3633e-09,2.732) circle (1.3pt);
\draw[cFourB, line width=.5pt] (axis cs:2.0611e-09,2.817) circle (1.3pt);
\draw[cFourB, line width=.5pt] (axis cs:3.6173e-09,2.844) circle (1.3pt);
\draw[cFourB, line width=.5pt] (axis cs:5.2632e-09,2.829) circle (1.3pt);
\draw[cTeacher, line width=1.1pt] (axis cs:4.2311e-09,2.670) -- (axis cs:4.2311e-09,3.010);
\end{axis}
\end{tikzpicture}
\caption{\textbf{The teacher agrees to stop where the student stopped, and almost nowhere along a response that
never stops.} The students' first training responses (external-SFT start, 7{,}168-token budget), scored token by token
by the thinking teacher; color: \protect\sizedot{cFourB}\,4B, \protect\sizedot{cOneSevenB}\,1.7B, \protect\sizedot{cZeroSixB}\,0.6B.
(a)~The teacher's probability of \think{}, median over responses, at each position up to 2{,}000 tokens before the
point where the student closed its reasoning (solid) or before the end of a response that ran to the budget (dashed).
(b)~Filled: the probability at the position where the student closed. Open: the largest probability anywhere along a
response that ran to the budget. Black ticks: medians. Numbers: Appendix Table~\ref{tab:probe}.}
\label{fig:probe}
\end{figure}

%% file: figures/fig_filter.tex
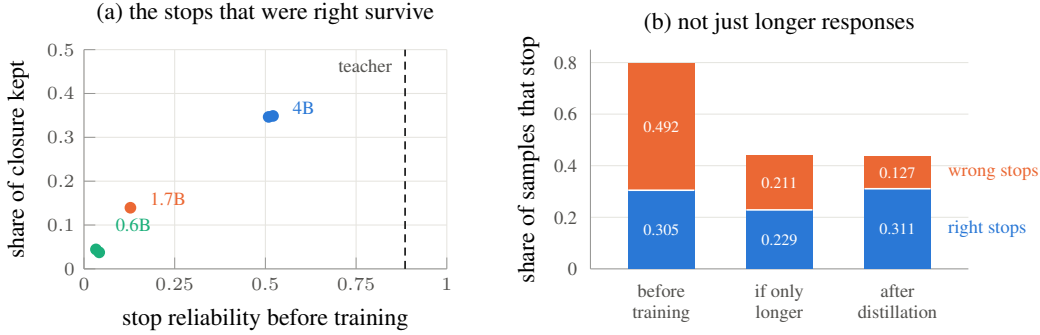
\begin{figure}[t]
\centering
\begin{tikzpicture}
\begin{axis}[paperstyle, scale only axis, width=4.8cm, height=2.9cm, xmin=0, xmax=1, ymin=0, ymax=0.5, xtick={0,0.25,0.5,0.75,1}, ytick={0,0.1,0.2,0.3,0.4,0.5}, axis x line*=bottom, axis y line*=left, clip=false, title={(a) the stops that were right survive}, xlabel={stop reliability before training}, ylabel={share of closure kept}]
\draw[cTeacher, densely dashed, line width=.6pt] (axis cs:0.885,0) -- (axis cs:0.885,0.5);
\node[font=\scriptsize, text=cMuted, anchor=north east] at (axis cs:0.875,0.5) {teacher};
\fill[cFourB] (axis cs:0.5210,0.3488) circle (2.2pt);
\fill[cFourB] (axis cs:0.5090,0.3466) circle (2.2pt);
\fill[cOneSevenB] (axis cs:0.1280,0.1394) circle (2.2pt);
\fill[cZeroSixB] (axis cs:0.0420,0.0376) circle (2.2pt);
\fill[cZeroSixB] (axis cs:0.0330,0.0444) circle (2.2pt);
\node[font=\scriptsize, text=cFourB, anchor=west] at (axis cs:0.5410,0.3688) {\,4B};
\node[font=\scriptsize, text=cOneSevenB, anchor=west] at (axis cs:0.1480,0.1594) {\,1.7B};
\node[font=\scriptsize, text=cZeroSixB, anchor=south west] at (axis cs:0.0530,0.0644) {\,0.6B};
\end{axis}
\begin{scope}
\draw[cGrid] (6.700,0.682) -- (11.700,0.682);
\draw[cGrid] (6.700,1.365) -- (11.700,1.365);
\draw[cGrid] (6.700,2.047) -- (11.700,2.047);
\draw[cGrid] (6.700,2.729) -- (11.700,2.729);
\draw[cMuted!60] (6.700,0) -- (11.700,0);
\draw[cMuted!60] (6.700,0) -- (6.700,2.900);
\node[font=\scriptsize, text=cMuted, anchor=east] at (6.650,0.000) {0};
\node[font=\scriptsize, text=cMuted, anchor=east] at (6.650,0.682) {0.2};
\node[font=\scriptsize, text=cMuted, anchor=east] at (6.650,1.365) {0.4};
\node[font=\scriptsize, text=cMuted, anchor=east] at (6.650,2.047) {0.6};
\node[font=\scriptsize, text=cMuted, anchor=east] at (6.650,2.729) {0.8};
\fill[cFourB] (7.200,0) rectangle (8.075,1.040);
\fill[cOneSevenB] (7.200,1.040) rectangle (8.075,2.719);
\draw[white, line width=.6pt] (7.200,1.040) -- (8.075,1.040);
\node[font=\tiny, text=white] at (7.638,0.520) {0.305};
\node[font=\tiny, text=white] at (7.638,1.879) {0.492};
\node[font=\scriptsize, text=cMuted, align=center, anchor=north] at (7.638,-0.08) {before\\training};
\fill[cFourB] (8.762,0) rectangle (9.637,0.780);
\fill[cOneSevenB] (8.762,0.780) rectangle (9.637,1.499);
\draw[white, line width=.6pt] (8.762,0.780) -- (9.637,0.780);
\node[font=\tiny, text=white] at (9.200,0.390) {0.229};
\node[font=\tiny, text=white] at (9.200,1.139) {0.211};
\node[font=\scriptsize, text=cMuted, align=center, anchor=north] at (9.200,-0.08) {if only\\longer};
\fill[cFourB] (10.325,0) rectangle (11.200,1.060);
\fill[cOneSevenB] (10.325,1.060) rectangle (11.200,1.493);
\draw[white, line width=.6pt] (10.325,1.060) -- (11.200,1.060);
\node[font=\tiny, text=white] at (10.762,0.530) {0.311};
\node[font=\tiny, text=white] at (10.762,1.276) {0.127};
\node[font=\scriptsize, text=cMuted, align=center, anchor=north] at (10.762,-0.08) {after\\distillation};
\node[font=\scriptsize, text=cFourB, anchor=west] at (11.309,0.530) {right stops};
\node[font=\scriptsize, text=cOneSevenB, anchor=west] at (11.309,1.276) {wrong stops};
\node[font=\footnotesize, rotate=90, anchor=south] at (6.150,1.450) {share of samples that stop};
\node[font=\footnotesize] at (9.200,3.250) {(b) not just longer responses};
\end{scope}
\end{tikzpicture}
\caption{\textbf{Distillation filters stops: it keeps the ones that were right.} (a)~Each run from the
external-SFT start: the fraction of its stops that landed on a correct answer over the first four steps, against the
share of its early closure it keeps late in training (color: size; the teacher's stop reliability dashed). Runs from the
teacher-SFT start, which close little from the first step, and the 20-step runs of the officially post-trained 0.6B are
in Appendix Table~\ref{tab:app-reliability}. (b)~Could the lost stops simply be responses grown too long for the
limit? Each bar splits the officially post-trained 0.6B's samples that stop into those whose answer is right (blue)
and wrong (orange). Left: before distillation. Right: after 20 thinking-mode steps. Middle: what lengthening alone
would give, if every starting response were stretched by one factor, chosen so that as many samples stop as after
distillation. Lengthening alone would cut right and wrong stops alike; distillation instead keeps the right stops and
removes mostly the wrong ones (Appendix Table~\ref{tab:app-nullmodel}).}
\label{fig:filter}
\end{figure}

%% file: sections/05_unfinished.tex
\section{What the samples that do not stop contain}
\label{sec:unfinished}

The small students often reach the right value but do not commit to it. From the external-SFT start they rarely mark
it; from the teacher-SFT start they mark it and write past it. Below 4B, distillation trades wrong answers for no answer. We sort every evaluation sample by what it marks
(Figure~\ref{fig:overview}b; per size in Appendix Figure~\ref{fig:cases}). From the external-SFT start, samples whose marked answers are all wrong shrink at every size,
for instance from 0.433 to 0.127 of all samples at 0.6B. At 4B their place is taken by right answers, and samples
without any marked answer stay where they were. At 1.7B and 0.6B their place is taken mostly by samples that mark no
answer, which rise from 0.373 to 0.518 and from 0.328 to 0.579, while right answers rise only a little; every such sample ran to the evaluation limit.

\input{tables/tab_unfinished}

From the external-SFT start, these unfinished samples, the ones cut off at the length limit, almost never mark an
answer, yet many of them already contain the correct value, early. Among the samples that reach the limit, only 0.066,
0.055 and 0.107 mark any answer at 4B, 1.7B and 0.6B
(Table~\ref{tab:unfinished}), so lenient scoring cannot credit the rest, whatever their reasoning contains. We search
the text of each unfinished sample for the correct value, on problems whose answer has at least three digits and does not
occur in the statement, and compare with how often the answer of a different problem appears. The correct value appears
in 0.211, 0.298 and 0.240 of these samples, four to seven times the chance rate, and more often than before distillation.
It first appears around the first fifth of the text, and the student then writes tens of thousands of further
characters. The unfinished training responses look the same: after the collapse, the correct value appears in 0.309 of
the 4B's, against 0.019 by chance.

Forcing a stop points the same way. When we append \think{} to the unfinished samples of
the officially post-trained 0.6B and let it answer \citep{muennighoff2025s1}, it answers correctly in 0.364 of them on
problems its starting model had solved three or four times in four, and in 0.049 on problems the starting model never
solved (Appendix~\ref{app:block3}): where the student can solve, much of what these samples lack is the stop. The
smallest students' unfinished samples also often end in repetition: 0.084 at
4B, about as many as for the teacher (0.077), but 0.197 and 0.276 at 1.7B and 0.6B. From the teacher-SFT start the
failure takes another form. The 1.7B and 0.6B distilled from it never finish a sample at evaluation, yet they mark 185.5
and 150.3 answers per sample on average; they mark a first answer halfway through, at 0.44 and 0.50 of the output against
0.85 for the teacher, and then overwrite it: when a correct answer was marked, the last marked answer is wrong in 0.437
and 0.180 of those samples, against 0.004 for the teacher (Appendix~\ref{app:block3}).

%% file: tables/tab_unfinished.tex
\begin{table}[t]
\centering
\caption{\textbf{From the external-SFT start, the unfinished samples rarely mark an answer, but the correct
value often appears in them, early; from the teacher-SFT start they mark answers but never finish.} Samples truncated at the cap (30{,}720 tokens; GSM8K-200 8{,}192), benchmarks pooled.
Value present: the correct value occurs in the text (answers of $\ge$3 digits not in the statement); chance: another
problem's answer; the dot plot shows the two (dot: present, tick: chance). First at, characters after it: medians.
Loop: a line repeated $\ge$5 times at the end.
Details: Table~\ref{tab:app-unfinished-bench}.}
\label{tab:unfinished}
\footnotesize
\setlength{\tabcolsep}{3.4pt}
\renewcommand{\arraystretch}{0.95}
\begin{tabular}{@{}lcccccccc@{}}
\toprule
 & All samples & \multicolumn{7}{c}{Truncated samples} \\
\cmidrule(lr){2-2}\cmidrule(lr){3-9}
Model & Truncated & \makecell{With a\\marked answer} & \makecell{Correct value\\present} & Chance & \makecell{present\\vs.\ chance} & \makecell{First\\at} & \makecell{Chars.\\after it} & Loop \\
\midrule
Teacher (thinking) & 0.045 & 0.173 & 0.184 & 0.063 & \vsdot{cTeacher}{0.1839}{0.0632} & 0.48 & 23,492 & 0.077 \\
\midrule
\multicolumn{9}{@{}l}{\textit{External-SFT start, before and after thinking-mode distillation}} \\
\quad \sizedot{cFourB}\,\textcolor{cMuted}{\textit{4B start}} & \textcolor{cMuted}{0.274} & \textcolor{cMuted}{0.035} & \textcolor{cMuted}{0.067} & \textcolor{cMuted}{0.028} & \vsdot{cFourB!45}{0.0668}{0.0276} & \textcolor{cMuted}{0.08} & \textcolor{cMuted}{33,181} & \textcolor{cMuted}{0.115} \\
\quad\phantom{\sizedot{cMuted}\,}after & 0.284 & 0.066 & 0.211 & 0.049 & \vsdot{cFourB}{0.2108}{0.0491} & 0.21 & 29,796 & 0.084 \\
\addlinespace[1.5pt]
\quad \sizedot{cOneSevenB}\,\textcolor{cMuted}{\textit{1.7B start}} & \textcolor{cMuted}{0.387} & \textcolor{cMuted}{0.047} & \textcolor{cMuted}{0.126} & \textcolor{cMuted}{0.024} & \vsdot{cOneSevenB!45}{0.1260}{0.0244} & \textcolor{cMuted}{0.07} & \textcolor{cMuted}{25,693} & \textcolor{cMuted}{0.163} \\
\quad\phantom{\sizedot{cMuted}\,}after & 0.540 & 0.055 & 0.298 & 0.043 & \vsdot{cOneSevenB}{0.2976}{0.0433} & 0.14 & 26,362 & 0.197 \\
\addlinespace[1.5pt]
\quad \sizedot{cZeroSixB}\,\textcolor{cMuted}{\textit{0.6B start}} & \textcolor{cMuted}{0.392} & \textcolor{cMuted}{0.178} & \textcolor{cMuted}{0.169} & \textcolor{cMuted}{0.036} & \vsdot{cZeroSixB!45}{0.1693}{0.0357} & \textcolor{cMuted}{0.06} & \textcolor{cMuted}{23,474} & \textcolor{cMuted}{0.202} \\
\quad\phantom{\sizedot{cMuted}\,}after & 0.627 & 0.107 & 0.240 & 0.037 & \vsdot{cZeroSixB}{0.2404}{0.0372} & 0.21 & 20,757 & 0.276 \\
\midrule
\multicolumn{9}{@{}l}{\textit{Teacher-SFT start, after distillation}} \\
\quad \sizedot{cOneSevenB}\,1.7B & 1.000 & 0.640 & 0.614 & 0.053 & \vsdot{cOneSevenB}{0.6142}{0.0535} & 0.07 & 23,486 & 0.650 \\
\quad \sizedot{cZeroSixB}\,0.6B & 1.000 & 0.506 & 0.506 & 0.045 & \vsdot{cZeroSixB}{0.5059}{0.0455} & 0.07 & 22,972 & 0.567 \\
\bottomrule
\end{tabular}
\end{table}

%% file: sections/06_related.tex
\section{Related work}
\label{sec:related}

On-policy distillation trains the student on its own samples \citep{ross2011dagger,agarwal2024gkd,gu2024minillm,tml2025opd}
rather than on the teacher's outputs \citep{hinton2015distilling,kim2016sequence}. Recent variants change the divergence
\citep{jin2026eopd,fu2026revisiting} or the credit assigned over time \citep{liu2026beyond}, cut rollouts where the
teacher's signal weakens \citep{ziheng2026esr,xin2026kat}, or mask the stop token's advantage to contain length
\citep{he2026simpleopd}. Analyses relate success to compatible thinking patterns \citep{li2026rethinking}, report length
exploitation \citep{wang2026demystify}, find that distillation improves sampling efficiency more than coverage
\citep{ge2026tts}, and describe outcome-confounded supervision
\citep{ma2026outcome} and privileged teachers that harm thinking students \citep{kaur2026rethinking}. We hold one recipe
fixed across student sizes and locate the loss of stopping in the states the student reaches, not in the token.
For reinforcement learning with verifiable rewards, single-sample gains have been found to stay within what the base
model reaches in many samples \citep{yue2025rlvr}. For on-policy distillation, how far a student can rise, in thinking
and in non-thinking mode and as a function of its size, has not been measured; Section~\ref{sec:ceiling} does so.

Prior work treats stopping as a problem of its own, separate from solving. Long reasoning overspends on easy problems
\citep{chen2024overthinking} and abandons promising lines too early \citep{wang2025underthinking}; models carry a
signal that they have already answered correctly \citep{zhang2025probing}, and inference-time methods impose a stop
through budget forcing \citep{muennighoff2025s1}, early exit \citep{yang2026deer} or abstention
\citep{davidov2026quit}. These methods decide when a trained model should stop; we ask what training teaches about
stopping. We find that on-policy distillation filters a student's stops rather than teaching them.

Small dense models learn long reasoning from a strong model's traces \citep{guo2025deepseekr1}, but the smallest learn
better from short chains than from long ones \citep{li2025smallmodels}. In imitation learning, a learner that lacks
information the expert acts on cannot recover the expert's decisions: an imitation gap remains
\citep{weihs2021imitationgap}, and learners trained off-policy latch onto correlates \citep{dehaan2019causal,swamy2022unobserved}.
Stopping is such a decision: the teacher stops at the end of a finished derivation, which the small student rarely
reaches.

%% file: sections/07_conclusion.tex
\section{Conclusion}
\label{sec:conclusion}

We analyzed on-policy distillation at small scale across three student sizes, both modes and two sources of
fine-tuning data. It transfers solving but, in thinking mode, not stopping. Solving improves at every size up to two
ceilings, the student's own reach before training and a share of the teacher that shrinks with the student, which we
measured across both modes and all sizes. In non-thinking mode every student keeps stopping; in thinking mode
distillation filters stops rather than teaching them, so the smaller the student, the less stopping survives, and the
smallest students reach the right value without committing to it. Our diagnostic, which separates answering,
correctness and stopping, is what makes this visible: an end-point score alone would read the small students' lost
stops as lost solving. These results locate where methods for small students should act: along the responses that run
on, where the teacher gives no signal to stop, rather than at the stop token itself. We hope they offer guidance for
distilling small reasoning models: how far solving can rise, why stopping fails in thinking mode, and what the
unfinished outputs of the small students contain.

%% file: sections/11_statements.tex
\section*{AI use statement}

We used generative AI tools mainly for two purposes: to improve the writing of this paper, and to help write analysis and
plotting scripts. The authors designed the study, ran all training and evaluation, and checked every analysis, number and
claim against the recorded logs and data. Generative AI was not used to generate data or to run the experiments;
translation and dataset cleaning are not applicable to this work. We have reviewed all AI-assisted content and take
responsibility for the final content of this work, including text, claims or artifacts produced with the aid of generative AI.

\section*{Ethics statement}

This study distills mathematical reasoning between openly released language models and evaluates it on
public benchmarks. It involves no human subjects, no personal data and no new data collection: the
training prompts are a public competition-mathematics set, and the cold-start solutions come from a public
corpus or from the teacher itself. We report how a widely used recipe can fail on small students so that
the failure can be detected, and we see no direct route to misuse.

\section*{Reproducibility statement}

Section~\ref{sec:setup} and Appendix~\ref{app:setup} specify the teacher, the students, the objective, the
hyperparameters, the data pipeline, every training run and the evaluation protocol. The teacher and the
pretrained-only and officially post-trained students are public checkpoints. We will release the code, the derived data
and the configuration of every run upon acceptance.

%% file: appendix/A_setup.tex
\section{Experimental details}
\label{app:setup}

\paragraph{Training problems.}
We started from the 17,398 problems of DAPO-Math-17k \citep{yu2025dapo}. Cleaning left 15,827; removing exact duplicates
(924), near duplicates with a 13-gram Jaccard similarity of at least 0.7 (167), and problems that also occur in AIME
2024--2026, AMC23 or MATH (1,139) left 13,597. Every prompt ends with the instruction to reason step by step and to put
the final answer in \verb|\boxed{}|.

\paragraph{The two cold starts.}
The short external solutions are all 5,878 solutions of at most 2,048 tokens (mean 1,530) in a quality-filtered 45k
subset of OpenR1-Math-220k (Apache-2.0), written as a thinking block followed by a boxed answer. The teacher's own
solutions are, for each of 5,000 training problems, the shortest correct thinking-mode sample among the teacher's four
(temperature 0.6), kept if it has at most 7,000 tokens and is not predominantly Chinese: 1,524 solutions with a mean of
4,809 tokens. The same fine-tuning recipe is used at every size (Table~\ref{tab:app-hparams}).

\input{figures/fig_valk}

\input{tables/lean_runs}

\input{tables/lean_hparams}

%% file: figures/fig_valk.tex
\begin{figure}[t]
\centering
\begin{tikzpicture}
\begin{groupplot}[group style={group size=3 by 4, horizontal sep=0.30cm, vertical sep=0.24cm,
    xticklabels at=edge bottom, yticklabels at=edge left, ylabels at=edge left, group name=vk},
  paperstyle, scale only axis, width=3.90cm, height=1.62cm,
  xmin=0, xmax=212, enlarge x limits=0.02, clip=false,
  xtick={0,20,100,142,212}, axis x line*=bottom, axis y line*=left, title style={yshift=-3pt},
  ylabel style={align=center, font=\scriptsize}]
\nextgroupplot[ymin=0, ymax=0.50, ytick={0,0.25,0.5}, yticklabels={0,,0.5}, title={\textbf{Non-thinking OPD}}, ylabel={AIME 2025\\$p@1$}]
\fill[cNeutral!28] (axis cs:0,0) rectangle (axis cs:20,0.500);
\addplot[cFourB, line width=.75pt, line join=round, densely dashed, opacity=.8, forget plot] coordinates {(0,0.0417) (20,0.1708) (40,0.1417) (60,0.1708) (80,0.1667) (99,0.1750)};
\addplot[cOneSevenB, line width=.75pt, line join=round, densely dashed, opacity=.8, forget plot] coordinates {(0,0.0167) (20,0.0542) (40,0.0542) (60,0.0583) (80,0.0458) (99,0.0625)};
\addplot[cFourB, line width=1.15pt, line join=round, mark=*, mark size=.95pt, forget plot] coordinates {(0,0.0708) (20,0.1708) (40,0.1500) (60,0.1750) (80,0.1750) (100,0.1833) (120,0.1833) (140,0.1708) (160,0.1625) (180,0.1917) (200,0.1667) (212,0.1708)};
\addplot[cOneSevenB, line width=1.15pt, line join=round, mark=*, mark size=.95pt, forget plot] coordinates {(0,0.0333) (20,0.0792) (40,0.0542) (60,0.0458) (80,0.0500) (100,0.0625) (120,0.0458) (140,0.0583) (160,0.0958) (180,0.0667) (200,0.0500) (212,0.0458)};
\addplot[cZeroSixB, line width=1.15pt, line join=round, mark=*, mark size=.95pt, forget plot] coordinates {(0,0.0000) (20,0.0167) (40,0.0042) (60,0.0250) (80,0.0167) (100,0.0208) (120,0.0042) (140,0.0083) (160,0.0083) (180,0.0042) (200,0.0125) (212,0.0125)};
\nextgroupplot[ymin=0, ymax=0.50, ytick={0,0.25,0.5}, yticklabels={0,,0.5}, title={\textbf{Thinking OPD, external SFT}}, legend style={at={(0.5,1)}, anchor=south, yshift=14pt, legend columns=7, /tikz/every even column/.append style={column sep=0.9em}}]
\addlegendimage{legend image code/.code={\fill[cFourB] (0cm,-0.6ex) rectangle (0.25cm,0.6ex);}}
\addlegendentry{4B}
\addlegendimage{legend image code/.code={\fill[cOneSevenB] (0cm,-0.6ex) rectangle (0.25cm,0.6ex);}}
\addlegendentry{1.7B}
\addlegendimage{legend image code/.code={\fill[cZeroSixB] (0cm,-0.6ex) rectangle (0.25cm,0.6ex);}}
\addlegendentry{0.6B}
\addlegendimage{cMuted, line width=1.15pt, line join=round, mark=*, mark size=.95pt}
\addlegendentry{run}
\addlegendimage{cMuted, line width=.65pt, line join=round, opacity=.55}
\addlegendentry{second run}
\addlegendimage{cMuted, line width=.75pt, line join=round, densely dashed, opacity=.8}
\addlegendentry{easy-to-hard curriculum}
\addlegendimage{legend image code/.code={\fill[cNeutral!35] (0cm,-0.8ex) rectangle (0.3cm,0.8ex);}}
\addlegendentry{early steps}
\fill[cNeutral!28] (axis cs:0,0) rectangle (axis cs:20,0.500);
\addplot[cFourB, line width=.65pt, line join=round, opacity=.55, forget plot] coordinates {(0,0.1750) (20,0.1833) (40,0.1625) (60,0.1833) (80,0.1917) (100,0.2625) (120,0.2042) (140,0.2500) (160,0.2292)};
\addplot[cZeroSixB, line width=.65pt, line join=round, opacity=.55, forget plot] coordinates {(0,0.0083) (20,0.0000) (40,0.0000) (60,0.0000)};
\addplot[cFourB, line width=1.15pt, line join=round, mark=*, mark size=.95pt, forget plot] coordinates {(0,0.1958) (20,0.1875) (40,0.1917) (60,0.2042) (80,0.1917) (100,0.2333) (120,0.1875) (140,0.2333) (160,0.2125)};
\addplot[cOneSevenB, line width=1.15pt, line join=round, mark=*, mark size=.95pt, forget plot] coordinates {(0,0.0583) (20,0.0375) (40,0.0583) (60,0.0750) (80,0.0917) (100,0.0625) (120,0.0708) (140,0.0917) (160,0.0708) (180,0.0708) (200,0.0833)};
\addplot[cZeroSixB, line width=1.15pt, line join=round, mark=*, mark size=.95pt, forget plot] coordinates {(0,0.0000) (20,0.0000) (40,0.0000) (60,0.0000) (80,0.0042) (100,0.0042) (120,0.0042) (140,0.0042) (160,0.0083) (180,0.0042) (200,0.0042)};
\nextgroupplot[ymin=0, ymax=0.50, ytick={0,0.25,0.5}, yticklabels={0,,0.5}, title={\textbf{Thinking OPD, teacher SFT}}]
\fill[cNeutral!28] (axis cs:0,0) rectangle (axis cs:20,0.500);
\addplot[cFourB, line width=1.15pt, line join=round, mark=*, mark size=.95pt, forget plot] coordinates {(0,0.1250) (20,0.1625) (40,0.2042) (60,0.2042) (80,0.2333) (100,0.2208) (120,0.2458) (140,0.2042) (160,0.2375) (180,0.2333) (200,0.2250) (212,0.2542)};
\addplot[cOneSevenB, line width=1.15pt, line join=round, mark=*, mark size=.95pt, forget plot] coordinates {(0,0.0083) (20,0.0458) (40,0.0458) (60,0.0375) (80,0.0542) (100,0.0750) (120,0.0792) (140,0.0875) (142,0.0667)};
\addplot[cZeroSixB, line width=1.15pt, line join=round, mark=*, mark size=.95pt, forget plot] coordinates {(0,0.0042) (20,0.0042) (40,0.0000) (60,0.0083) (80,0.0083) (100,0.0042) (120,0.0083) (140,0.0083) (142,0.0042)};
\nextgroupplot[ymin=0, ymax=0.75, ytick={0,0.25,0.5,0.75}, yticklabels={0,,0.5,}, ylabel={AMC23\\$p@1$}]
\fill[cNeutral!28] (axis cs:0,0) rectangle (axis cs:20,0.750);
\addplot[cFourB, line width=.75pt, line join=round, densely dashed, opacity=.8, forget plot] coordinates {(0,0.3688) (20,0.5250) (40,0.5969) (60,0.5625) (80,0.5750) (99,0.6062)};
\addplot[cOneSevenB, line width=.75pt, line join=round, densely dashed, opacity=.8, forget plot] coordinates {(0,0.2437) (20,0.3875) (40,0.4000) (60,0.4156) (80,0.4000) (99,0.4313)};
\addplot[cFourB, line width=1.15pt, line join=round, mark=*, mark size=.95pt, forget plot] coordinates {(0,0.4156) (20,0.5594) (40,0.5687) (60,0.5500) (80,0.6125) (100,0.5375) (120,0.6156) (140,0.5938) (160,0.5750) (180,0.5938) (200,0.5906) (212,0.5469)};
\addplot[cOneSevenB, line width=1.15pt, line join=round, mark=*, mark size=.95pt, forget plot] coordinates {(0,0.2219) (20,0.3438) (40,0.4000) (60,0.3719) (80,0.3875) (100,0.3781) (120,0.4094) (140,0.4094) (160,0.4313) (180,0.3969) (200,0.3906) (212,0.4406)};
\addplot[cZeroSixB, line width=1.15pt, line join=round, mark=*, mark size=.95pt, forget plot] coordinates {(0,0.0437) (20,0.2313) (40,0.2531) (60,0.2437) (80,0.2781) (100,0.2625) (120,0.2625) (140,0.2906) (160,0.2969) (180,0.2594) (200,0.3031) (212,0.3187)};
\nextgroupplot[ymin=0, ymax=0.75, ytick={0,0.25,0.5,0.75}, yticklabels={0,,0.5,}]
\fill[cNeutral!28] (axis cs:0,0) rectangle (axis cs:20,0.750);
\addplot[cFourB, line width=.65pt, line join=round, opacity=.55, forget plot] coordinates {(0,0.5781) (20,0.5563) (40,0.5906) (60,0.6062) (80,0.6219) (100,0.6531) (120,0.6188) (140,0.6281) (160,0.6719)};
\addplot[cZeroSixB, line width=.65pt, line join=round, opacity=.55, forget plot] coordinates {(0,0.1781) (20,0.0469) (40,0.1375) (60,0.1719)};
\addplot[cFourB, line width=1.15pt, line join=round, mark=*, mark size=.95pt, forget plot] coordinates {(0,0.5844) (20,0.5219) (40,0.5969) (60,0.5906) (80,0.5844) (100,0.6094) (120,0.6500) (140,0.6625) (160,0.6531)};
\addplot[cOneSevenB, line width=1.15pt, line join=round, mark=*, mark size=.95pt, forget plot] coordinates {(0,0.3000) (20,0.2687) (40,0.2969) (60,0.3187) (80,0.3469) (100,0.3406) (120,0.3406) (140,0.3250) (160,0.3281) (180,0.3531) (200,0.3438)};
\addplot[cZeroSixB, line width=1.15pt, line join=round, mark=*, mark size=.95pt, forget plot] coordinates {(0,0.1844) (20,0.0500) (40,0.1125) (60,0.2000) (80,0.1781) (100,0.1875) (120,0.1406) (140,0.1406) (160,0.1938) (180,0.1688) (200,0.1375)};
\nextgroupplot[ymin=0, ymax=0.75, ytick={0,0.25,0.5,0.75}, yticklabels={0,,0.5,}]
\fill[cNeutral!28] (axis cs:0,0) rectangle (axis cs:20,0.750);
\addplot[cFourB, line width=1.15pt, line join=round, mark=*, mark size=.95pt, forget plot] coordinates {(0,0.3844) (20,0.5969) (40,0.5625) (60,0.5844) (80,0.6281) (100,0.5906) (120,0.6156) (140,0.6188) (160,0.6469) (180,0.6500) (200,0.6500) (212,0.6687)};
\addplot[cOneSevenB, line width=1.15pt, line join=round, mark=*, mark size=.95pt, forget plot] coordinates {(0,0.1156) (20,0.2875) (40,0.3406) (60,0.3250) (80,0.3281) (100,0.3531) (120,0.3594) (140,0.3250) (142,0.3031)};
\addplot[cZeroSixB, line width=1.15pt, line join=round, mark=*, mark size=.95pt, forget plot] coordinates {(0,0.0312) (20,0.0813) (40,0.1719) (60,0.1594) (80,0.1781) (100,0.1656) (120,0.1406) (140,0.2000) (142,0.1500)};
\nextgroupplot[ymin=0, ymax=0.50, ytick={0,0.25,0.5}, yticklabels={0,,0.5}, ylabel={AIME 2025\\$p@8$}]
\fill[cNeutral!28] (axis cs:0,0) rectangle (axis cs:20,0.500);
\addplot[cFourB, line width=.75pt, line join=round, densely dashed, opacity=.8, forget plot] coordinates {(0,0.2000) (20,0.3333) (40,0.2667) (60,0.4000) (80,0.3333) (99,0.4000)};
\addplot[cOneSevenB, line width=.75pt, line join=round, densely dashed, opacity=.8, forget plot] coordinates {(0,0.0667) (20,0.2667) (40,0.2000) (60,0.1333) (80,0.1667) (99,0.2000)};
\addplot[cFourB, line width=1.15pt, line join=round, mark=*, mark size=.95pt, forget plot] coordinates {(0,0.1667) (20,0.4000) (40,0.2667) (60,0.4000) (80,0.3333) (100,0.3000) (120,0.3333) (140,0.2667) (160,0.3000) (180,0.4333) (200,0.3667) (212,0.3000)};
\addplot[cOneSevenB, line width=1.15pt, line join=round, mark=*, mark size=.95pt, forget plot] coordinates {(0,0.1000) (20,0.2333) (40,0.2333) (60,0.1333) (80,0.2000) (100,0.2000) (120,0.2000) (140,0.2000) (160,0.2333) (180,0.1667) (200,0.1667) (212,0.1667)};
\addplot[cZeroSixB, line width=1.15pt, line join=round, mark=*, mark size=.95pt, forget plot] coordinates {(0,0.0000) (20,0.1333) (40,0.0333) (60,0.1000) (80,0.1000) (100,0.1667) (120,0.0333) (140,0.0667) (160,0.0667) (180,0.0333) (200,0.0667) (212,0.0667)};
\nextgroupplot[ymin=0, ymax=0.50, ytick={0,0.25,0.5}, yticklabels={0,,0.5}]
\fill[cNeutral!28] (axis cs:0,0) rectangle (axis cs:20,0.500);
\addplot[cFourB, line width=.65pt, line join=round, opacity=.55, forget plot] coordinates {(0,0.3000) (20,0.3667) (40,0.3000) (60,0.3333) (80,0.3333) (100,0.4333) (120,0.3667) (140,0.4000) (160,0.4000)};
\addplot[cZeroSixB, line width=.65pt, line join=round, opacity=.55, forget plot] coordinates {(0,0.0667) (20,0.0000) (40,0.0000) (60,0.0000)};
\addplot[cFourB, line width=1.15pt, line join=round, mark=*, mark size=.95pt, forget plot] coordinates {(0,0.3667) (20,0.3333) (40,0.3333) (60,0.3667) (80,0.4000) (100,0.4333) (120,0.3667) (140,0.4000) (160,0.3000)};
\addplot[cOneSevenB, line width=1.15pt, line join=round, mark=*, mark size=.95pt, forget plot] coordinates {(0,0.1667) (20,0.2000) (40,0.1667) (60,0.1333) (80,0.1667) (100,0.2000) (120,0.2000) (140,0.1667) (160,0.2000) (180,0.1667) (200,0.2000)};
\addplot[cZeroSixB, line width=1.15pt, line join=round, mark=*, mark size=.95pt, forget plot] coordinates {(0,0.0000) (20,0.0000) (40,0.0000) (60,0.0000) (80,0.0333) (100,0.0333) (120,0.0333) (140,0.0333) (160,0.0667) (180,0.0333) (200,0.0333)};
\nextgroupplot[ymin=0, ymax=0.50, ytick={0,0.25,0.5}, yticklabels={0,,0.5}]
\fill[cNeutral!28] (axis cs:0,0) rectangle (axis cs:20,0.500);
\addplot[cFourB, line width=1.15pt, line join=round, mark=*, mark size=.95pt, forget plot] coordinates {(0,0.3333) (20,0.2667) (40,0.3667) (60,0.3333) (80,0.4000) (100,0.3667) (120,0.4000) (140,0.3667) (160,0.3667) (180,0.3667) (200,0.3667) (212,0.3667)};
\addplot[cOneSevenB, line width=1.15pt, line join=round, mark=*, mark size=.95pt, forget plot] coordinates {(0,0.0333) (20,0.1333) (40,0.1333) (60,0.1667) (80,0.1667) (100,0.2333) (120,0.1667) (140,0.2333) (142,0.2000)};
\addplot[cZeroSixB, line width=1.15pt, line join=round, mark=*, mark size=.95pt, forget plot] coordinates {(0,0.0333) (20,0.0333) (40,0.0000) (60,0.0667) (80,0.0667) (100,0.0333) (120,0.0667) (140,0.0667) (142,0.0333)};
\nextgroupplot[ymin=0, ymax=1.00, ytick={0,0.25,0.5,0.75,1}, yticklabels={0,,0.5,,1}, ylabel={AMC23\\$p@8$}]
\fill[cNeutral!28] (axis cs:0,0) rectangle (axis cs:20,1.000);
\addplot[cFourB, line width=.75pt, line join=round, densely dashed, opacity=.8, forget plot] coordinates {(0,0.7000) (20,0.8000) (40,0.8750) (60,0.8500) (80,0.8250) (99,0.8500)};
\addplot[cOneSevenB, line width=.75pt, line join=round, densely dashed, opacity=.8, forget plot] coordinates {(0,0.6500) (20,0.6750) (40,0.7500) (60,0.6750) (80,0.6250) (99,0.6750)};
\addplot[cFourB, line width=1.15pt, line join=round, mark=*, mark size=.95pt, forget plot] coordinates {(0,0.7500) (20,0.9000) (40,0.8000) (60,0.8250) (80,0.8250) (100,0.8250) (120,0.9000) (140,0.8500) (160,0.9000) (180,0.7750) (200,0.9000) (212,0.8500)};
\addplot[cOneSevenB, line width=1.15pt, line join=round, mark=*, mark size=.95pt, forget plot] coordinates {(0,0.5750) (20,0.6500) (40,0.7250) (60,0.7000) (80,0.7750) (100,0.6750) (120,0.6750) (140,0.7750) (160,0.7500) (180,0.7000) (200,0.7250) (212,0.8000)};
\addplot[cZeroSixB, line width=1.15pt, line join=round, mark=*, mark size=.95pt, forget plot] coordinates {(0,0.2250) (20,0.5250) (40,0.6000) (60,0.5500) (80,0.6000) (100,0.5000) (120,0.5750) (140,0.6000) (160,0.5500) (180,0.5250) (200,0.6250) (212,0.6000)};
\nextgroupplot[ymin=0, ymax=1.00, ytick={0,0.25,0.5,0.75,1}, yticklabels={0,,0.5,,1}]
\fill[cNeutral!28] (axis cs:0,0) rectangle (axis cs:20,1.000);
\addplot[cFourB, line width=.65pt, line join=round, opacity=.55, forget plot] coordinates {(0,0.9250) (20,0.8000) (40,0.7500) (60,0.7750) (80,0.7750) (100,0.8000) (120,0.8250) (140,0.7750) (160,0.8250)};
\addplot[cZeroSixB, line width=.65pt, line join=round, opacity=.55, forget plot] coordinates {(0,0.5000) (20,0.1750) (40,0.2750) (60,0.2750)};
\addplot[cFourB, line width=1.15pt, line join=round, mark=*, mark size=.95pt, forget plot] coordinates {(0,0.8750) (20,0.7250) (40,0.8250) (60,0.8000) (80,0.7500) (100,0.8250) (120,0.8250) (140,0.8750) (160,0.7750)};
\addplot[cOneSevenB, line width=1.15pt, line join=round, mark=*, mark size=.95pt, forget plot] coordinates {(0,0.6500) (20,0.4750) (40,0.4750) (60,0.5000) (80,0.5000) (100,0.4750) (120,0.4500) (140,0.4750) (160,0.5000) (180,0.5750) (200,0.5500)};
\addplot[cZeroSixB, line width=1.15pt, line join=round, mark=*, mark size=.95pt, forget plot] coordinates {(0,0.4250) (20,0.2000) (40,0.3000) (60,0.4250) (80,0.3500) (100,0.3500) (120,0.2500) (140,0.2750) (160,0.2750) (180,0.3500) (200,0.3000)};
\nextgroupplot[ymin=0, ymax=1.00, ytick={0,0.25,0.5,0.75,1}, yticklabels={0,,0.5,,1}]
\fill[cNeutral!28] (axis cs:0,0) rectangle (axis cs:20,1.000);
\addplot[cFourB, line width=1.15pt, line join=round, mark=*, mark size=.95pt, forget plot] coordinates {(0,0.7000) (20,0.7750) (40,0.7500) (60,0.7750) (80,0.8250) (100,0.8250) (120,0.7750) (140,0.8000) (160,0.8000) (180,0.8250) (200,0.8250) (212,0.8000)};
\addplot[cOneSevenB, line width=1.15pt, line join=round, mark=*, mark size=.95pt, forget plot] coordinates {(0,0.3500) (20,0.5000) (40,0.5500) (60,0.4250) (80,0.4750) (100,0.4750) (120,0.5250) (140,0.5250) (142,0.5000)};
\addplot[cZeroSixB, line width=1.15pt, line join=round, mark=*, mark size=.95pt, forget plot] coordinates {(0,0.1750) (20,0.2750) (40,0.3250) (60,0.3750) (80,0.3500) (100,0.3250) (120,0.2250) (140,0.3250) (142,0.3250)};
\end{groupplot}
\path (vk c1r4.south west) -- (vk c3r4.south east) node[midway, below=10pt, font=\footnotesize] {training step};
\end{tikzpicture}
\caption{\textbf{Thinking or not, distillation converges early.} Lenient
$p@1$ and $p@8$ on validation problems held out from training, AIME 2025 (30 problems) and AMC23 (40), 8 samples each,
16{,}384-token limit, every 20 steps; ticks at 142 and 212 mark the evaluated steps. After the shaded first steps no curve rises clearly. In thinking mode from the external-SFT start the curves dip early, when the students stop
closing their reasoning (Section~\ref{sec:stopping}). Thin: a second run of the same setting; dashed: non-thinking
runs with an easy-to-hard curriculum (99 steps).}
\label{fig:valk}
\end{figure}
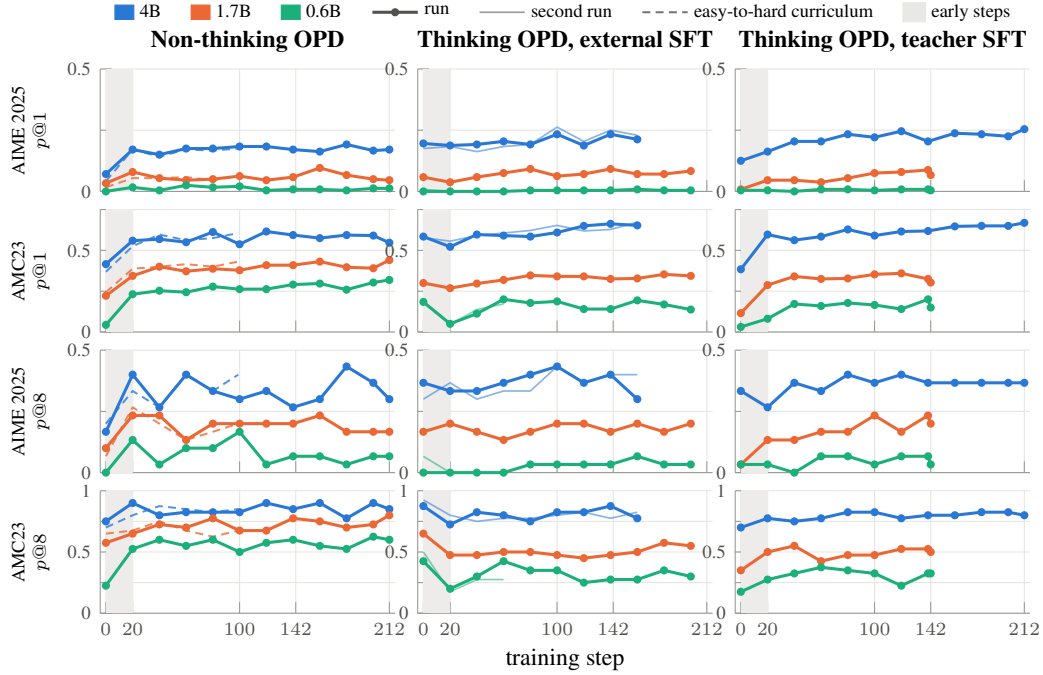

%% file: tables/lean_runs.tex
\begin{table}[h]
\centering
\footnotesize
\caption{\textbf{Training runs used in this paper.} All runs distill Qwen3-8B on the same 13{,}597 problems; the
student runs in the teacher's mode. \emph{Steps}: the last training step whose responses were stored. \emph{Evaluated}: the
step whose checkpoint was evaluated on the benchmarks. Runs 1 and 2 of one configuration differ only by being run twice
over the same data order.}
\label{tab:app-runs}
\setlength{\tabcolsep}{3pt}
\begin{tabular}{@{}lllccl@{}}
\toprule
Student start & Teacher & Budget & Steps & Evaluated & Used in \\
\midrule
\multicolumn{6}{@{}l}{\textit{Non-thinking route}} \\
\quad 4B pretrained-only & non-thinking & 7,168 & 212 & 212 & Section 3 \\
\quad 1.7B pretrained-only & non-thinking & 7,168 & 212 & 212 & Section 3 \\
\quad 0.6B pretrained-only & non-thinking & 7,168 & 212 & 212 & Section 3 \\
\quad 4B pretrained-only, curriculum & non-thinking & 7,168 & 99 & -- & Figure 1 \\
\quad 1.7B pretrained-only, curriculum & non-thinking & 7,168 & 99 & -- & Figure 1 \\
\addlinespace[2pt]
\multicolumn{6}{@{}l}{\textit{Thinking mode}} \\
\quad 4B external-SFT start, run 1 & thinking & 7,168 & 170 & -- & Section 4 \\
\quad 4B external-SFT start, run 2 & thinking & 7,168 & 167 & 142 & Sections 3--5 \\
\quad 1.7B external-SFT start & thinking & 7,168 & 212 & 142 & Sections 3--5 \\
\quad 0.6B external-SFT start, run 1 & thinking & 7,168 & 80 & -- & Section 4 \\
\quad 0.6B external-SFT start, run 2 & thinking & 7,168 & 212 & 142 & Sections 3--5 \\
\quad 4B teacher-SFT start & thinking & 7,168 & 212 & 142 & Sections 3--5 \\
\quad 1.7B teacher-SFT start & thinking & 7,168 & 142 & 142 & Sections 3--5 \\
\quad 0.6B teacher-SFT start & thinking & 7,168 & 142 & 142 & Sections 3--5 \\
\addlinespace[2pt]
\multicolumn{6}{@{}l}{\textit{Controls}} \\
\quad 0.6B external-SFT start & thinking & 16,384 & 7 & -- & budget, Section 4 \\
\quad 0.6B external-SFT start & thinking & 1,024 & 212 & -- & budget, Section 4 \\
\quad officially post-trained 0.6B, run 1 & thinking & 16,384 & 20 & -- & start, Section 4 \\
\quad officially post-trained 0.6B, run 2 & thinking & 16,384 & 20 & 20 & Sections 4--5, probes \\
\bottomrule
\end{tabular}
\end{table}

%% file: tables/lean_hparams.tex
\begin{table}[h]
\centering
\footnotesize
\caption{\textbf{Settings.} Distillation, the supervised cold starts, evaluation and in-training validation.}
\label{tab:app-hparams}
\setlength{\tabcolsep}{5pt}
\begin{tabularx}{\linewidth}{@{}l>{\raggedright\arraybackslash}X@{}}
\toprule
\multicolumn{2}{@{}l}{\textit{On-policy distillation (verl)}} \\
\quad prompts per step; updates per step & 128; 4 mini-batches of 32 \\
\quad optimizer & AdamW, learning rate $3\times10^{-6}$, 10 warm-up steps, then constant; weight decay 0.01; gradient clip 1.0 \\
\quad advantage & clipped per-token reverse KL, Eq.~(1), clip at $\pm 1$; no task reward \\
\quad loss aggregation & sum over tokens, mean over sequences, divided by a constant of 7{,}168 \\
\quad rollouts & 1 per prompt, temperature 1.0, budget 7{,}168 tokens (controls: 1{,}024 and 16{,}384) \\
\quad length & 2 epochs $=$ 212 steps; thinking mode evaluated at step 142 \\
\addlinespace[2pt]
\multicolumn{2}{@{}l}{\textit{Supervised cold start (same recipe at every size)}} \\
\quad data & 5{,}878 short external solutions, or 1{,}524 of the teacher's own correct solutions \\
\quad optimization & batch 16, learning rate $2\times10^{-5}$, cosine to $0.1\times$, warm-up ratio 0.01, 2 epochs \\
\quad maximum length & 4{,}096 tokens (short solutions), 10{,}240 (teacher solutions) \\
\addlinespace[2pt]
\multicolumn{2}{@{}l}{\textit{Evaluation}} \\
\quad benchmarks (problems, samples, token limit) & AIME 2026 (30, 48, 30{,}720); AMC23 (40, 48, 30{,}720); GSM8K-200 (200, 32, 8{,}192) \\
\quad non-thinking route only & GPQA-Diamond (198, 8, 16{,}384) \\
\quad sampling & temperature 0.6, top-$p$ 0.95, top-$k$ 20 \\
\quad in-training validation & every 20 steps; AIME 2025 and AMC23, 8 samples, 16{,}384 tokens, temperature 0.6 \\
\addlinespace[2pt]
\multicolumn{2}{@{}l}{\textit{Compute}} \\
\quad hardware & 3 nodes of 4 A100-40GB: the teacher on one node (vLLM), the student on two (FSDP) \\
\bottomrule
\end{tabularx}
\end{table}

%% file: appendix/B_scoring.tex
\section{Scoring and statistics}
\label{app:scoring}

\paragraph{Marked answers.}
A marked answer is the content of a \verb|\boxed{}| or of an \texttt{Answer:} / \texttt{final answer is} line. It is
compared with the reference after LaTeX normalization and, for numbers, by numeric value. The same matcher scores
training rewards, evaluation and every analysis in this paper.

\paragraph{Three scorings.}
\emph{Lenient}: at least one marked answer in the sample is correct, wherever it appears. \emph{Strict}: the sample also
finished, by closing its reasoning (thinking mode) or by stopping before the token limit (non-thinking mode), and the
answer after the close is correct. \emph{Committed}: the last marked \verb|\boxed{}| is correct; majority votes count one
committed answer per sample. A sample with no marked answer is wrong under all three.

\paragraph{Samples with several marked answers.}
Lenient scoring would reward a sample that lists several different answers. This credit is at most 0.025 of the
three-benchmark mean for every model except the 1.7B and 0.6B distilled from the teacher-SFT start, which mark
over a hundred answers per sample (Section~\ref{sec:unfinished}); removing it moves their lenient p@1 from 0.394 to 0.256
and from 0.262 to 0.231, close to their committed scores of 0.254 and 0.225.

\paragraph{The correct-value diagnostic.}
\label{app:scoring-value}
In Section~\ref{sec:unfinished} we ask whether the correct value appears anywhere in a sample that did not finish, marked
or not. Small numbers occur in almost any derivation, so we use only problems whose reference answer is an integer of
at least three digits that does not occur in the problem statement, and we report beside it how often the answer of a
different such problem appears in the same samples. It is not part of any score.

\paragraph{GPQA-Diamond.}
GPQA-Diamond has four options, so a sample that names several options would collect lenient credit by listing them. We
therefore evaluate it only for the non-thinking route, score it by the committed answer (strict beside it), and never
average it with the other benchmarks (Table~\ref{tab:app-gpqa}). We read the leading option letter of each marked answer, so that forms such as \verb|\boxed{\text{A}}| are credited.

\paragraph{Intervals.}
Every interval is a paired bootstrap over problems with 10,000 resamples: within each benchmark, problems are drawn with
replacement, the same draw is used for both models of a comparison, the metric is computed per benchmark and the
benchmarks are averaged without weights, as in the text. We report 95\% percentile intervals. We read a difference whose
interval contains zero as no difference, and call a change of about 0.1 or less small even when its interval excludes
zero.

%% file: appendix/C_ceiling.tex
\section{Supplementary results: how far solving rises}
\label{app:block1}

\paragraph{Scores per benchmark.}
Table~\ref{tab:app-results} gives lenient p@1 and p@n on every benchmark for the teacher and for every starting and
distilled student; Table~\ref{tab:results} in the main text gives their three-benchmark means.

\input{tables/tab_results_full}

\paragraph{Pass@$k$ and the problems behind the gain.}
Figure~\ref{fig:passk} shows, for every route and size, how often the starting and the distilled student solve a
problem at least once in $k$ samples; Table~\ref{tab:app-passk} gives the values and the $k^\star$ of every distilled
student, and Table~\ref{tab:app-bins} the numbers behind Figure~\ref{fig:bins}. On AMC23 and GSM8K-200 the two curves
close in as $k$ grows for every student; on AIME 2026, the hardest benchmark, the distilled curve pulls clearly ahead
for the 4B in thinking mode and falls behind at large $k$ for the 0.6B from the external-SFT start. The curves also give an exchange rate: one distilled attempt is worth the start's best of 4
attempts at 4B and of only 2 at 1.7B and 0.6B from the external-SFT start, and of 8 to 15 only where the start rarely
marked an answer, as the pretrained-only 0.6B (0.194 of its samples) and the teacher-SFT starts (0.366, 0.151 and
0.089) did.

\input{figures/fig_passk}

\input{tables/lean_passk}

\input{tables/lean_bins}

\paragraph{The ceiling per benchmark.}
No distilled student's p@1 exceeds its starting p@n on any benchmark (Table~\ref{tab:app-ceiling}a). The four cases
whose interval reaches zero are all on AIME 2026: the 4B from both thinking starts, and the 1.7B and 0.6B from the
teacher-SFT start, where both scores are close to zero. The starting p@n itself moves by 0.2 or more only for the 4B
in thinking mode on AIME 2026 and for the pretrained-only 0.6B on GSM8K-200, which before distillation often gave no answer
at all (Table~\ref{tab:app-ceiling}b).

\input{tables/lean_ceiling}

\paragraph{What distillation changes.}
Table~\ref{tab:app-decomp} splits each route's change into the marked-answer rate and the accuracy of the marked
answers. Accuracy rises on every route at every size. The marked-answer rate rises on the non-thinking route and from
the teacher-SFT start, whose students rarely marked an answer before distillation, and falls from the
external-SFT start at 1.7B and 0.6B.

\input{tables/lean_decomp}

\paragraph{GPQA-Diamond.}
On the non-thinking route every pretrained-only start is below chance, because it often runs to the limit without an
answer. Distillation raises all three sizes, but only the distilled 4B ends above chance (Table~\ref{tab:app-gpqa}).

\input{tables/lean_gpqa}

\paragraph{Majority vote.}
A vote over the $n$ committed answers \citep{wang2023selfconsistency} tells whether a student's samples agree on the
right answer. Distillation raises the 4B's vote from 0.656 to 0.792 from the external-SFT start. At 1.7B the thinking
routes tie the non-thinking route under the vote (0.511 and 0.548 against 0.519), although they fall below it one sample
at a time; at 0.6B the vote does not change with distillation (0.400 to 0.398).

%% file: tables/tab_results_full.tex
\begin{table}[h]
\centering
\caption{\textbf{Lenient $p@1$ and $p@n$ on every benchmark.} Distillation raises every student's mean lenient $p@1$,
but on no benchmark past the $p@n$ of the student it started from. Lenient scoring; $p@n$ over $n=48$ samples (AIME 2026, AMC23) or 32
(GSM8K-200), every sample we drew; mean: unweighted over the three benchmarks. Gray: the student before distillation.
The two ceilings: Figures~\ref{fig:ownceil} and~\ref{fig:teachceil}; paired intervals: Table~\ref{tab:app-ceiling}.}
\label{tab:app-results}
\footnotesize
\setlength{\tabcolsep}{5.2pt}
\renewcommand{\arraystretch}{0.95}
\begin{tabular}{@{}ll*{8}{c}@{}}
\toprule
 & & \multicolumn{2}{c}{AIME 2026} & \multicolumn{2}{c}{AMC23} & \multicolumn{2}{c}{GSM8K-200} & \multicolumn{2}{c}{Mean} \\
\cmidrule(lr){3-4}\cmidrule(lr){5-6}\cmidrule(lr){7-8}\cmidrule(l){9-10}
Model & & $p@1$ & $p@48$ & $p@1$ & $p@48$ & $p@1$ & $p@32$ & $p@1$ & $p@n$ \\
\midrule
Teacher (8B) & thinking & 0.674 & 0.900 & 0.960 & 1.000 & 0.937 & 0.965 & 0.857 & 0.955 \\
 & non-thinking & 0.163 & 0.667 & 0.694 & 0.975 & 0.929 & 0.965 & 0.595 & 0.869 \\
\addlinespace[3pt]
\multicolumn{10}{@{}l}{\textit{Non-thinking route: from the pretrained-only model, non-thinking teacher}} \\
\sizedot{cFourB}\,4B & start & \textcolor{cMuted}{0.083} & \textcolor{cMuted}{0.333} & \textcolor{cMuted}{0.438} & \textcolor{cMuted}{0.900} & \textcolor{cMuted}{0.761} & \textcolor{cMuted}{0.975} & \textcolor{cMuted}{0.427} & \textcolor{cMuted}{0.736} \\
 & after & 0.141 & 0.467 & 0.587 & 0.925 & 0.919 & 0.980 & 0.549 & 0.791 \\
\sizedot{cOneSevenB}\,1.7B & start & \textcolor{cMuted}{0.014} & \textcolor{cMuted}{0.133} & \textcolor{cMuted}{0.266} & \textcolor{cMuted}{0.825} & \textcolor{cMuted}{0.601} & \textcolor{cMuted}{0.980} & \textcolor{cMuted}{0.294} & \textcolor{cMuted}{0.646} \\
 & after & 0.079 & 0.300 & 0.410 & 0.900 & 0.830 & 0.975 & 0.440 & 0.725 \\
\sizedot{cZeroSixB}\,0.6B & start & \textcolor{cMuted}{0.004} & \textcolor{cMuted}{0.167} & \textcolor{cMuted}{0.053} & \textcolor{cMuted}{0.625} & \textcolor{cMuted}{0.087} & \textcolor{cMuted}{0.745} & \textcolor{cMuted}{0.048} & \textcolor{cMuted}{0.512} \\
 & after & 0.018 & 0.133 & 0.278 & 0.700 & 0.689 & 0.945 & 0.328 & 0.593 \\
\addlinespace[3pt]
\multicolumn{10}{@{}l}{\textit{Thinking mode, from the external-SFT start, thinking teacher}} \\
\sizedot{cFourB}\,4B & start & \textcolor{cMuted}{0.148} & \textcolor{cMuted}{0.400} & \textcolor{cMuted}{0.548} & \textcolor{cMuted}{0.925} & \textcolor{cMuted}{0.899} & \textcolor{cMuted}{0.980} & \textcolor{cMuted}{0.532} & \textcolor{cMuted}{0.768} \\
 & after & 0.329 & 0.700 & 0.728 & 0.900 & 0.896 & 0.965 & 0.651 & 0.855 \\
\sizedot{cOneSevenB}\,1.7B & start & \textcolor{cMuted}{0.029} & \textcolor{cMuted}{0.233} & \textcolor{cMuted}{0.299} & \textcolor{cMuted}{0.775} & \textcolor{cMuted}{0.747} & \textcolor{cMuted}{0.985} & \textcolor{cMuted}{0.358} & \textcolor{cMuted}{0.664} \\
 & after & 0.057 & 0.200 & 0.362 & 0.725 & 0.757 & 0.960 & 0.392 & 0.628 \\
\sizedot{cZeroSixB}\,0.6B & start & \textcolor{cMuted}{0.006} & \textcolor{cMuted}{0.167} & \textcolor{cMuted}{0.173} & \textcolor{cMuted}{0.675} & \textcolor{cMuted}{0.536} & \textcolor{cMuted}{0.930} & \textcolor{cMuted}{0.238} & \textcolor{cMuted}{0.591} \\
 & after & 0.004 & 0.067 & 0.212 & 0.600 & 0.667 & 0.895 & 0.294 & 0.521 \\
\addlinespace[3pt]
\multicolumn{10}{@{}l}{\textit{Thinking mode, from the teacher-SFT start, thinking teacher}} \\
\sizedot{cFourB}\,4B & start & \textcolor{cMuted}{0.116} & \textcolor{cMuted}{0.400} & \textcolor{cMuted}{0.412} & \textcolor{cMuted}{0.900} & \textcolor{cMuted}{0.461} & \textcolor{cMuted}{0.945} & \textcolor{cMuted}{0.330} & \textcolor{cMuted}{0.748} \\
 & after & 0.316 & 0.733 & 0.728 & 0.950 & 0.903 & 0.965 & 0.649 & 0.883 \\
\sizedot{cOneSevenB}\,1.7B & start & \textcolor{cMuted}{0.013} & \textcolor{cMuted}{0.100} & \textcolor{cMuted}{0.129} & \textcolor{cMuted}{0.550} & \textcolor{cMuted}{0.243} & \textcolor{cMuted}{0.900} & \textcolor{cMuted}{0.128} & \textcolor{cMuted}{0.517} \\
 & after & 0.067 & 0.233 & 0.372 & 0.675 & 0.743 & 0.950 & 0.394 & 0.619 \\
\sizedot{cZeroSixB}\,0.6B & start & \textcolor{cMuted}{0.001} & \textcolor{cMuted}{0.033} & \textcolor{cMuted}{0.029} & \textcolor{cMuted}{0.400} & \textcolor{cMuted}{0.135} & \textcolor{cMuted}{0.765} & \textcolor{cMuted}{0.055} & \textcolor{cMuted}{0.399} \\
 & after & 0.003 & 0.067 & 0.165 & 0.475 & 0.620 & 0.880 & 0.262 & 0.474 \\
\bottomrule
\end{tabular}
\end{table}

%% file: figures/fig_passk.tex
\begin{figure}[h]
\centering
\begin{tikzpicture}
\begin{groupplot}[group style={group size=3 by 3, horizontal sep=0.30cm, vertical sep=0.26cm,
    xticklabels at=edge bottom, yticklabels at=edge left, ylabels at=edge left, group name=pk},
  paperstyle, scale only axis, width=3.90cm, height=2.10cm,
  xmode=log, log basis x=2, xmin=1, xmax=48, ymin=0, ymax=1, enlargelimits=false, clip=false,
  xtick={1,2,4,8,16,32,48}, xticklabels={1,2,4,8,16,32,48}, ytick={0,0.25,0.5,0.75,1},
  yticklabels={0,,0.5,,1}, minor tick num=0, axis x line*=bottom, axis y line*=left,
  title style={yshift=-3pt}, ylabel style={align=center, font=\scriptsize}]
\nextgroupplot[title={\textbf{Non-thinking OPD}}, ylabel={AIME 2026\\lenient pass@$k$}]
\addplot[cTeacher!75, line width=.7pt, densely dotted, forget plot] coordinates {(1,0.1625) (2,0.2179) (3,0.2578) (4,0.2902) (5,0.3173) (6,0.3406) (7,0.3609) (8,0.3789) (9,0.3951) (10,0.4098) (11,0.4232) (12,0.4356) (13,0.4471) (14,0.4578) (15,0.4679) (16,0.4775) (17,0.4865) (18,0.4951) (19,0.5033) (20,0.5112) (21,0.5188) (22,0.5261) (23,0.5332) (24,0.5401) (25,0.5468) (26,0.5533) (27,0.5596) (28,0.5658) (29,0.5718) (30,0.5777) (31,0.5835) (32,0.5892) (33,0.5947) (34,0.6002) (35,0.6055) (36,0.6108) (37,0.6159) (38,0.6210) (39,0.6260) (40,0.6308) (41,0.6356) (42,0.6403) (43,0.6449) (44,0.6495) (45,0.6539) (46,0.6582) (47,0.6625) (48,0.6667)};
\addplot[cFourB, line width=.85pt, densely dashed, forget plot] coordinates {(1,0.0826) (2,0.1248) (3,0.1516) (4,0.1716) (5,0.1878) (6,0.2015) (7,0.2134) (8,0.2239) (9,0.2332) (10,0.2415) (11,0.2489) (12,0.2556) (13,0.2617) (14,0.2672) (15,0.2722) (16,0.2768) (17,0.2809) (18,0.2848) (19,0.2883) (20,0.2916) (21,0.2946) (22,0.2974) (23,0.3000) (24,0.3024) (25,0.3047) (26,0.3068) (27,0.3088) (28,0.3107) (29,0.3125) (30,0.3142) (31,0.3158) (32,0.3173) (33,0.3187) (34,0.3200) (35,0.3213) (36,0.3225) (37,0.3237) (38,0.3248) (39,0.3258) (40,0.3268) (41,0.3278) (42,0.3287) (43,0.3295) (44,0.3304) (45,0.3312) (46,0.3319) (47,0.3326) (48,0.3333)};
\addplot[cFourB, line width=1.25pt, forget plot] coordinates {(1,0.1410) (2,0.1811) (3,0.2096) (4,0.2320) (5,0.2504) (6,0.2660) (7,0.2795) (8,0.2913) (9,0.3018) (10,0.3113) (11,0.3199) (12,0.3278) (13,0.3350) (14,0.3417) (15,0.3479) (16,0.3537) (17,0.3591) (18,0.3642) (19,0.3691) (20,0.3737) (21,0.3781) (22,0.3823) (23,0.3864) (24,0.3903) (25,0.3942) (26,0.3979) (27,0.4015) (28,0.4050) (29,0.4085) (30,0.4119) (31,0.4153) (32,0.4186) (33,0.4218) (34,0.4250) (35,0.4282) (36,0.4314) (37,0.4345) (38,0.4376) (39,0.4406) (40,0.4436) (41,0.4466) (42,0.4496) (43,0.4525) (44,0.4554) (45,0.4582) (46,0.4611) (47,0.4639) (48,0.4667)};
\addplot[cOneSevenB, line width=.85pt, densely dashed, forget plot] coordinates {(1,0.0139) (2,0.0258) (3,0.0360) (4,0.0448) (5,0.0525) (6,0.0592) (7,0.0651) (8,0.0703) (9,0.0750) (10,0.0791) (11,0.0829) (12,0.0863) (13,0.0895) (14,0.0924) (15,0.0950) (16,0.0974) (17,0.0997) (18,0.1018) (19,0.1038) (20,0.1056) (21,0.1074) (22,0.1090) (23,0.1105) (24,0.1119) (25,0.1133) (26,0.1146) (27,0.1158) (28,0.1169) (29,0.1180) (30,0.1191) (31,0.1201) (32,0.1211) (33,0.1220) (34,0.1229) (35,0.1237) (36,0.1246) (37,0.1254) (38,0.1262) (39,0.1269) (40,0.1277) (41,0.1284) (42,0.1291) (43,0.1298) (44,0.1305) (45,0.1312) (46,0.1319) (47,0.1326) (48,0.1333)};
\addplot[cOneSevenB, line width=1.25pt, forget plot] coordinates {(1,0.0792) (2,0.1108) (3,0.1272) (4,0.1381) (5,0.1466) (6,0.1538) (7,0.1603) (8,0.1662) (9,0.1717) (10,0.1769) (11,0.1818) (12,0.1865) (13,0.1910) (14,0.1953) (15,0.1996) (16,0.2037) (17,0.2077) (18,0.2116) (19,0.2155) (20,0.2193) (21,0.2230) (22,0.2266) (23,0.2302) (24,0.2337) (25,0.2371) (26,0.2405) (27,0.2438) (28,0.2471) (29,0.2503) (30,0.2535) (31,0.2565) (32,0.2596) (33,0.2625) (34,0.2655) (35,0.2683) (36,0.2711) (37,0.2738) (38,0.2765) (39,0.2791) (40,0.2817) (41,0.2842) (42,0.2866) (43,0.2890) (44,0.2913) (45,0.2936) (46,0.2958) (47,0.2979) (48,0.3000)};
\addplot[cZeroSixB, line width=.85pt, densely dashed, forget plot] coordinates {(1,0.0042) (2,0.0083) (3,0.0124) (4,0.0165) (5,0.0205) (6,0.0246) (7,0.0285) (8,0.0325) (9,0.0364) (10,0.0403) (11,0.0442) (12,0.0480) (13,0.0519) (14,0.0556) (15,0.0594) (16,0.0631) (17,0.0668) (18,0.0705) (19,0.0741) (20,0.0777) (21,0.0813) (22,0.0848) (23,0.0884) (24,0.0918) (25,0.0953) (26,0.0987) (27,0.1021) (28,0.1055) (29,0.1088) (30,0.1121) (31,0.1154) (32,0.1187) (33,0.1219) (34,0.1251) (35,0.1283) (36,0.1314) (37,0.1345) (38,0.1376) (39,0.1406) (40,0.1436) (41,0.1466) (42,0.1496) (43,0.1525) (44,0.1554) (45,0.1582) (46,0.1611) (47,0.1639) (48,0.1667)};
\addplot[cZeroSixB, line width=1.25pt, forget plot] coordinates {(1,0.0181) (2,0.0323) (3,0.0437) (4,0.0528) (5,0.0602) (6,0.0662) (7,0.0712) (8,0.0755) (9,0.0791) (10,0.0823) (11,0.0851) (12,0.0876) (13,0.0899) (14,0.0920) (15,0.0940) (16,0.0959) (17,0.0976) (18,0.0994) (19,0.1010) (20,0.1026) (21,0.1041) (22,0.1056) (23,0.1071) (24,0.1085) (25,0.1099) (26,0.1112) (27,0.1125) (28,0.1138) (29,0.1151) (30,0.1163) (31,0.1175) (32,0.1187) (33,0.1198) (34,0.1209) (35,0.1220) (36,0.1230) (37,0.1241) (38,0.1251) (39,0.1260) (40,0.1270) (41,0.1279) (42,0.1287) (43,0.1296) (44,0.1304) (45,0.1312) (46,0.1319) (47,0.1326) (48,0.1333)};
\nextgroupplot[title={\textbf{Thinking OPD, external SFT}}, legend style={at={(0.5,1)}, anchor=south, yshift=13pt, legend columns=6, /tikz/every even column/.append style={column sep=0.9em}}]
\addlegendimage{legend image code/.code={\fill[cFourB] (0cm,-0.6ex) rectangle (0.25cm,0.6ex);}}
\addlegendentry{4B}
\addlegendimage{legend image code/.code={\fill[cOneSevenB] (0cm,-0.6ex) rectangle (0.25cm,0.6ex);}}
\addlegendentry{1.7B}
\addlegendimage{legend image code/.code={\fill[cZeroSixB] (0cm,-0.6ex) rectangle (0.25cm,0.6ex);}}
\addlegendentry{0.6B}
\addlegendimage{cMuted, line width=.85pt, densely dashed}
\addlegendentry{start}
\addlegendimage{cMuted, line width=1.25pt}
\addlegendentry{after distillation}
\addlegendimage{cTeacher!75, line width=.7pt, densely dotted}
\addlegendentry{mode-matched teacher}
\addplot[cTeacher!75, line width=.7pt, densely dotted, forget plot] coordinates {(1,0.6743) (2,0.7619) (3,0.7995) (4,0.8212) (5,0.8356) (6,0.8460) (7,0.8537) (8,0.8598) (9,0.8647) (10,0.8686) (11,0.8719) (12,0.8747) (13,0.8771) (14,0.8792) (15,0.8810) (16,0.8826) (17,0.8841) (18,0.8854) (19,0.8867) (20,0.8878) (21,0.8888) (22,0.8898) (23,0.8906) (24,0.8915) (25,0.8922) (26,0.8930) (27,0.8936) (28,0.8943) (29,0.8949) (30,0.8954) (31,0.8959) (32,0.8964) (33,0.8969) (34,0.8973) (35,0.8977) (36,0.8980) (37,0.8984) (38,0.8987) (39,0.8989) (40,0.8992) (41,0.8994) (42,0.8996) (43,0.8997) (44,0.8998) (45,0.8999) (46,0.9000) (47,0.9000) (48,0.9000)};
\addplot[cFourB, line width=.85pt, densely dashed, forget plot] coordinates {(1,0.1479) (2,0.2007) (3,0.2286) (4,0.2476) (5,0.2620) (6,0.2735) (7,0.2830) (8,0.2911) (9,0.2982) (10,0.3045) (11,0.3101) (12,0.3152) (13,0.3199) (14,0.3243) (15,0.3283) (16,0.3322) (17,0.3357) (18,0.3391) (19,0.3423) (20,0.3454) (21,0.3483) (22,0.3511) (23,0.3538) (24,0.3563) (25,0.3588) (26,0.3612) (27,0.3635) (28,0.3657) (29,0.3678) (30,0.3699) (31,0.3719) (32,0.3739) (33,0.3758) (34,0.3777) (35,0.3795) (36,0.3813) (37,0.3830) (38,0.3847) (39,0.3864) (40,0.3880) (41,0.3897) (42,0.3912) (43,0.3928) (44,0.3943) (45,0.3957) (46,0.3972) (47,0.3986) (48,0.4000)};
\addplot[cFourB, line width=1.25pt, forget plot] coordinates {(1,0.3292) (2,0.3972) (3,0.4292) (4,0.4506) (5,0.4676) (6,0.4822) (7,0.4951) (8,0.5068) (9,0.5175) (10,0.5273) (11,0.5364) (12,0.5448) (13,0.5528) (14,0.5602) (15,0.5673) (16,0.5739) (17,0.5803) (18,0.5863) (19,0.5921) (20,0.5976) (21,0.6029) (22,0.6080) (23,0.6129) (24,0.6176) (25,0.6222) (26,0.6266) (27,0.6308) (28,0.6350) (29,0.6390) (30,0.6429) (31,0.6466) (32,0.6503) (33,0.6539) (34,0.6574) (35,0.6608) (36,0.6641) (37,0.6674) (38,0.6706) (39,0.6737) (40,0.6768) (41,0.6799) (42,0.6828) (43,0.6858) (44,0.6887) (45,0.6916) (46,0.6944) (47,0.6972) (48,0.7000)};
\addplot[cOneSevenB, line width=.85pt, densely dashed, forget plot] coordinates {(1,0.0292) (2,0.0513) (3,0.0683) (4,0.0818) (5,0.0928) (6,0.1019) (7,0.1097) (8,0.1166) (9,0.1227) (10,0.1283) (11,0.1336) (12,0.1384) (13,0.1430) (14,0.1474) (15,0.1516) (16,0.1556) (17,0.1595) (18,0.1632) (19,0.1667) (20,0.1702) (21,0.1735) (22,0.1767) (23,0.1798) (24,0.1829) (25,0.1858) (26,0.1886) (27,0.1914) (28,0.1940) (29,0.1966) (30,0.1991) (31,0.2016) (32,0.2039) (33,0.2062) (34,0.2085) (35,0.2106) (36,0.2128) (37,0.2148) (38,0.2168) (39,0.2187) (40,0.2206) (41,0.2224) (42,0.2241) (43,0.2258) (44,0.2274) (45,0.2290) (46,0.2305) (47,0.2319) (48,0.2333)};
\addplot[cOneSevenB, line width=1.25pt, forget plot] coordinates {(1,0.0569) (2,0.0821) (3,0.0963) (4,0.1058) (5,0.1129) (6,0.1185) (7,0.1231) (8,0.1271) (9,0.1307) (10,0.1339) (11,0.1369) (12,0.1396) (13,0.1423) (14,0.1447) (15,0.1471) (16,0.1494) (17,0.1515) (18,0.1536) (19,0.1556) (20,0.1576) (21,0.1595) (22,0.1613) (23,0.1631) (24,0.1648) (25,0.1665) (26,0.1682) (27,0.1698) (28,0.1714) (29,0.1729) (30,0.1745) (31,0.1760) (32,0.1775) (33,0.1789) (34,0.1804) (35,0.1818) (36,0.1832) (37,0.1847) (38,0.1861) (39,0.1875) (40,0.1889) (41,0.1903) (42,0.1917) (43,0.1931) (44,0.1944) (45,0.1958) (46,0.1972) (47,0.1986) (48,0.2000)};
\addplot[cZeroSixB, line width=.85pt, densely dashed, forget plot] coordinates {(1,0.0056) (2,0.0109) (3,0.0161) (4,0.0212) (5,0.0261) (6,0.0308) (7,0.0354) (8,0.0399) (9,0.0442) (10,0.0485) (11,0.0526) (12,0.0566) (13,0.0605) (14,0.0643) (15,0.0680) (16,0.0716) (17,0.0752) (18,0.0786) (19,0.0820) (20,0.0854) (21,0.0887) (22,0.0919) (23,0.0951) (24,0.0982) (25,0.1013) (26,0.1043) (27,0.1073) (28,0.1103) (29,0.1132) (30,0.1161) (31,0.1190) (32,0.1219) (33,0.1248) (34,0.1276) (35,0.1304) (36,0.1332) (37,0.1361) (38,0.1389) (39,0.1416) (40,0.1444) (41,0.1472) (42,0.1500) (43,0.1528) (44,0.1556) (45,0.1583) (46,0.1611) (47,0.1639) (48,0.1667)};
\addplot[cZeroSixB, line width=1.25pt, forget plot] coordinates {(1,0.0035) (2,0.0068) (3,0.0099) (4,0.0129) (5,0.0157) (6,0.0183) (7,0.0208) (8,0.0232) (9,0.0255) (10,0.0276) (11,0.0297) (12,0.0316) (13,0.0334) (14,0.0351) (15,0.0367) (16,0.0383) (17,0.0397) (18,0.0411) (19,0.0425) (20,0.0437) (21,0.0449) (22,0.0461) (23,0.0471) (24,0.0482) (25,0.0492) (26,0.0501) (27,0.0511) (28,0.0519) (29,0.0528) (30,0.0536) (31,0.0545) (32,0.0552) (33,0.0560) (34,0.0568) (35,0.0575) (36,0.0582) (37,0.0590) (38,0.0597) (39,0.0604) (40,0.0611) (41,0.0618) (42,0.0625) (43,0.0632) (44,0.0639) (45,0.0646) (46,0.0653) (47,0.0660) (48,0.0667)};
\nextgroupplot[title={\textbf{Thinking OPD, teacher SFT}}]
\addplot[cTeacher!75, line width=.7pt, densely dotted, forget plot] coordinates {(1,0.6743) (2,0.7619) (3,0.7995) (4,0.8212) (5,0.8356) (6,0.8460) (7,0.8537) (8,0.8598) (9,0.8647) (10,0.8686) (11,0.8719) (12,0.8747) (13,0.8771) (14,0.8792) (15,0.8810) (16,0.8826) (17,0.8841) (18,0.8854) (19,0.8867) (20,0.8878) (21,0.8888) (22,0.8898) (23,0.8906) (24,0.8915) (25,0.8922) (26,0.8930) (27,0.8936) (28,0.8943) (29,0.8949) (30,0.8954) (31,0.8959) (32,0.8964) (33,0.8969) (34,0.8973) (35,0.8977) (36,0.8980) (37,0.8984) (38,0.8987) (39,0.8989) (40,0.8992) (41,0.8994) (42,0.8996) (43,0.8997) (44,0.8998) (45,0.8999) (46,0.9000) (47,0.9000) (48,0.9000)};
\addplot[cFourB, line width=.85pt, densely dashed, forget plot] coordinates {(1,0.1160) (2,0.1723) (3,0.2060) (4,0.2296) (5,0.2480) (6,0.2630) (7,0.2756) (8,0.2865) (9,0.2960) (10,0.3043) (11,0.3116) (12,0.3182) (13,0.3241) (14,0.3295) (15,0.3344) (16,0.3388) (17,0.3429) (18,0.3467) (19,0.3503) (20,0.3536) (21,0.3566) (22,0.3595) (23,0.3623) (24,0.3649) (25,0.3673) (26,0.3697) (27,0.3719) (28,0.3740) (29,0.3760) (30,0.3779) (31,0.3797) (32,0.3815) (33,0.3831) (34,0.3847) (35,0.3862) (36,0.3877) (37,0.3891) (38,0.3904) (39,0.3916) (40,0.3928) (41,0.3939) (42,0.3949) (43,0.3959) (44,0.3969) (45,0.3977) (46,0.3986) (47,0.3993) (48,0.4000)};
\addplot[cFourB, line width=1.25pt, forget plot] coordinates {(1,0.3160) (2,0.3979) (3,0.4445) (4,0.4770) (5,0.5022) (6,0.5230) (7,0.5408) (8,0.5565) (9,0.5703) (10,0.5828) (11,0.5941) (12,0.6043) (13,0.6136) (14,0.6221) (15,0.6299) (16,0.6371) (17,0.6438) (18,0.6499) (19,0.6557) (20,0.6610) (21,0.6659) (22,0.6706) (23,0.6749) (24,0.6790) (25,0.6828) (26,0.6864) (27,0.6898) (28,0.6930) (29,0.6961) (30,0.6989) (31,0.7017) (32,0.7043) (33,0.7067) (34,0.7091) (35,0.7113) (36,0.7134) (37,0.7155) (38,0.7174) (39,0.7193) (40,0.7211) (41,0.7228) (42,0.7244) (43,0.7260) (44,0.7276) (45,0.7291) (46,0.7305) (47,0.7319) (48,0.7333)};
\addplot[cOneSevenB, line width=.85pt, densely dashed, forget plot] coordinates {(1,0.0132) (2,0.0249) (3,0.0351) (4,0.0442) (5,0.0521) (6,0.0590) (7,0.0651) (8,0.0703) (9,0.0749) (10,0.0788) (11,0.0822) (12,0.0851) (13,0.0876) (14,0.0897) (15,0.0915) (16,0.0930) (17,0.0943) (18,0.0954) (19,0.0963) (20,0.0970) (21,0.0976) (22,0.0981) (23,0.0985) (24,0.0989) (25,0.0991) (26,0.0993) (27,0.0995) (28,0.0996) (29,0.0997) (30,0.0998) (31,0.0999) (32,0.0999) (33,0.0999) (34,0.1000) (35,0.1000) (36,0.1000) (37,0.1000) (38,0.1000) (39,0.1000) (40,0.1000) (41,0.1000) (42,0.1000) (43,0.1000) (44,0.1000) (45,0.1000) (46,0.1000) (47,0.1000) (48,0.1000)};
\addplot[cOneSevenB, line width=1.25pt, forget plot] coordinates {(1,0.0674) (2,0.0994) (3,0.1185) (4,0.1320) (5,0.1425) (6,0.1512) (7,0.1585) (8,0.1648) (9,0.1702) (10,0.1750) (11,0.1792) (12,0.1830) (13,0.1864) (14,0.1896) (15,0.1924) (16,0.1951) (17,0.1976) (18,0.1999) (19,0.2020) (20,0.2040) (21,0.2059) (22,0.2077) (23,0.2094) (24,0.2109) (25,0.2124) (26,0.2138) (27,0.2152) (28,0.2164) (29,0.2176) (30,0.2187) (31,0.2198) (32,0.2208) (33,0.2218) (34,0.2227) (35,0.2236) (36,0.2245) (37,0.2253) (38,0.2261) (39,0.2269) (40,0.2277) (41,0.2284) (42,0.2291) (43,0.2298) (44,0.2305) (45,0.2312) (46,0.2319) (47,0.2326) (48,0.2333)};
\addplot[cZeroSixB, line width=.85pt, densely dashed, forget plot] coordinates {(1,0.0014) (2,0.0027) (3,0.0041) (4,0.0054) (5,0.0066) (6,0.0079) (7,0.0091) (8,0.0103) (9,0.0114) (10,0.0126) (11,0.0137) (12,0.0147) (13,0.0158) (14,0.0168) (15,0.0177) (16,0.0187) (17,0.0196) (18,0.0205) (19,0.0213) (20,0.0222) (21,0.0230) (22,0.0237) (23,0.0245) (24,0.0252) (25,0.0259) (26,0.0265) (27,0.0271) (28,0.0277) (29,0.0283) (30,0.0288) (31,0.0293) (32,0.0298) (33,0.0302) (34,0.0306) (35,0.0310) (36,0.0314) (37,0.0317) (38,0.0320) (39,0.0323) (40,0.0325) (41,0.0327) (42,0.0329) (43,0.0330) (44,0.0332) (45,0.0332) (46,0.0333) (47,0.0333) (48,0.0333)};
\addplot[cZeroSixB, line width=1.25pt, forget plot] coordinates {(1,0.0028) (2,0.0055) (3,0.0081) (4,0.0106) (5,0.0130) (6,0.0154) (7,0.0177) (8,0.0198) (9,0.0220) (10,0.0240) (11,0.0260) (12,0.0279) (13,0.0297) (14,0.0315) (15,0.0332) (16,0.0349) (17,0.0365) (18,0.0380) (19,0.0395) (20,0.0409) (21,0.0423) (22,0.0436) (23,0.0449) (24,0.0461) (25,0.0473) (26,0.0484) (27,0.0495) (28,0.0506) (29,0.0516) (30,0.0526) (31,0.0536) (32,0.0545) (33,0.0554) (34,0.0562) (35,0.0571) (36,0.0579) (37,0.0587) (38,0.0595) (39,0.0603) (40,0.0610) (41,0.0617) (42,0.0625) (43,0.0632) (44,0.0639) (45,0.0646) (46,0.0653) (47,0.0660) (48,0.0667)};
\nextgroupplot[ylabel={AMC23\\lenient pass@$k$}]
\addplot[cTeacher!75, line width=.7pt, densely dotted, forget plot] coordinates {(1,0.6937) (2,0.7896) (3,0.8362) (4,0.8644) (5,0.8834) (6,0.8970) (7,0.9074) (8,0.9154) (9,0.9220) (10,0.9273) (11,0.9318) (12,0.9357) (13,0.9390) (14,0.9419) (15,0.9444) (16,0.9467) (17,0.9487) (18,0.9506) (19,0.9523) (20,0.9538) (21,0.9552) (22,0.9565) (23,0.9577) (24,0.9588) (25,0.9599) (26,0.9609) (27,0.9618) (28,0.9627) (29,0.9635) (30,0.9643) (31,0.9651) (32,0.9658) (33,0.9665) (34,0.9672) (35,0.9678) (36,0.9684) (37,0.9690) (38,0.9696) (39,0.9702) (40,0.9708) (41,0.9713) (42,0.9718) (43,0.9724) (44,0.9729) (45,0.9734) (46,0.9740) (47,0.9745) (48,0.9750)};
\addplot[cFourB, line width=.85pt, densely dashed, forget plot] coordinates {(1,0.4380) (2,0.5652) (3,0.6308) (4,0.6742) (5,0.7061) (6,0.7309) (7,0.7506) (8,0.7668) (9,0.7802) (10,0.7914) (11,0.8010) (12,0.8092) (13,0.8163) (14,0.8226) (15,0.8281) (16,0.8331) (17,0.8376) (18,0.8416) (19,0.8454) (20,0.8488) (21,0.8520) (22,0.8550) (23,0.8578) (24,0.8605) (25,0.8630) (26,0.8654) (27,0.8677) (28,0.8698) (29,0.8719) (30,0.8739) (31,0.8759) (32,0.8777) (33,0.8795) (34,0.8812) (35,0.8829) (36,0.8845) (37,0.8861) (38,0.8876) (39,0.8890) (40,0.8904) (41,0.8918) (42,0.8931) (43,0.8943) (44,0.8956) (45,0.8967) (46,0.8979) (47,0.8990) (48,0.9000)};
\addplot[cFourB, line width=1.25pt, forget plot] coordinates {(1,0.5865) (2,0.6921) (3,0.7488) (4,0.7847) (5,0.8094) (6,0.8273) (7,0.8410) (8,0.8518) (9,0.8606) (10,0.8680) (11,0.8742) (12,0.8795) (13,0.8842) (14,0.8883) (15,0.8919) (16,0.8952) (17,0.8981) (18,0.9008) (19,0.9032) (20,0.9053) (21,0.9073) (22,0.9091) (23,0.9108) (24,0.9123) (25,0.9137) (26,0.9150) (27,0.9162) (28,0.9173) (29,0.9182) (30,0.9191) (31,0.9199) (32,0.9207) (33,0.9213) (34,0.9219) (35,0.9224) (36,0.9229) (37,0.9233) (38,0.9237) (39,0.9240) (40,0.9242) (41,0.9244) (42,0.9246) (43,0.9247) (44,0.9249) (45,0.9249) (46,0.9250) (47,0.9250) (48,0.9250)};
\addplot[cOneSevenB, line width=.85pt, densely dashed, forget plot] coordinates {(1,0.2661) (2,0.3942) (3,0.4685) (4,0.5170) (5,0.5514) (6,0.5773) (7,0.5978) (8,0.6146) (9,0.6289) (10,0.6412) (11,0.6520) (12,0.6617) (13,0.6705) (14,0.6786) (15,0.6860) (16,0.6929) (17,0.6994) (18,0.7056) (19,0.7114) (20,0.7169) (21,0.7223) (22,0.7274) (23,0.7323) (24,0.7370) (25,0.7416) (26,0.7461) (27,0.7504) (28,0.7547) (29,0.7588) (30,0.7628) (31,0.7668) (32,0.7706) (33,0.7744) (34,0.7781) (35,0.7818) (36,0.7854) (37,0.7889) (38,0.7924) (39,0.7958) (40,0.7992) (41,0.8026) (42,0.8059) (43,0.8091) (44,0.8124) (45,0.8156) (46,0.8187) (47,0.8219) (48,0.8250)};
\addplot[cOneSevenB, line width=1.25pt, forget plot] coordinates {(1,0.4099) (2,0.5200) (3,0.5849) (4,0.6298) (5,0.6638) (6,0.6909) (7,0.7134) (8,0.7324) (9,0.7487) (10,0.7628) (11,0.7751) (12,0.7860) (13,0.7956) (14,0.8042) (15,0.8118) (16,0.8187) (17,0.8249) (18,0.8305) (19,0.8357) (20,0.8404) (21,0.8447) (22,0.8487) (23,0.8524) (24,0.8558) (25,0.8590) (26,0.8620) (27,0.8648) (28,0.8675) (29,0.8700) (30,0.8723) (31,0.8745) (32,0.8766) (33,0.8787) (34,0.8806) (35,0.8824) (36,0.8841) (37,0.8858) (38,0.8874) (39,0.8889) (40,0.8903) (41,0.8917) (42,0.8931) (43,0.8943) (44,0.8956) (45,0.8967) (46,0.8979) (47,0.8990) (48,0.9000)};
\addplot[cZeroSixB, line width=.85pt, densely dashed, forget plot] coordinates {(1,0.0531) (2,0.0969) (3,0.1343) (4,0.1671) (5,0.1965) (6,0.2232) (7,0.2475) (8,0.2699) (9,0.2907) (10,0.3099) (11,0.3278) (12,0.3446) (13,0.3603) (14,0.3750) (15,0.3889) (16,0.4020) (17,0.4144) (18,0.4261) (19,0.4372) (20,0.4478) (21,0.4579) (22,0.4675) (23,0.4767) (24,0.4855) (25,0.4939) (26,0.5019) (27,0.5097) (28,0.5171) (29,0.5243) (30,0.5312) (31,0.5378) (32,0.5443) (33,0.5505) (34,0.5565) (35,0.5623) (36,0.5679) (37,0.5734) (38,0.5787) (39,0.5839) (40,0.5889) (41,0.5938) (42,0.5986) (43,0.6032) (44,0.6078) (45,0.6122) (46,0.6166) (47,0.6208) (48,0.6250)};
\addplot[cZeroSixB, line width=1.25pt, forget plot] coordinates {(1,0.2776) (2,0.3867) (3,0.4478) (4,0.4874) (5,0.5154) (6,0.5365) (7,0.5532) (8,0.5668) (9,0.5783) (10,0.5882) (11,0.5968) (12,0.6046) (13,0.6115) (14,0.6178) (15,0.6236) (16,0.6289) (17,0.6338) (18,0.6384) (19,0.6426) (20,0.6465) (21,0.6502) (22,0.6537) (23,0.6569) (24,0.6599) (25,0.6627) (26,0.6654) (27,0.6679) (28,0.6703) (29,0.6725) (30,0.6747) (31,0.6767) (32,0.6786) (33,0.6804) (34,0.6821) (35,0.6837) (36,0.6853) (37,0.6868) (38,0.6882) (39,0.6896) (40,0.6909) (41,0.6921) (42,0.6934) (43,0.6945) (44,0.6957) (45,0.6968) (46,0.6979) (47,0.6990) (48,0.7000)};
\nextgroupplot[]
\addplot[cTeacher!75, line width=.7pt, densely dotted, forget plot] coordinates {(1,0.9599) (2,0.9848) (3,0.9919) (4,0.9952) (5,0.9970) (6,0.9981) (7,0.9989) (8,0.9993) (9,0.9996) (10,0.9998) (11,0.9999) (12,0.9999) (13,1.0000) (14,1.0000) (15,1.0000) (16,1.0000) (17,1.0000) (18,1.0000) (19,1.0000) (20,1.0000) (21,1.0000) (22,1.0000) (23,1.0000) (24,1.0000) (25,1.0000) (26,1.0000) (27,1.0000) (28,1.0000) (29,1.0000) (30,1.0000) (31,1.0000) (32,1.0000) (33,1.0000) (34,1.0000) (35,1.0000) (36,1.0000) (37,1.0000) (38,1.0000) (39,1.0000) (40,1.0000) (41,1.0000) (42,1.0000) (43,1.0000) (44,1.0000) (45,1.0000) (46,1.0000) (47,1.0000) (48,1.0000)};
\addplot[cFourB, line width=.85pt, densely dashed, forget plot] coordinates {(1,0.5484) (2,0.6553) (3,0.7147) (4,0.7561) (5,0.7873) (6,0.8114) (7,0.8305) (8,0.8457) (9,0.8579) (10,0.8677) (11,0.8757) (12,0.8822) (13,0.8875) (14,0.8919) (15,0.8955) (16,0.8985) (17,0.9011) (18,0.9032) (19,0.9050) (20,0.9066) (21,0.9079) (22,0.9091) (23,0.9102) (24,0.9111) (25,0.9120) (26,0.9128) (27,0.9135) (28,0.9142) (29,0.9148) (30,0.9154) (31,0.9160) (32,0.9166) (33,0.9171) (34,0.9177) (35,0.9182) (36,0.9187) (37,0.9193) (38,0.9198) (39,0.9203) (40,0.9208) (41,0.9214) (42,0.9219) (43,0.9224) (44,0.9229) (45,0.9234) (46,0.9240) (47,0.9245) (48,0.9250)};
\addplot[cFourB, line width=1.25pt, forget plot] coordinates {(1,0.7276) (2,0.7982) (3,0.8252) (4,0.8401) (5,0.8499) (6,0.8571) (7,0.8627) (8,0.8672) (9,0.8709) (10,0.8740) (11,0.8766) (12,0.8789) (13,0.8809) (14,0.8826) (15,0.8842) (16,0.8856) (17,0.8869) (18,0.8881) (19,0.8891) (20,0.8901) (21,0.8910) (22,0.8918) (23,0.8925) (24,0.8932) (25,0.8939) (26,0.8945) (27,0.8950) (28,0.8956) (29,0.8960) (30,0.8965) (31,0.8969) (32,0.8973) (33,0.8976) (34,0.8980) (35,0.8983) (36,0.8985) (37,0.8988) (38,0.8990) (39,0.8992) (40,0.8994) (41,0.8995) (42,0.8997) (43,0.8998) (44,0.8999) (45,0.8999) (46,0.9000) (47,0.9000) (48,0.9000)};
\addplot[cOneSevenB, line width=.85pt, densely dashed, forget plot] coordinates {(1,0.2995) (2,0.3924) (3,0.4463) (4,0.4841) (5,0.5132) (6,0.5369) (7,0.5569) (8,0.5742) (9,0.5894) (10,0.6030) (11,0.6152) (12,0.6264) (13,0.6366) (14,0.6461) (15,0.6549) (16,0.6631) (17,0.6707) (18,0.6779) (19,0.6847) (20,0.6910) (21,0.6970) (22,0.7026) (23,0.7080) (24,0.7130) (25,0.7178) (26,0.7222) (27,0.7265) (28,0.7305) (29,0.7343) (30,0.7379) (31,0.7412) (32,0.7444) (33,0.7474) (34,0.7502) (35,0.7529) (36,0.7554) (37,0.7577) (38,0.7599) (39,0.7619) (40,0.7639) (41,0.7656) (42,0.7673) (43,0.7688) (44,0.7703) (45,0.7716) (46,0.7728) (47,0.7740) (48,0.7750)};
\addplot[cOneSevenB, line width=1.25pt, forget plot] coordinates {(1,0.3625) (2,0.4262) (3,0.4582) (4,0.4796) (5,0.4958) (6,0.5090) (7,0.5205) (8,0.5306) (9,0.5399) (10,0.5485) (11,0.5565) (12,0.5641) (13,0.5713) (14,0.5781) (15,0.5846) (16,0.5909) (17,0.5969) (18,0.6027) (19,0.6084) (20,0.6138) (21,0.6190) (22,0.6241) (23,0.6291) (24,0.6339) (25,0.6387) (26,0.6432) (27,0.6477) (28,0.6521) (29,0.6564) (30,0.6606) (31,0.6647) (32,0.6688) (33,0.6728) (34,0.6767) (35,0.6805) (36,0.6842) (37,0.6879) (38,0.6916) (39,0.6952) (40,0.6987) (41,0.7021) (42,0.7056) (43,0.7089) (44,0.7122) (45,0.7155) (46,0.7187) (47,0.7219) (48,0.7250)};
\addplot[cZeroSixB, line width=.85pt, densely dashed, forget plot] coordinates {(1,0.1729) (2,0.2669) (3,0.3272) (4,0.3710) (5,0.4055) (6,0.4341) (7,0.4586) (8,0.4797) (9,0.4983) (10,0.5147) (11,0.5292) (12,0.5422) (13,0.5537) (14,0.5641) (15,0.5734) (16,0.5817) (17,0.5893) (18,0.5961) (19,0.6023) (20,0.6079) (21,0.6131) (22,0.6178) (23,0.6221) (24,0.6261) (25,0.6298) (26,0.6332) (27,0.6364) (28,0.6394) (29,0.6422) (30,0.6448) (31,0.6473) (32,0.6497) (33,0.6519) (34,0.6540) (35,0.6560) (36,0.6580) (37,0.6598) (38,0.6615) (39,0.6632) (40,0.6648) (41,0.6663) (42,0.6677) (43,0.6691) (44,0.6704) (45,0.6717) (46,0.6729) (47,0.6740) (48,0.6750)};
\addplot[cZeroSixB, line width=1.25pt, forget plot] coordinates {(1,0.2120) (2,0.2746) (3,0.3070) (4,0.3290) (5,0.3463) (6,0.3609) (7,0.3739) (8,0.3857) (9,0.3966) (10,0.4069) (11,0.4165) (12,0.4256) (13,0.4343) (14,0.4425) (15,0.4504) (16,0.4579) (17,0.4651) (18,0.4720) (19,0.4787) (20,0.4850) (21,0.4912) (22,0.4971) (23,0.5028) (24,0.5083) (25,0.5137) (26,0.5189) (27,0.5239) (28,0.5287) (29,0.5334) (30,0.5379) (31,0.5424) (32,0.5466) (33,0.5508) (34,0.5548) (35,0.5587) (36,0.5625) (37,0.5662) (38,0.5698) (39,0.5732) (40,0.5766) (41,0.5799) (42,0.5830) (43,0.5861) (44,0.5890) (45,0.5919) (46,0.5947) (47,0.5974) (48,0.6000)};
\nextgroupplot[]
\addplot[cTeacher!75, line width=.7pt, densely dotted, forget plot] coordinates {(1,0.9599) (2,0.9848) (3,0.9919) (4,0.9952) (5,0.9970) (6,0.9981) (7,0.9989) (8,0.9993) (9,0.9996) (10,0.9998) (11,0.9999) (12,0.9999) (13,1.0000) (14,1.0000) (15,1.0000) (16,1.0000) (17,1.0000) (18,1.0000) (19,1.0000) (20,1.0000) (21,1.0000) (22,1.0000) (23,1.0000) (24,1.0000) (25,1.0000) (26,1.0000) (27,1.0000) (28,1.0000) (29,1.0000) (30,1.0000) (31,1.0000) (32,1.0000) (33,1.0000) (34,1.0000) (35,1.0000) (36,1.0000) (37,1.0000) (38,1.0000) (39,1.0000) (40,1.0000) (41,1.0000) (42,1.0000) (43,1.0000) (44,1.0000) (45,1.0000) (46,1.0000) (47,1.0000) (48,1.0000)};
\addplot[cFourB, line width=.85pt, densely dashed, forget plot] coordinates {(1,0.4120) (2,0.5429) (3,0.6078) (4,0.6498) (5,0.6803) (6,0.7039) (7,0.7228) (8,0.7382) (9,0.7510) (10,0.7618) (11,0.7710) (12,0.7790) (13,0.7860) (14,0.7922) (15,0.7977) (16,0.8028) (17,0.8075) (18,0.8118) (19,0.8159) (20,0.8197) (21,0.8234) (22,0.8269) (23,0.8303) (24,0.8336) (25,0.8368) (26,0.8399) (27,0.8430) (28,0.8460) (29,0.8489) (30,0.8518) (31,0.8546) (32,0.8574) (33,0.8602) (34,0.8630) (35,0.8657) (36,0.8684) (37,0.8711) (38,0.8738) (39,0.8764) (40,0.8791) (41,0.8817) (42,0.8843) (43,0.8870) (44,0.8896) (45,0.8922) (46,0.8948) (47,0.8974) (48,0.9000)};
\addplot[cFourB, line width=1.25pt, forget plot] coordinates {(1,0.7281) (2,0.8069) (3,0.8386) (4,0.8559) (5,0.8670) (6,0.8751) (7,0.8815) (8,0.8868) (9,0.8913) (10,0.8954) (11,0.8990) (12,0.9023) (13,0.9053) (14,0.9081) (15,0.9108) (16,0.9132) (17,0.9155) (18,0.9176) (19,0.9196) (20,0.9215) (21,0.9233) (22,0.9250) (23,0.9266) (24,0.9282) (25,0.9296) (26,0.9310) (27,0.9324) (28,0.9336) (29,0.9349) (30,0.9360) (31,0.9371) (32,0.9382) (33,0.9392) (34,0.9402) (35,0.9411) (36,0.9420) (37,0.9428) (38,0.9436) (39,0.9444) (40,0.9451) (41,0.9458) (42,0.9465) (43,0.9472) (44,0.9478) (45,0.9484) (46,0.9489) (47,0.9495) (48,0.9500)};
\addplot[cOneSevenB, line width=.85pt, densely dashed, forget plot] coordinates {(1,0.1286) (2,0.2100) (3,0.2651) (4,0.3047) (5,0.3347) (6,0.3583) (7,0.3774) (8,0.3934) (9,0.4070) (10,0.4187) (11,0.4290) (12,0.4380) (13,0.4461) (14,0.4534) (15,0.4600) (16,0.4660) (17,0.4715) (18,0.4765) (19,0.4811) (20,0.4854) (21,0.4894) (22,0.4931) (23,0.4966) (24,0.4999) (25,0.5030) (26,0.5060) (27,0.5088) (28,0.5114) (29,0.5140) (30,0.5164) (31,0.5188) (32,0.5210) (33,0.5232) (34,0.5253) (35,0.5274) (36,0.5293) (37,0.5313) (38,0.5331) (39,0.5350) (40,0.5368) (41,0.5385) (42,0.5403) (43,0.5420) (44,0.5436) (45,0.5452) (46,0.5469) (47,0.5484) (48,0.5500)};
\addplot[cOneSevenB, line width=1.25pt, forget plot] coordinates {(1,0.3719) (2,0.4429) (3,0.4785) (4,0.5016) (5,0.5185) (6,0.5320) (7,0.5433) (8,0.5530) (9,0.5616) (10,0.5692) (11,0.5762) (12,0.5824) (13,0.5882) (14,0.5934) (15,0.5983) (16,0.6027) (17,0.6068) (18,0.6106) (19,0.6142) (20,0.6175) (21,0.6207) (22,0.6236) (23,0.6264) (24,0.6291) (25,0.6316) (26,0.6341) (27,0.6364) (28,0.6387) (29,0.6408) (30,0.6430) (31,0.6450) (32,0.6471) (33,0.6490) (34,0.6510) (35,0.6529) (36,0.6547) (37,0.6565) (38,0.6583) (39,0.6601) (40,0.6619) (41,0.6636) (42,0.6653) (43,0.6670) (44,0.6686) (45,0.6702) (46,0.6719) (47,0.6734) (48,0.6750)};
\addplot[cZeroSixB, line width=.85pt, densely dashed, forget plot] coordinates {(1,0.0286) (2,0.0542) (3,0.0771) (4,0.0976) (5,0.1162) (6,0.1331) (7,0.1485) (8,0.1627) (9,0.1757) (10,0.1877) (11,0.1989) (12,0.2093) (13,0.2190) (14,0.2282) (15,0.2368) (16,0.2450) (17,0.2527) (18,0.2601) (19,0.2671) (20,0.2738) (21,0.2802) (22,0.2863) (23,0.2922) (24,0.2979) (25,0.3033) (26,0.3086) (27,0.3137) (28,0.3187) (29,0.3235) (30,0.3282) (31,0.3328) (32,0.3373) (33,0.3416) (34,0.3459) (35,0.3501) (36,0.3542) (37,0.3583) (38,0.3623) (39,0.3662) (40,0.3701) (41,0.3739) (42,0.3778) (43,0.3815) (44,0.3853) (45,0.3890) (46,0.3927) (47,0.3964) (48,0.4000)};
\addplot[cZeroSixB, line width=1.25pt, forget plot] coordinates {(1,0.1646) (2,0.2257) (3,0.2590) (4,0.2811) (5,0.2976) (6,0.3110) (7,0.3224) (8,0.3324) (9,0.3413) (10,0.3494) (11,0.3567) (12,0.3635) (13,0.3697) (14,0.3756) (15,0.3810) (16,0.3861) (17,0.3908) (18,0.3954) (19,0.3996) (20,0.4037) (21,0.4076) (22,0.4113) (23,0.4149) (24,0.4183) (25,0.4216) (26,0.4248) (27,0.4279) (28,0.4309) (29,0.4338) (30,0.4366) (31,0.4393) (32,0.4420) (33,0.4446) (34,0.4471) (35,0.4495) (36,0.4519) (37,0.4542) (38,0.4564) (39,0.4585) (40,0.4606) (41,0.4627) (42,0.4646) (43,0.4665) (44,0.4684) (45,0.4701) (46,0.4718) (47,0.4734) (48,0.4750)};
\nextgroupplot[ylabel={GSM8K-200\\lenient pass@$k$}]
\addplot[cTeacher!75, line width=.7pt, densely dotted, forget plot] coordinates {(1,0.9291) (2,0.9447) (3,0.9501) (4,0.9530) (5,0.9550) (6,0.9565) (7,0.9577) (8,0.9587) (9,0.9596) (10,0.9603) (11,0.9609) (12,0.9615) (13,0.9620) (14,0.9624) (15,0.9628) (16,0.9631) (17,0.9634) (18,0.9637) (19,0.9639) (20,0.9641) (21,0.9643) (22,0.9644) (23,0.9645) (24,0.9647) (25,0.9648) (26,0.9648) (27,0.9649) (28,0.9649) (29,0.9650) (30,0.9650) (31,0.9650) (32,0.9650)};
\addplot[cFourB, line width=.85pt, densely dashed, forget plot] coordinates {(1,0.7609) (2,0.8951) (3,0.9324) (4,0.9465) (5,0.9533) (6,0.9572) (7,0.9598) (8,0.9617) (9,0.9632) (10,0.9645) (11,0.9656) (12,0.9666) (13,0.9675) (14,0.9682) (15,0.9690) (16,0.9696) (17,0.9702) (18,0.9707) (19,0.9712) (20,0.9717) (21,0.9721) (22,0.9725) (23,0.9728) (24,0.9732) (25,0.9735) (26,0.9738) (27,0.9740) (28,0.9743) (29,0.9745) (30,0.9747) (31,0.9748) (32,0.9750)};
\addplot[cFourB, line width=1.25pt, forget plot] coordinates {(1,0.9187) (2,0.9413) (3,0.9494) (4,0.9545) (5,0.9583) (6,0.9613) (7,0.9638) (8,0.9659) (9,0.9676) (10,0.9691) (11,0.9704) (12,0.9715) (13,0.9725) (14,0.9734) (15,0.9741) (16,0.9748) (17,0.9754) (18,0.9760) (19,0.9765) (20,0.9769) (21,0.9773) (22,0.9777) (23,0.9780) (24,0.9783) (25,0.9786) (26,0.9789) (27,0.9791) (28,0.9793) (29,0.9795) (30,0.9797) (31,0.9798) (32,0.9800)};
\addplot[cOneSevenB, line width=.85pt, densely dashed, forget plot] coordinates {(1,0.6009) (2,0.7771) (3,0.8473) (4,0.8829) (5,0.9041) (6,0.9181) (7,0.9281) (8,0.9357) (9,0.9417) (10,0.9466) (11,0.9508) (12,0.9543) (13,0.9574) (14,0.9601) (15,0.9625) (16,0.9646) (17,0.9665) (18,0.9683) (19,0.9698) (20,0.9712) (21,0.9725) (22,0.9736) (23,0.9746) (24,0.9756) (25,0.9764) (26,0.9771) (27,0.9778) (28,0.9784) (29,0.9789) (30,0.9793) (31,0.9797) (32,0.9800)};
\addplot[cOneSevenB, line width=1.25pt, forget plot] coordinates {(1,0.8297) (2,0.8883) (3,0.9092) (4,0.9204) (5,0.9279) (6,0.9334) (7,0.9378) (8,0.9414) (9,0.9444) (10,0.9470) (11,0.9493) (12,0.9513) (13,0.9532) (14,0.9548) (15,0.9564) (16,0.9578) (17,0.9591) (18,0.9604) (19,0.9616) (20,0.9628) (21,0.9639) (22,0.9650) (23,0.9661) (24,0.9672) (25,0.9682) (26,0.9692) (27,0.9702) (28,0.9712) (29,0.9722) (30,0.9731) (31,0.9741) (32,0.9750)};
\addplot[cZeroSixB, line width=.85pt, densely dashed, forget plot] coordinates {(1,0.0873) (2,0.1566) (3,0.2147) (4,0.2649) (5,0.3089) (6,0.3479) (7,0.3828) (8,0.4142) (9,0.4426) (10,0.4685) (11,0.4922) (12,0.5139) (13,0.5339) (14,0.5525) (15,0.5697) (16,0.5858) (17,0.6008) (18,0.6148) (19,0.6280) (20,0.6404) (21,0.6520) (22,0.6630) (23,0.6733) (24,0.6831) (25,0.6924) (26,0.7011) (27,0.7094) (28,0.7173) (29,0.7248) (30,0.7319) (31,0.7386) (32,0.7450)};
\addplot[cZeroSixB, line width=1.25pt, forget plot] coordinates {(1,0.6887) (2,0.7762) (3,0.8149) (4,0.8379) (5,0.8539) (6,0.8660) (7,0.8755) (8,0.8834) (9,0.8900) (10,0.8957) (11,0.9007) (12,0.9052) (13,0.9091) (14,0.9127) (15,0.9160) (16,0.9189) (17,0.9216) (18,0.9241) (19,0.9264) (20,0.9285) (21,0.9305) (22,0.9323) (23,0.9340) (24,0.9355) (25,0.9370) (26,0.9384) (27,0.9397) (28,0.9409) (29,0.9420) (30,0.9431) (31,0.9441) (32,0.9450)};
\nextgroupplot[]
\addplot[cTeacher!75, line width=.7pt, densely dotted, forget plot] coordinates {(1,0.9372) (2,0.9465) (3,0.9500) (4,0.9520) (5,0.9534) (6,0.9544) (7,0.9552) (8,0.9559) (9,0.9565) (10,0.9571) (11,0.9576) (12,0.9581) (13,0.9585) (14,0.9589) (15,0.9594) (16,0.9597) (17,0.9601) (18,0.9605) (19,0.9608) (20,0.9612) (21,0.9615) (22,0.9618) (23,0.9622) (24,0.9625) (25,0.9628) (26,0.9631) (27,0.9634) (28,0.9637) (29,0.9641) (30,0.9644) (31,0.9647) (32,0.9650)};
\addplot[cFourB, line width=.85pt, densely dashed, forget plot] coordinates {(1,0.8992) (2,0.9361) (3,0.9483) (4,0.9549) (5,0.9593) (6,0.9625) (7,0.9649) (8,0.9669) (9,0.9685) (10,0.9698) (11,0.9710) (12,0.9720) (13,0.9729) (14,0.9737) (15,0.9744) (16,0.9750) (17,0.9756) (18,0.9761) (19,0.9766) (20,0.9770) (21,0.9774) (22,0.9777) (23,0.9781) (24,0.9784) (25,0.9786) (26,0.9789) (27,0.9791) (28,0.9793) (29,0.9795) (30,0.9797) (31,0.9798) (32,0.9800)};
\addplot[cFourB, line width=1.25pt, forget plot] coordinates {(1,0.8964) (2,0.9303) (3,0.9409) (4,0.9467) (5,0.9503) (6,0.9528) (7,0.9546) (8,0.9560) (9,0.9571) (10,0.9580) (11,0.9588) (12,0.9594) (13,0.9600) (14,0.9605) (15,0.9609) (16,0.9614) (17,0.9617) (18,0.9621) (19,0.9624) (20,0.9627) (21,0.9629) (22,0.9632) (23,0.9634) (24,0.9636) (25,0.9638) (26,0.9640) (27,0.9642) (28,0.9644) (29,0.9645) (30,0.9647) (31,0.9648) (32,0.9650)};
\addplot[cOneSevenB, line width=.85pt, densely dashed, forget plot] coordinates {(1,0.7467) (2,0.8545) (3,0.8938) (4,0.9144) (5,0.9271) (6,0.9358) (7,0.9422) (8,0.9473) (9,0.9514) (10,0.9548) (11,0.9578) (12,0.9604) (13,0.9627) (14,0.9648) (15,0.9667) (16,0.9684) (17,0.9699) (18,0.9714) (19,0.9727) (20,0.9739) (21,0.9751) (22,0.9762) (23,0.9772) (24,0.9782) (25,0.9792) (26,0.9801) (27,0.9810) (28,0.9818) (29,0.9826) (30,0.9834) (31,0.9842) (32,0.9850)};
\addplot[cOneSevenB, line width=1.25pt, forget plot] coordinates {(1,0.7572) (2,0.8493) (3,0.8819) (4,0.8992) (5,0.9103) (6,0.9181) (7,0.9239) (8,0.9285) (9,0.9321) (10,0.9352) (11,0.9378) (12,0.9400) (13,0.9419) (14,0.9436) (15,0.9452) (16,0.9466) (17,0.9478) (18,0.9490) (19,0.9501) (20,0.9511) (21,0.9520) (22,0.9529) (23,0.9537) (24,0.9545) (25,0.9553) (26,0.9560) (27,0.9567) (28,0.9574) (29,0.9581) (30,0.9587) (31,0.9594) (32,0.9600)};
\addplot[cZeroSixB, line width=.85pt, densely dashed, forget plot] coordinates {(1,0.5356) (2,0.6816) (3,0.7455) (4,0.7823) (5,0.8069) (6,0.8247) (7,0.8383) (8,0.8491) (9,0.8578) (10,0.8650) (11,0.8712) (12,0.8764) (13,0.8811) (14,0.8852) (15,0.8890) (16,0.8924) (17,0.8956) (18,0.8986) (19,0.9014) (20,0.9041) (21,0.9066) (22,0.9091) (23,0.9114) (24,0.9137) (25,0.9159) (26,0.9181) (27,0.9202) (28,0.9222) (29,0.9243) (30,0.9262) (31,0.9281) (32,0.9300)};
\addplot[cZeroSixB, line width=1.25pt, forget plot] coordinates {(1,0.6669) (2,0.7454) (3,0.7777) (4,0.7982) (5,0.8131) (6,0.8247) (7,0.8340) (8,0.8416) (9,0.8480) (10,0.8535) (11,0.8582) (12,0.8623) (13,0.8659) (14,0.8691) (15,0.8720) (16,0.8746) (17,0.8769) (18,0.8790) (19,0.8809) (20,0.8827) (21,0.8843) (22,0.8857) (23,0.8870) (24,0.8883) (25,0.8894) (26,0.8904) (27,0.8913) (28,0.8922) (29,0.8930) (30,0.8937) (31,0.8944) (32,0.8950)};
\nextgroupplot[]
\addplot[cTeacher!75, line width=.7pt, densely dotted, forget plot] coordinates {(1,0.9372) (2,0.9465) (3,0.9500) (4,0.9520) (5,0.9534) (6,0.9544) (7,0.9552) (8,0.9559) (9,0.9565) (10,0.9571) (11,0.9576) (12,0.9581) (13,0.9585) (14,0.9589) (15,0.9594) (16,0.9597) (17,0.9601) (18,0.9605) (19,0.9608) (20,0.9612) (21,0.9615) (22,0.9618) (23,0.9622) (24,0.9625) (25,0.9628) (26,0.9631) (27,0.9634) (28,0.9637) (29,0.9641) (30,0.9644) (31,0.9647) (32,0.9650)};
\addplot[cFourB, line width=.85pt, densely dashed, forget plot] coordinates {(1,0.4611) (2,0.6497) (3,0.7429) (4,0.7954) (5,0.8280) (6,0.8497) (7,0.8651) (8,0.8766) (9,0.8855) (10,0.8926) (11,0.8985) (12,0.9034) (13,0.9077) (14,0.9115) (15,0.9148) (16,0.9178) (17,0.9205) (18,0.9230) (19,0.9252) (20,0.9274) (21,0.9293) (22,0.9312) (23,0.9329) (24,0.9346) (25,0.9361) (26,0.9376) (27,0.9390) (28,0.9403) (29,0.9416) (30,0.9428) (31,0.9439) (32,0.9450)};
\addplot[cFourB, line width=1.25pt, forget plot] coordinates {(1,0.9030) (2,0.9277) (3,0.9370) (4,0.9423) (5,0.9458) (6,0.9485) (7,0.9506) (8,0.9524) (9,0.9540) (10,0.9553) (11,0.9565) (12,0.9576) (13,0.9585) (14,0.9594) (15,0.9602) (16,0.9609) (17,0.9615) (18,0.9621) (19,0.9626) (20,0.9630) (21,0.9634) (22,0.9637) (23,0.9640) (24,0.9643) (25,0.9645) (26,0.9646) (27,0.9648) (28,0.9649) (29,0.9649) (30,0.9650) (31,0.9650) (32,0.9650)};
\addplot[cOneSevenB, line width=.85pt, densely dashed, forget plot] coordinates {(1,0.2427) (2,0.3975) (3,0.5021) (4,0.5762) (5,0.6307) (6,0.6722) (7,0.7047) (8,0.7308) (9,0.7522) (10,0.7700) (11,0.7850) (12,0.7979) (13,0.8091) (14,0.8189) (15,0.8276) (16,0.8353) (17,0.8422) (18,0.8485) (19,0.8542) (20,0.8594) (21,0.8642) (22,0.8687) (23,0.8728) (24,0.8767) (25,0.8802) (26,0.8836) (27,0.8868) (28,0.8897) (29,0.8925) (30,0.8952) (31,0.8977) (32,0.9000)};
\addplot[cOneSevenB, line width=1.25pt, forget plot] coordinates {(1,0.7430) (2,0.8419) (3,0.8760) (4,0.8936) (5,0.9046) (6,0.9122) (7,0.9179) (8,0.9223) (9,0.9259) (10,0.9289) (11,0.9314) (12,0.9335) (13,0.9354) (14,0.9370) (15,0.9384) (16,0.9397) (17,0.9409) (18,0.9419) (19,0.9429) (20,0.9437) (21,0.9445) (22,0.9452) (23,0.9459) (24,0.9465) (25,0.9471) (26,0.9476) (27,0.9481) (28,0.9486) (29,0.9490) (30,0.9493) (31,0.9497) (32,0.9500)};
\addplot[cZeroSixB, line width=.85pt, densely dashed, forget plot] coordinates {(1,0.1353) (2,0.2392) (3,0.3201) (4,0.3840) (5,0.4353) (6,0.4770) (7,0.5115) (8,0.5403) (9,0.5648) (10,0.5859) (11,0.6042) (12,0.6202) (13,0.6345) (14,0.6472) (15,0.6587) (16,0.6691) (17,0.6786) (18,0.6873) (19,0.6954) (20,0.7029) (21,0.7098) (22,0.7163) (23,0.7225) (24,0.7282) (25,0.7336) (26,0.7388) (27,0.7437) (28,0.7483) (29,0.7528) (30,0.7570) (31,0.7611) (32,0.7650)};
\addplot[cZeroSixB, line width=1.25pt, forget plot] coordinates {(1,0.6197) (2,0.7049) (3,0.7404) (4,0.7621) (5,0.7776) (6,0.7896) (7,0.7994) (8,0.8076) (9,0.8147) (10,0.8208) (11,0.8263) (12,0.8312) (13,0.8357) (14,0.8397) (15,0.8435) (16,0.8469) (17,0.8501) (18,0.8531) (19,0.8559) (20,0.8585) (21,0.8610) (22,0.8633) (23,0.8654) (24,0.8675) (25,0.8694) (26,0.8712) (27,0.8729) (28,0.8745) (29,0.8760) (30,0.8774) (31,0.8788) (32,0.8800)};
\end{groupplot}
\path (pk c1r3.south west) -- (pk c3r3.south east) node[midway, below=13pt, font=\footnotesize] {number of samples $k$ (log scale)};
\end{tikzpicture}
\caption{\textbf{Lenient pass@$k$ per benchmark, before (dashed) and after (solid) distillation.} Each curve runs
to the number of samples drawn, 48 on AIME 2026 and AMC23 and 32 on GSM8K-200; its right end is the $p@n$ of
Table~\ref{tab:app-results}. On AMC23 and GSM8K-200 the two curves close in as $k$ grows in every panel; on AIME 2026, the
hardest benchmark, the distilled curve pulls clearly ahead for the 4B in thinking mode and falls behind at large $k$ for
the 0.6B from the external-SFT start.
Three-benchmark values and $k^\star$: Table~\ref{tab:app-passk}.}
\label{fig:passk}
\end{figure}
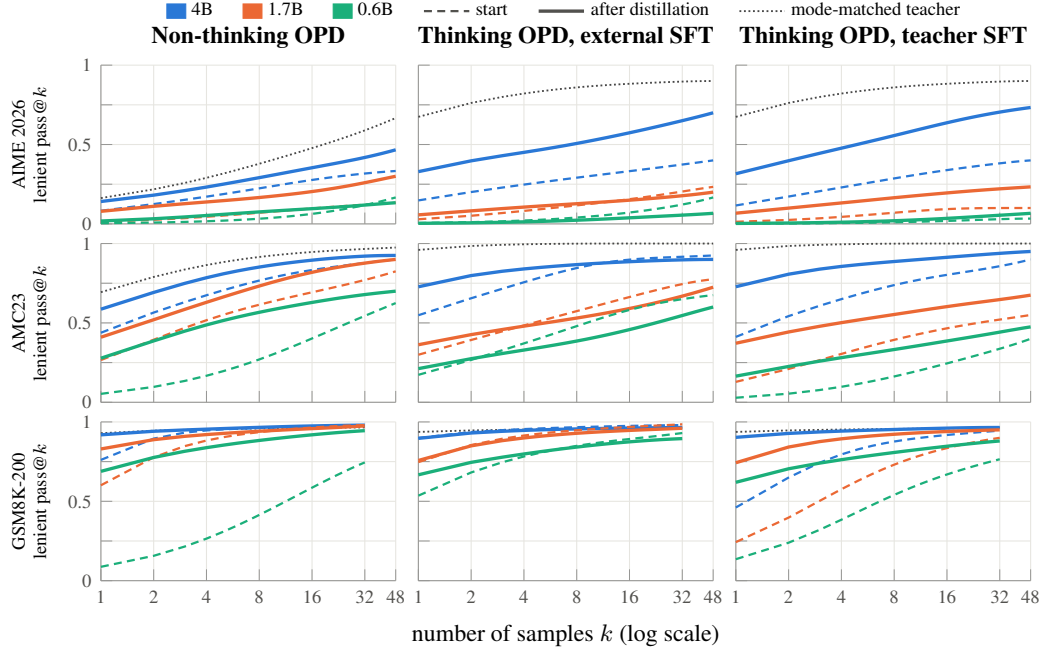

%% file: tables/lean_passk.tex
\begin{table}[h]
\centering
\footnotesize
\caption{\textbf{Lenient pass@$k$ (Figure~\ref{fig:passk}).} Three-benchmark mean of the unbiased per-problem
estimate; beyond $k=32$ GSM8K-200 (32 samples) enters with its $p@32$, so $k=48$ is the mean $p@n$. $k^\star$: the smallest $k$ at which the starting
model's pass@$k$ reaches the distilled model's $p@1$. Starting points in gray.}
\label{tab:app-passk}
\setlength{\tabcolsep}{4.2pt}
\begin{tabular}{@{}lcccccccc@{}}
\toprule
Model & $k=1$ & $k=2$ & $k=4$ & $k=8$ & $k=16$ & $k=32$ & $k=48$ & $k^\star$ \\
\midrule
Teacher (thinking) & 0.857 & 0.898 & 0.923 & 0.938 & 0.948 & 0.954 & 0.955 &  \\
Teacher (non-thinking) & 0.595 & 0.651 & 0.703 & 0.751 & 0.796 & 0.840 & 0.869 &  \\
\midrule
\textcolor{cMuted}{4B pretrained-only} & \textcolor{cMuted}{0.427} & \textcolor{cMuted}{0.528} & \textcolor{cMuted}{0.597} & \textcolor{cMuted}{0.651} & \textcolor{cMuted}{0.693} & \textcolor{cMuted}{0.723} & \textcolor{cMuted}{0.736} &  \\
\quad + non-thinking route & 0.549 & 0.605 & 0.657 & 0.703 & 0.741 & 0.773 & 0.791 & 3 \\
\textcolor{cMuted}{4B external-SFT start} & \textcolor{cMuted}{0.532} & \textcolor{cMuted}{0.597} & \textcolor{cMuted}{0.653} & \textcolor{cMuted}{0.701} & \textcolor{cMuted}{0.735} & \textcolor{cMuted}{0.757} & \textcolor{cMuted}{0.768} &  \\
\quad + thinking-mode distillation & 0.651 & 0.709 & 0.746 & 0.777 & 0.807 & 0.838 & 0.855 & 4 \\
\textcolor{cMuted}{4B teacher-SFT start} & \textcolor{cMuted}{0.330} & \textcolor{cMuted}{0.455} & \textcolor{cMuted}{0.558} & \textcolor{cMuted}{0.634} & \textcolor{cMuted}{0.686} & \textcolor{cMuted}{0.728} & \textcolor{cMuted}{0.748} &  \\
\quad + thinking-mode distillation & 0.649 & 0.711 & 0.758 & 0.799 & 0.837 & 0.869 & 0.883 & 10 \\
\midrule
\textcolor{cMuted}{1.7B pretrained-only} & \textcolor{cMuted}{0.294} & \textcolor{cMuted}{0.399} & \textcolor{cMuted}{0.482} & \textcolor{cMuted}{0.540} & \textcolor{cMuted}{0.585} & \textcolor{cMuted}{0.624} & \textcolor{cMuted}{0.646} &  \\
\quad + non-thinking route & 0.440 & 0.506 & 0.563 & 0.613 & 0.660 & 0.704 & 0.725 & 3 \\
\textcolor{cMuted}{1.7B external-SFT start} & \textcolor{cMuted}{0.358} & \textcolor{cMuted}{0.433} & \textcolor{cMuted}{0.493} & \textcolor{cMuted}{0.546} & \textcolor{cMuted}{0.596} & \textcolor{cMuted}{0.644} & \textcolor{cMuted}{0.664} &  \\
\quad + thinking-mode distillation & 0.392 & 0.453 & 0.495 & 0.529 & 0.562 & 0.602 & 0.628 & 2 \\
\textcolor{cMuted}{1.7B teacher-SFT start} & \textcolor{cMuted}{0.128} & \textcolor{cMuted}{0.211} & \textcolor{cMuted}{0.308} & \textcolor{cMuted}{0.398} & \textcolor{cMuted}{0.465} & \textcolor{cMuted}{0.507} & \textcolor{cMuted}{0.517} &  \\
\quad + thinking-mode distillation & 0.394 & 0.461 & 0.509 & 0.547 & 0.579 & 0.606 & 0.619 & 8 \\
\midrule
\textcolor{cMuted}{0.6B pretrained-only} & \textcolor{cMuted}{0.048} & \textcolor{cMuted}{0.087} & \textcolor{cMuted}{0.149} & \textcolor{cMuted}{0.239} & \textcolor{cMuted}{0.350} & \textcolor{cMuted}{0.469} & \textcolor{cMuted}{0.512} &  \\
\quad + non-thinking route & 0.328 & 0.398 & 0.459 & 0.509 & 0.548 & 0.581 & 0.593 & 15 \\
\textcolor{cMuted}{0.6B external-SFT start} & \textcolor{cMuted}{0.238} & \textcolor{cMuted}{0.320} & \textcolor{cMuted}{0.392} & \textcolor{cMuted}{0.456} & \textcolor{cMuted}{0.515} & \textcolor{cMuted}{0.567} & \textcolor{cMuted}{0.591} &  \\
\quad + thinking-mode distillation & 0.294 & 0.342 & 0.380 & 0.417 & 0.457 & 0.499 & 0.521 & 2 \\
\textcolor{cMuted}{0.6B teacher-SFT start} & \textcolor{cMuted}{0.055} & \textcolor{cMuted}{0.099} & \textcolor{cMuted}{0.162} & \textcolor{cMuted}{0.238} & \textcolor{cMuted}{0.311} & \textcolor{cMuted}{0.377} & \textcolor{cMuted}{0.399} &  \\
\quad + thinking-mode distillation & 0.262 & 0.312 & 0.351 & 0.387 & 0.423 & 0.459 & 0.474 & 11 \\
\bottomrule
\end{tabular}
\end{table}

%% file: tables/lean_bins.tex
\begin{table}[h]
\centering
\footnotesize
\caption{\textbf{Where the $p@1$ gain comes from (Figure~\ref{fig:bins}).} Problems of the three benchmarks
grouped by the starting model's lenient solve rate over its $n$ samples. Per group: number of problems, mean solve rate
at the start $\to$ after distillation, and the group's contribution to the three-benchmark $p@1$ gain (the three
contributions add up to the gain). Enter / leave: problems never solved at the start and solved at least once after,
and the reverse.}
\label{tab:app-bins}
\scriptsize
\setlength{\tabcolsep}{1.9pt}
\begin{tabular}{@{}llcccccccccc@{}}
\toprule
 & & \multicolumn{3}{c}{Never solved at the start} & \multicolumn{3}{c}{Solved in fewer than half} & \multicolumn{3}{c}{Solved in at least half} & \\
\cmidrule(lr){3-5}\cmidrule(lr){6-8}\cmidrule(lr){9-11}
Route & Size & problems & after & contrib. & problems & start $\to$ after & contrib. & problems & start $\to$ after & contrib. & \makecell{enter /\\leave} \\
\midrule
non-thinking & 4B & 29 & 0.025 & $+0.004$ & 42 & 0.219\,$\to$\,0.437 & $+0.048$ & 199 & 0.819\,$\to$\,0.967 & $+0.070$ & 8 / 2 \\
 & 1.7B & 37 & 0.017 & $+0.005$ & 86 & 0.250\,$\to$\,0.523 & $+0.080$ & 147 & 0.747\,$\to$\,0.946 & $+0.061$ & 11 / 4 \\
 & 0.6B & 91 & 0.183 & $+0.034$ & 175 & 0.098\,$\to$\,0.736 & $+0.243$ & 4 & 0.633\,$\to$\,1.000 & $+0.002$ & 49 / 7 \\
external SFT & 4B & 25 & 0.046 & $+0.013$ & 36 & 0.204\,$\to$\,0.496 & $+0.086$ & 209 & 0.952\,$\to$\,0.953 & $+0.021$ & 9 / 4 \\
 & 1.7B & 35 & 0.002 & $+0.000$ & 61 & 0.200\,$\to$\,0.268 & $+0.024$ & 174 & 0.862\,$\to$\,0.869 & $+0.010$ & 3 / 11 \\
 & 0.6B & 52 & 0.019 & $+0.003$ & 94 & 0.192\,$\to$\,0.301 & $+0.021$ & 124 & 0.776\,$\to$\,0.909 & $+0.032$ & 9 / 22 \\
teacher SFT & 4B & 33 & 0.149 & $+0.020$ & 121 & 0.262\,$\to$\,0.822 & $+0.204$ & 116 & 0.694\,$\to$\,0.990 & $+0.095$ & 16 / 0 \\
 & 1.7B & 65 & 0.069 & $+0.016$ & 184 & 0.225\,$\to$\,0.764 & $+0.232$ & 21 & 0.602\,$\to$\,0.970 & $+0.018$ & 22 / 3 \\
 & 0.6B & 100 & 0.072 & $+0.015$ & 169 & 0.164\,$\to$\,0.724 & $+0.191$ & 1 & 0.531\,$\to$\,1.000 & $+0.001$ & 34 / 7 \\
\bottomrule
\end{tabular}
\end{table}

%% file: tables/lean_ceiling.tex
\begin{table}[h]
\centering
\footnotesize
\caption{\textbf{The ceiling per benchmark.} (a) Distilled lenient p@1 minus the starting model's lenient p@n; no
difference is above zero. (b) Change in lenient p@n from start to end. Paired 95\% bootstrap intervals.}
\label{tab:app-ceiling}
\scriptsize
\setlength{\tabcolsep}{3pt}
\begin{tabular}{@{}llcccc@{}}
\toprule
Route & Size & AIME 2026 & AMC23 & GSM8K-200 & Mean \\
\midrule
\multicolumn{6}{@{}l}{\textit{(a) distilled p@1 minus starting p@n}} \\
non-thinking route & 4B & $-0.192$\,{\scriptsize[-0.328, -0.075]} & $-0.314$\,{\scriptsize[-0.421, -0.210]} & $-0.056$\,{\scriptsize[-0.084, -0.032]} & $-0.187$\,{\scriptsize[-0.246, -0.133]} \\
 & 1.7B & $-0.054$\,{\scriptsize[-0.125, -0.002]} & $-0.415$\,{\scriptsize[-0.538, -0.296]} & $-0.150$\,{\scriptsize[-0.188, -0.114]} & $-0.206$\,{\scriptsize[-0.255, -0.160]} \\
 & 0.6B & $-0.149$\,{\scriptsize[-0.283, -0.032]} & $-0.347$\,{\scriptsize[-0.473, -0.222]} & $-0.056$\,{\scriptsize[-0.106, -0.008]} & $-0.184$\,{\scriptsize[-0.247, -0.126]} \\
external-SFT start & 4B & $-0.071$\,{\scriptsize[-0.176, +0.023]} & $-0.197$\,{\scriptsize[-0.294, -0.112]} & $-0.084$\,{\scriptsize[-0.114, -0.056]} & $-0.117$\,{\scriptsize[-0.165, -0.074]} \\
 & 1.7B & $-0.176$\,{\scriptsize[-0.311, -0.063]} & $-0.412$\,{\scriptsize[-0.550, -0.282]} & $-0.228$\,{\scriptsize[-0.270, -0.187]} & $-0.272$\,{\scriptsize[-0.335, -0.212]} \\
 & 0.6B & $-0.163$\,{\scriptsize[-0.296, -0.033]} & $-0.463$\,{\scriptsize[-0.602, -0.328]} & $-0.263$\,{\scriptsize[-0.315, -0.214]} & $-0.296$\,{\scriptsize[-0.365, -0.234]} \\
teacher-SFT start & 4B & $-0.084$\,{\scriptsize[-0.185, +0.003]} & $-0.172$\,{\scriptsize[-0.265, -0.090]} & $-0.042$\,{\scriptsize[-0.074, -0.011]} & $-0.099$\,{\scriptsize[-0.146, -0.058]} \\
 & 1.7B & $-0.033$\,{\scriptsize[-0.102, +0.019]} & $-0.178$\,{\scriptsize[-0.289, -0.082]} & $-0.157$\,{\scriptsize[-0.194, -0.121]} & $-0.123$\,{\scriptsize[-0.166, -0.083]} \\
 & 0.6B & $-0.031$\,{\scriptsize[-0.094, +0.002]} & $-0.235$\,{\scriptsize[-0.359, -0.124]} & $-0.145$\,{\scriptsize[-0.191, -0.100]} & $-0.137$\,{\scriptsize[-0.186, -0.092]} \\
\addlinespace[3pt]
\multicolumn{6}{@{}l}{\textit{(b) change in p@n, start to end}} \\
non-thinking route & 4B & $+0.133$\,{\scriptsize[+0.033, +0.267]} & $+0.025$\,{\scriptsize[+0.000, +0.075]} & $+0.005$\,{\scriptsize[-0.015, +0.030]} & $+0.054$\,{\scriptsize[+0.014, +0.102]} \\
 & 1.7B & $+0.167$\,{\scriptsize[+0.033, +0.300]} & $+0.075$\,{\scriptsize[-0.025, +0.175]} & $-0.005$\,{\scriptsize[-0.025, +0.015]} & $+0.079$\,{\scriptsize[+0.024, +0.139]} \\
 & 0.6B & $-0.033$\,{\scriptsize[-0.167, +0.100]} & $+0.075$\,{\scriptsize[-0.075, +0.225]} & $+0.200$\,{\scriptsize[+0.145, +0.260]} & $+0.081$\,{\scriptsize[+0.008, +0.151]} \\
external-SFT start & 4B & $+0.300$\,{\scriptsize[+0.133, +0.467]} & $-0.025$\,{\scriptsize[-0.075, +0.000]} & $-0.015$\,{\scriptsize[-0.035, +0.000]} & $+0.087$\,{\scriptsize[+0.031, +0.147]} \\
 & 1.7B & $-0.033$\,{\scriptsize[-0.100, +0.000]} & $-0.050$\,{\scriptsize[-0.175, +0.075]} & $-0.025$\,{\scriptsize[-0.055, +0.000]} & $-0.036$\,{\scriptsize[-0.084, +0.008]} \\
 & 0.6B & $-0.100$\,{\scriptsize[-0.200, +0.000]} & $-0.075$\,{\scriptsize[-0.250, +0.075]} & $-0.035$\,{\scriptsize[-0.075, +0.005]} & $-0.070$\,{\scriptsize[-0.137, -0.007]} \\
teacher-SFT start & 4B & $+0.333$\,{\scriptsize[+0.167, +0.500]} & $+0.050$\,{\scriptsize[+0.000, +0.125]} & $+0.020$\,{\scriptsize[+0.005, +0.040]} & $+0.134$\,{\scriptsize[+0.075, +0.198]} \\
 & 1.7B & $+0.133$\,{\scriptsize[+0.033, +0.267]} & $+0.125$\,{\scriptsize[+0.000, +0.250]} & $+0.050$\,{\scriptsize[+0.015, +0.090]} & $+0.103$\,{\scriptsize[+0.047, +0.165]} \\
 & 0.6B & $+0.033$\,{\scriptsize[+0.000, +0.100]} & $+0.075$\,{\scriptsize[-0.050, +0.200]} & $+0.115$\,{\scriptsize[+0.065, +0.170]} & $+0.074$\,{\scriptsize[+0.025, +0.127]} \\
\bottomrule
\end{tabular}
\end{table}

%% file: tables/lean_decomp.tex
\begin{table}[h]
\centering
\footnotesize
\caption{\textbf{What distillation changes, start to end, on the three-benchmark mean.} Differences with paired 95\%
bootstrap intervals. \emph{Answer accuracy}: accuracy of the samples that mark an answer, on the problems both models
answered.}
\label{tab:app-decomp}
\scriptsize
\setlength{\tabcolsep}{3pt}
\begin{tabular}{@{}llcccc@{}}
\toprule
Route & Size & \makecell{marked-answer\\rate} & \makecell{answer\\accuracy} & lenient p@1 & lenient p@n \\
\midrule
non-thinking route & 4B & $+0.115$\,{\scriptsize[+0.088, +0.143]} & $+0.063$\,{\scriptsize[+0.041, +0.088]} & $+0.121$\,{\scriptsize[+0.097, +0.149]} & $+0.054$\,{\scriptsize[+0.014, +0.102]} \\
 & 1.7B & $+0.134$\,{\scriptsize[+0.106, +0.161]} & $+0.080$\,{\scriptsize[+0.054, +0.111]} & $+0.146$\,{\scriptsize[+0.118, +0.177]} & $+0.079$\,{\scriptsize[+0.024, +0.139]} \\
 & 0.6B & $+0.633$\,{\scriptsize[+0.598, +0.667]} & $+0.092$\,{\scriptsize[+0.065, +0.118]} & $+0.280$\,{\scriptsize[+0.249, +0.310]} & $+0.081$\,{\scriptsize[+0.008, +0.151]} \\
\addlinespace[2pt]
external-SFT start & 4B & $-0.006$\,{\scriptsize[-0.043, +0.033]} & $+0.190$\,{\scriptsize[+0.142, +0.243]} & $+0.119$\,{\scriptsize[+0.083, +0.158]} & $+0.087$\,{\scriptsize[+0.031, +0.147]} \\
 & 1.7B & $-0.145$\,{\scriptsize[-0.180, -0.110]} & $+0.106$\,{\scriptsize[+0.072, +0.145]} & $+0.034$\,{\scriptsize[+0.013, +0.057]} & $-0.036$\,{\scriptsize[-0.084, +0.008]} \\
 & 0.6B & $-0.251$\,{\scriptsize[-0.290, -0.210]} & $+0.115$\,{\scriptsize[+0.089, +0.142]} & $+0.056$\,{\scriptsize[+0.035, +0.077]} & $-0.070$\,{\scriptsize[-0.137, -0.007]} \\
\addlinespace[2pt]
teacher-SFT start & 4B & $+0.368$\,{\scriptsize[+0.334, +0.403]} & $+0.104$\,{\scriptsize[+0.063, +0.147]} & $+0.319$\,{\scriptsize[+0.284, +0.357]} & $+0.134$\,{\scriptsize[+0.075, +0.198]} \\
 & 1.7B & $+0.314$\,{\scriptsize[+0.281, +0.349]} & $+0.061$\,{\scriptsize[+0.018, +0.110]} & $+0.266$\,{\scriptsize[+0.232, +0.301]} & $+0.103$\,{\scriptsize[+0.047, +0.165]} \\
 & 0.6B & $+0.244$\,{\scriptsize[+0.214, +0.276]} & $+0.087$\,{\scriptsize[+0.051, +0.124]} & $+0.207$\,{\scriptsize[+0.180, +0.237]} & $+0.074$\,{\scriptsize[+0.025, +0.127]} \\
\bottomrule
\end{tabular}
\end{table}

%% file: tables/lean_gpqa.tex
\begin{table}[h]
\centering
\footnotesize
\caption{\textbf{GPQA-Diamond, non-thinking route only, scored by the last marked answer.} With four options,
chance is 0.250. Differences with paired 95\% bootstrap intervals.}
\label{tab:app-gpqa}
\scriptsize
\setlength{\tabcolsep}{3pt}
\begin{tabular}{@{}lccccc@{}}
\toprule
Model & \makecell{committed p@1\\start $\to$ end} & change & \makecell{end minus\\chance} & \makecell{strict p@1\\end} & \makecell{truncation\\start $\to$ end} \\
\midrule
Teacher (non-thinking) & 0.499 & -- & $+0.249$\,{\scriptsize[+0.200, +0.300]} & 0.497 & 0.026 \\
4B & 0.157\,$\to$\,0.397 & $+0.239$\,{\scriptsize[+0.198, +0.282]} & $+0.146$\,{\scriptsize[+0.099, +0.195]} & 0.361 & 0.462\,$\to$\,0.125 \\
1.7B & 0.191\,$\to$\,0.275 & $+0.084$\,{\scriptsize[+0.050, +0.117]} & $+0.025$\,{\scriptsize[-0.011, +0.062]} & 0.270 & 0.181\,$\to$\,0.111 \\
0.6B & 0.049\,$\to$\,0.234 & $+0.185$\,{\scriptsize[+0.155, +0.215]} & $-0.016$\,{\scriptsize[-0.047, +0.017]} & 0.233 & 0.581\,$\to$\,0.154 \\
\bottomrule
\end{tabular}
\end{table}

%% file: appendix/D_stopping.tex
\section{Supplementary results: why small students do not learn to stop}
\label{app:block2}

\paragraph{Training dynamics in full.}
Figure~\ref{fig:collapse} adds to the closure rate of Figure~\ref{fig:closure} the median response length relative to
the budget, the accuracy of the closed responses and the fraction of responses with a marked answer, over the same
steps. The responses fill the budget in the step in which closure collapses, and the marked answers disappear with it.

\input{figures/fig2_collapse}

\paragraph{The collapse in every thinking-mode run.}
Table~\ref{tab:app-controls} lists the closure of every thinking-mode run over the first 30 steps. Each run from the
external-SFT start drops at step 6 at the 7,168-token budget and at step 7 at 16,384 tokens; the officially
post-trained 0.6B drops at step 6 and partly recovers afterwards. The runs from the teacher-SFT start and the
1,024-token run close little from the first step on, because their responses already reach the budget.

\input{tables/lean_controls}

\paragraph{Stop reliability and the closure each run keeps.}
Table~\ref{tab:app-reliability} gives, for every thinking-mode run, how often its closed responses were correct over
the first four steps and how much of its early closure it keeps late in training. Among the runs from the
external-SFT start, both follow size. The officially post-trained 0.6B, whose stops were right in about a third of
cases, keeps more than half of its closure over its 20 steps.

\input{tables/lean_reliability}

The 4B also stops finishing its training responses, but at evaluation, where the limit is 30,720 tokens, its longer
reasoning still ends in time, which is why its marked-answer rate there does not fall.

\subsection{Probes}
\label{app:probes}

\paragraph{The teacher's stop signal (Table~\ref{tab:probe}, Figure~\ref{fig:probe}).}
For each student from the external-SFT start we took its responses at the first training step, about a hundred per student at
the 7,168-token budget, and had the thinking teacher score every token. For responses that closed, we read the teacher's
probability of \think{} at the position where the student closed. For responses that ran to the budget, we read the
largest probability the teacher gave \think{} at any position and report its median over responses.

\input{tables/tab2_probe}

\paragraph{The stop token on identical prefixes (Table~\ref{tab:app-stoplogit}).}
We cut prefixes from the officially post-trained 0.6B's own outputs at four kinds of position (where it closed, just after a correct or a wrong marked answer, and at token 500) and asked the model
before and after 20 steps of thinking-mode distillation for its probability of \think{} as the next token. Where the
starting model had closed, both models rank \think{} first. Everywhere else the token lies tens of thousands of ranks
deep in both, so the changes there, although their intervals exclude zero, move a probability that is negligible in
both models.

\input{tables/lean_stoplogit}

\paragraph{A lengthening-only null model (Table~\ref{tab:app-nullmodel}).}
If distillation only made every response longer, closes would disappear wherever the longer response no longer fits
the limit, whether the answer was right or wrong. Stretching the starting model's responses by one factor, fitted so
that closure matches the observed value, removes most of the wrong closes that were lost, but also a quarter of the
correct ones, which the distilled model in fact keeps.

\input{tables/lean_nullmodel}

%% file: figures/fig2_collapse.tex
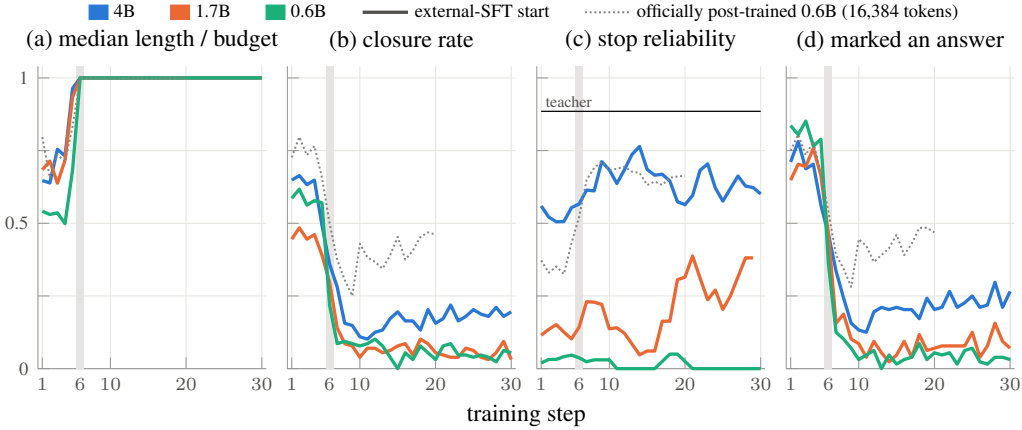
\begin{figure}[t]
\centering
\begin{tikzpicture}
\begin{groupplot}[group style={group size=4 by 1, horizontal sep=0.28cm,
    yticklabels at=edge left, group name=collapse},
  paperstyle, scale only axis, width=3.02cm, height=4.00cm,
  xmin=1, xmax=30, ymin=0, ymax=1.04, enlarge x limits=0.02, clip=false,
  xtick={1,6,10,20,30}, ytick={0,0.25,0.5,0.75,1}, yticklabels={0,,0.5,,1},
  axis x line*=bottom, axis y line*=left, title style={yshift=-4pt}]
\nextgroupplot[title={(a) median length / budget}]
\fill[cNeutral!40] (axis cs:5.5,0) rectangle (axis cs:6.5,1.04);
\addplot[cFourB, line width=1.25pt, line join=round, forget plot] coordinates {(1,0.6470) (2,0.6387) (3,0.7546) (4,0.7312) (5,0.9637) (6,1.0000) (7,1.0000) (8,1.0000) (9,1.0000) (10,1.0000) (11,1.0000) (12,1.0000) (13,1.0000) (14,1.0000) (15,1.0000) (16,1.0000) (17,1.0000) (18,1.0000) (19,1.0000) (20,1.0000) (21,1.0000) (22,1.0000) (23,1.0000) (24,1.0000) (25,1.0000) (26,1.0000) (27,1.0000) (28,1.0000) (29,1.0000) (30,1.0000)};
\addplot[cOneSevenB, line width=1.25pt, line join=round, forget plot] coordinates {(1,0.6842) (2,0.7140) (3,0.6378) (4,0.7169) (5,0.9337) (6,0.9999) (7,1.0000) (8,1.0000) (9,1.0000) (10,1.0000) (11,1.0000) (12,1.0000) (13,1.0000) (14,1.0000) (15,1.0000) (16,1.0000) (17,1.0000) (18,1.0000) (19,1.0000) (20,1.0000) (21,1.0000) (22,1.0000) (23,1.0000) (24,1.0000) (25,1.0000) (26,1.0000) (27,1.0000) (28,1.0000) (29,1.0000) (30,1.0000)};
\addplot[cZeroSixB, line width=1.25pt, line join=round, forget plot] coordinates {(1,0.5416) (2,0.5303) (3,0.5357) (4,0.4987) (5,0.6829) (6,0.9999) (7,1.0000) (8,1.0000) (9,1.0000) (10,1.0000) (11,1.0000) (12,1.0000) (13,1.0000) (14,1.0000) (15,1.0000) (16,1.0000) (17,1.0000) (18,1.0000) (19,1.0000) (20,1.0000) (21,1.0000) (22,1.0000) (23,1.0000) (24,1.0000) (25,1.0000) (26,1.0000) (27,1.0000) (28,1.0000) (29,1.0000) (30,1.0000)};
\addplot[line width=.75pt, line join=round, cTeacher!50, densely dotted, forget plot] coordinates {(1,0.7965) (2,0.6558) (3,0.7369) (4,0.7113) (5,0.8332) (6,0.9995) (7,0.9999) (8,0.9999) (9,0.9999) (10,0.9999) (11,0.9999) (12,0.9999) (13,1.0000) (14,0.9999) (15,0.9999) (16,0.9999) (17,0.9999) (18,0.9998) (19,0.9998) (20,0.9998)};
\nextgroupplot[title={(b) closure rate}, legend style={at={(1.046,1)}, anchor=south, yshift=13pt, legend columns=5, /tikz/every even column/.append style={column sep=1.1em}}]
\addlegendimage{legend image code/.code={\fill[cFourB] (0cm,-0.6ex) rectangle (0.25cm,0.6ex);}}
\addlegendentry{4B}
\addlegendimage{legend image code/.code={\fill[cOneSevenB] (0cm,-0.6ex) rectangle (0.25cm,0.6ex);}}
\addlegendentry{1.7B}
\addlegendimage{legend image code/.code={\fill[cZeroSixB] (0cm,-0.6ex) rectangle (0.25cm,0.6ex);}}
\addlegendentry{0.6B}
\addlegendimage{cMuted, line width=1.25pt, line join=round}
\addlegendentry{external-SFT start}
\addlegendimage{line width=.75pt, line join=round, cTeacher!50, densely dotted}
\addlegendentry{officially post-trained 0.6B (16{,}384 tokens)}
\fill[cNeutral!40] (axis cs:5.5,0) rectangle (axis cs:6.5,1.04);
\addplot[cFourB, line width=1.25pt, line join=round, forget plot] coordinates {(1,0.6484) (2,0.6641) (3,0.6328) (4,0.6484) (5,0.4922) (6,0.3594) (7,0.2812) (8,0.1562) (9,0.1484) (10,0.1094) (11,0.1016) (12,0.1250) (13,0.1328) (14,0.1719) (15,0.1953) (16,0.1641) (17,0.1641) (18,0.1328) (19,0.2031) (20,0.1562) (21,0.1719) (22,0.2188) (23,0.1641) (24,0.1797) (25,0.2031) (26,0.1875) (27,0.1797) (28,0.2109) (29,0.1797) (30,0.1953)};
\addplot[cOneSevenB, line width=1.25pt, line join=round, forget plot] coordinates {(1,0.4453) (2,0.4844) (3,0.4453) (4,0.4609) (5,0.3906) (6,0.2969) (7,0.1406) (8,0.0859) (9,0.0781) (10,0.0391) (11,0.0703) (12,0.0703) (13,0.0547) (14,0.0625) (15,0.0781) (16,0.0859) (17,0.0469) (18,0.1016) (19,0.0859) (20,0.0547) (21,0.0469) (22,0.0391) (23,0.0391) (24,0.0703) (25,0.0625) (26,0.0391) (27,0.0312) (28,0.0547) (29,0.0938) (30,0.0312)};
\addplot[cZeroSixB, line width=1.25pt, line join=round, forget plot] coordinates {(1,0.5859) (2,0.6172) (3,0.5625) (4,0.5781) (5,0.5703) (6,0.2188) (7,0.0859) (8,0.0938) (9,0.0859) (10,0.0781) (11,0.0859) (12,0.1016) (13,0.0781) (14,0.0391) (15,0.0000) (16,0.0547) (17,0.0312) (18,0.0781) (19,0.0547) (20,0.0312) (21,0.0781) (22,0.0859) (23,0.0469) (24,0.0469) (25,0.0391) (26,0.0469) (27,0.0391) (28,0.0234) (29,0.0625) (30,0.0547)};
\addplot[line width=.75pt, line join=round, cTeacher!50, densely dotted, forget plot] coordinates {(1,0.7266) (2,0.7969) (3,0.7344) (4,0.7656) (5,0.6562) (6,0.5000) (7,0.3750) (8,0.3047) (9,0.2500) (10,0.4297) (11,0.3828) (12,0.3672) (13,0.3438) (14,0.3906) (15,0.4531) (16,0.3750) (17,0.4062) (18,0.4531) (19,0.4688) (20,0.4609)};
\nextgroupplot[title={(c) stop reliability}]
\fill[cNeutral!40] (axis cs:5.5,0) rectangle (axis cs:6.5,1.04);
\draw[cTeacher, line width=.5pt] (axis cs:1,0.885) -- (axis cs:30,0.885) node[pos=0, above right, inner sep=1.5pt, font=\tiny, text=cMuted] {teacher};
\addplot[cFourB, line width=1.25pt, line join=round, forget plot] coordinates {(1,0.5596) (2,0.5212) (3,0.5051) (4,0.5060) (5,0.5536) (6,0.5661) (7,0.6142) (8,0.6115) (9,0.7120) (10,0.6838) (11,0.6367) (12,0.6843) (13,0.7367) (14,0.7642) (15,0.6849) (16,0.6654) (17,0.6685) (18,0.6453) (19,0.5739) (20,0.5641) (21,0.5952) (22,0.6825) (23,0.7043) (24,0.6219) (25,0.5762) (26,0.6197) (27,0.6625) (28,0.6280) (29,0.6228) (30,0.6009)};
\addplot[cOneSevenB, line width=1.25pt, line join=round, forget plot] coordinates {(1,0.1150) (2,0.1351) (3,0.1515) (4,0.1257) (5,0.1023) (6,0.1425) (7,0.2304) (8,0.2286) (9,0.2212) (10,0.1370) (11,0.1407) (12,0.1217) (13,0.0847) (14,0.0476) (15,0.0606) (16,0.0606) (17,0.1632) (18,0.1632) (19,0.3060) (20,0.3146) (21,0.3873) (22,0.3111) (23,0.2370) (24,0.2704) (25,0.2037) (26,0.2500) (28,0.3810) (29,0.3810)};
\addplot[cZeroSixB, line width=1.25pt, line join=round, forget plot] coordinates {(1,0.0197) (2,0.0317) (3,0.0318) (4,0.0412) (5,0.0465) (6,0.0375) (7,0.0238) (8,0.0303) (9,0.0303) (10,0.0303) (11,0.0000) (12,0.0000) (13,0.0000) (14,0.0000) (16,0.0000) (18,0.0500) (19,0.0500) (21,0.0000) (22,0.0000) (23,0.0000) (24,0.0000) (25,0.0000) (26,0.0000) (27,0.0000) (29,0.0000) (30,0.0000)};
\addplot[line width=.75pt, line join=round, cTeacher!50, densely dotted, forget plot] coordinates {(1,0.3725) (2,0.3299) (3,0.3515) (4,0.3253) (5,0.4313) (6,0.5208) (7,0.6490) (8,0.6907) (9,0.7109) (10,0.6825) (11,0.6874) (12,0.6949) (13,0.6774) (14,0.6732) (15,0.6316) (16,0.6434) (17,0.6319) (18,0.6569) (19,0.6608) (20,0.6635)};
\nextgroupplot[title={(d) marked an answer}]
\fill[cNeutral!40] (axis cs:5.5,0) rectangle (axis cs:6.5,1.04);
\addplot[cFourB, line width=1.25pt, line join=round, forget plot] coordinates {(1,0.7109) (2,0.7812) (3,0.6875) (4,0.7031) (5,0.5625) (6,0.4609) (7,0.3359) (8,0.2422) (9,0.1562) (10,0.1328) (11,0.1250) (12,0.1953) (13,0.2109) (14,0.2031) (15,0.2109) (16,0.2031) (17,0.2031) (18,0.1719) (19,0.2422) (20,0.2031) (21,0.2109) (22,0.2656) (23,0.2109) (24,0.2266) (25,0.2500) (26,0.2109) (27,0.2188) (28,0.2969) (29,0.2109) (30,0.2656)};
\addplot[cOneSevenB, line width=1.25pt, line join=round, forget plot] coordinates {(1,0.6484) (2,0.7031) (3,0.6953) (4,0.7578) (5,0.6641) (6,0.4609) (7,0.1562) (8,0.1875) (9,0.1016) (10,0.0859) (11,0.0391) (12,0.0938) (13,0.0547) (14,0.0234) (15,0.0469) (16,0.0938) (17,0.0391) (18,0.1172) (19,0.0625) (20,0.0703) (21,0.0781) (22,0.0781) (23,0.0781) (24,0.0781) (25,0.1250) (26,0.0391) (27,0.0781) (28,0.1562) (29,0.0938) (30,0.0703)};
\addplot[cZeroSixB, line width=1.25pt, line join=round, forget plot] coordinates {(1,0.8359) (2,0.8047) (3,0.8516) (4,0.7656) (5,0.7891) (6,0.3594) (7,0.1250) (8,0.1016) (9,0.0703) (10,0.0312) (11,0.0469) (12,0.0625) (13,0.0000) (14,0.0469) (15,0.0156) (16,0.0312) (17,0.0391) (18,0.0859) (19,0.0312) (20,0.0547) (21,0.0469) (22,0.0547) (23,0.0156) (24,0.0625) (25,0.0703) (26,0.0234) (27,0.0156) (28,0.0391) (29,0.0391) (30,0.0312)};
\addplot[line width=.75pt, line join=round, cTeacher!50, densely dotted, forget plot] coordinates {(1,0.7500) (2,0.8047) (3,0.7344) (4,0.7812) (5,0.6719) (6,0.5391) (7,0.3906) (8,0.3438) (9,0.2812) (10,0.4453) (11,0.4219) (12,0.3672) (13,0.3906) (14,0.4141) (15,0.4609) (16,0.3906) (17,0.4297) (18,0.4844) (19,0.4844) (20,0.4688)};
\end{groupplot}
\path (collapse c2r1.south east) -- (collapse c3r1.south west) node[midway, below=10pt, font=\footnotesize] {training step};
\end{tikzpicture}
\caption{\textbf{In thinking-mode training, the responses fill the budget at step 6, closure collapses in
the same step, and the students stop marking answers.} Training responses, 7{,}168-token budget unless noted. (c) Stop
reliability, the accuracy of closed responses: 3-step centered moving average, omitted below five closed responses; black
line: the thinking teacher, 0.885.}
\label{fig:collapse}
\end{figure}

%% file: tables/lean_controls.tex
\begin{table}[h]
\centering
\footnotesize
\caption{\textbf{Every thinking-mode run loses closure early.} Closure rate of the training responses, averaged over
step windows, and the first step at which it falls below three quarters of its steps 1--4 level. Runs from the
teacher-SFT start and the 1{,}024-token run begin with little closure to lose.}
\label{tab:app-controls}
\setlength{\tabcolsep}{5pt}
\begin{tabular}{@{}lccccc@{}}
\toprule
Run & steps 1--5 & 6--10 & 11--20 & 21--30 & first step below $0.75\times$ \\
\midrule
\multicolumn{6}{@{}l}{\textit{External-SFT start, 7,168 tokens}} \\
\quad 4B, run 1 & 0.617 & 0.211 & 0.155 & 0.189 & 6 \\
\quad 4B, run 2 & 0.633 & 0.208 & 0.170 & 0.212 & 6 \\
\quad 1.7B & 0.445 & 0.128 & 0.071 & 0.051 & 6 \\
\quad 0.6B, run 1 & 0.605 & 0.145 & 0.058 & 0.042 & 6 \\
\quad 0.6B, run 2 & 0.583 & 0.113 & 0.055 & 0.052 & 6 \\
\addlinespace[2pt]
\multicolumn{6}{@{}l}{\textit{Other rollout budgets}} \\
\quad 0.6B, 16,384 tokens & 0.608 & 0.297 & -- & -- & 7 \\
\quad 0.6B, 1,024 tokens & 0.125 & 0.105 & 0.080 & 0.064 & 5 \\
\addlinespace[2pt]
\multicolumn{6}{@{}l}{\textit{Teacher-SFT start, 7,168 tokens}} \\
\quad 4B & 0.225 & 0.231 & 0.145 & 0.148 & 11 \\
\quad 1.7B & 0.078 & 0.033 & 0.039 & 0.023 & 6 \\
\quad 0.6B & 0.117 & 0.070 & 0.032 & 0.055 & 6 \\
\addlinespace[2pt]
\multicolumn{6}{@{}l}{\textit{Officially post-trained 0.6B, 16,384 tokens}} \\
\quad run 1 & 0.736 & 0.372 & 0.410 & -- & 6 \\
\quad run 2 & 0.731 & 0.366 & 0.438 & -- & 6 \\
\bottomrule
\end{tabular}
\end{table}

%% file: tables/lean_reliability.tex
\begin{table}[h]
\centering
\footnotesize
\caption{\textbf{Stop reliability before training, and the closure each run keeps.} \emph{Stop reliability}: the
fraction of closed training responses whose answer is correct over steps 1--4; the thinking teacher reaches 0.885 on its own samples. \emph{Late closure}: the closure rate over the last
stored steps of the run (steps 11--20 for the 20-step runs). \emph{Kept}: late closure divided by the closure over steps
1--4.}
\label{tab:app-reliability}
\setlength{\tabcolsep}{5pt}
\begin{tabular}{@{}lcccc@{}}
\toprule
Run & stop reliability & closure, steps 1--4 & late closure & kept \\
\midrule
4B external-SFT start, run 1 & 0.521 & 0.648 & 0.226 & 0.349 \\
4B external-SFT start, run 2 & 0.509 & 0.652 & 0.226 & 0.347 \\
1.7B external-SFT start & 0.128 & 0.459 & 0.064 & 0.139 \\
0.6B external-SFT start, run 1 & 0.042 & 0.611 & 0.023 & 0.038 \\
0.6B external-SFT start, run 2 & 0.033 & 0.586 & 0.026 & 0.044 \\
4B teacher-SFT start & 0.259 & 0.219 & 0.211 & 0.963 \\
1.7B teacher-SFT start & 0.025 & 0.078 & 0.003 & 0.038 \\
0.6B teacher-SFT start & 0.000 & 0.119 & 0.002 & 0.017 \\
officially post-trained 0.6B, run 1 & 0.349 & 0.756 & 0.410 & 0.542 \\
officially post-trained 0.6B, run 2 & 0.362 & 0.750 & 0.438 & 0.584 \\
\bottomrule
\end{tabular}
\end{table}

%% file: tables/tab2_probe.tex
\begin{table}[h]
\centering
\caption{\textbf{The teacher endorses the student's own stops and gives almost no stop signal along a
truncated trajectory.} The thinking teacher scores each student's step-1 rollouts (7{,}168-token budget);
$p$ is its probability of \think. Advantage: what the objective gives the stop token. Stop reliability:
accuracy of the rollouts that closed, steps 1--4. Details: Appendix~\ref{app:probes}.}
\label{tab:probe}
\footnotesize
\setlength{\tabcolsep}{4.5pt}
\begin{tabular}{@{}lccccccc@{}}
\toprule
 & \multicolumn{3}{c}{Samples the student stopped} & \multicolumn{3}{c}{Truncated samples} & \makecell{Before\\training} \\
\cmidrule(lr){2-4}\cmidrule(lr){5-7}\cmidrule(lr){8-8}
Student & $n$ & \makecell{$p$ at the\\stop} & \makecell{Stop-token\\advantage} & $n$ & \makecell{Ever\\$p>0.1$} & \makecell{Median of\\max $p$} & \makecell{Stop\\reliability} \\
\midrule
\sizedot{cFourB}\,4B external-SFT start & 83 & 0.976 & $+0.027$ & 36 & 0.000 & $4.4\times10^{-9}$ & 0.521 \\
\sizedot{cOneSevenB}\,1.7B external-SFT start & 57 & 0.764 & $+0.032$ & 40 & 0.175 & $5.8\times10^{-8}$ & 0.128 \\
\sizedot{cZeroSixB}\,0.6B external-SFT start & 75 & 0.673 & $-0.062$ & 23 & 0.043 & $1.0\times10^{-7}$ & 0.042 \\
\midrule
\textcolor{cMuted}{\textit{Teacher (thinking), own samples}} &  &  &  &  &  &  & \textcolor{cMuted}{0.885} \\
\bottomrule
\end{tabular}
\end{table}

%% file: tables/lean_stoplogit.tex
\begin{table}[h]
\centering
\footnotesize
\caption{\textbf{On identical prefixes, distillation leaves the stop token where it was.} The officially post-trained
0.6B before and after 20 steps of thinking-mode distillation reads the same prefixes, cut from the starting model's own
outputs, and we compare the log-probability each assigns to \think{} as the next token. Two disjoint prefix sets.
$\Delta$: after minus before, with a 95\% bootstrap interval. Rank 0: \think{} is the most likely next token.}
\label{tab:app-stoplogit}
\scriptsize
\setlength{\tabcolsep}{3pt}
\begin{tabular}{@{}lcccccc@{}}
\toprule
 & & \multicolumn{2}{c}{$\log p(\think)$} & & & median rank \\
\cmidrule(lr){3-4}
Position in the prefix & $n$ & before & after & $\Delta$ & 95\% interval & before\,$\to$\,after \\
\midrule
\multicolumn{7}{@{}l}{\textit{Prefix set 1}} \\
\quad where the starting model closed & 205 & $-0.000$ & $-0.001$ & $-0.001$ & [$-0.003$, $-0.000$] & 0\,$\to$\,0 \\
\quad just after a correct marked answer & 71 & $-25.865$ & $-27.235$ & $-1.370$ & [$-1.591$, $-1.157$] & 64,401\,$\to$\,67,323 \\
\quad just after a wrong marked answer & 135 & $-23.113$ & $-24.032$ & $-0.919$ & [$-1.089$, $-0.753$] & 48,935\,$\to$\,38,821 \\
\quad at token 500 & 256 & $-31.530$ & $-36.190$ & $-4.660$ & [$-4.953$, $-4.360$] & 79,233\,$\to$\,80,691 \\
\addlinespace[2pt]
\multicolumn{7}{@{}l}{\textit{Prefix set 2}} \\
\quad where the starting model closed & 203 & $-0.000$ & $-0.001$ & $-0.000$ & [$-0.000$, $-0.000$] & 0\,$\to$\,0 \\
\quad just after a correct marked answer & 86 & $-25.913$ & $-27.273$ & $-1.360$ & [$-1.572$, $-1.158$] & 63,738\,$\to$\,65,571 \\
\quad just after a wrong marked answer & 118 & $-24.355$ & $-25.397$ & $-1.042$ & [$-1.220$, $-0.869$] & 53,955\,$\to$\,43,624 \\
\quad at token 500 & 256 & $-31.874$ & $-36.346$ & $-4.472$ & [$-4.754$, $-4.197$] & 77,146\,$\to$\,77,038 \\
\bottomrule
\end{tabular}
\end{table}

%% file: tables/lean_nullmodel.tex
\begin{table}[h]
\centering
\footnotesize
\caption{\textbf{Lengthening alone explains most lost wrong stops but not the kept correct ones.} 128 problems, 4
samples each, from the officially post-trained 0.6B before and after 20 thinking-mode steps. The null model stretches
every starting sample's closing position by one factor $k$, chosen so that its closure matches the observed one, and
keeps the close only if it still falls within the 16{,}384-token limit. Fractions of
all samples.}
\label{tab:app-nullmodel}
\setlength{\tabcolsep}{6pt}
\begin{tabular}{@{}lccc@{}}
\toprule
 & closed & closed and correct & closed and wrong \\
\midrule
before training & 0.797 & 0.305 & 0.492 \\
null model, $k = 1.71$ & 0.439 & 0.229 & 0.211 \\
after 20 thinking-mode steps & 0.438 & 0.311 & 0.127 \\
\bottomrule
\end{tabular}
\end{table}

%% file: appendix/E_unfinished.tex
\section{Supplementary results: what the samples that do not stop contain}
\label{app:block3}

\paragraph{The answer cases per benchmark.}
Table~\ref{tab:app-cases} splits every model's samples, benchmark by benchmark, into those with no marked answer, those
whose marked answers are all wrong, and those with a correct one. From the external-SFT start the share of only-wrong
samples falls on every benchmark and at every size; at 1.7B and 0.6B the share with no marked answer rises on every
benchmark. From the teacher-SFT start distillation mostly adds right answers, because those starts rarely marked an
answer at all.

\input{tables/lean_cases}

\input{figures/fig_cases}

\paragraph{The unfinished samples per benchmark.}
Table~\ref{tab:app-unfinished-bench} repeats Table~\ref{tab:unfinished} benchmark by benchmark for the distilled students
from the external-SFT start and for the teacher. For the students the correct value is found above chance on every benchmark, most
clearly on GSM8K-200.

\input{tables/lean_unfinished}

\paragraph{Forcing a stop.}
We appended \think{} to every unfinished sample of the officially post-trained 0.6B on 128 training problems and let the
model continue (Table~\ref{tab:app-forced}). Before and after distillation, a forced stop recovers a correct answer mostly
on problems the starting model could solve, and rarely on problems it never solved. Many continuations run out of their
1,024 tokens without answering, so these rates are lower bounds.

\input{tables/lean_forced}

\paragraph{Where the first marked answer appears.}
Figure~\ref{fig:anatomy} places every evaluated model by where its first marked answer appears in the output and by how
often it overwrites a correct marked answer. The teacher answers late and almost never overwrites (0.85 and 0.004). The
1.7B and 0.6B distilled from the teacher-SFT start answer at about half of the output and overwrite in 0.437 and
0.180 of the samples in which a correct answer appeared. From the external-SFT start the same sizes answer late (0.89
and 0.86), and they overwrite more as they get smaller.

\input{figures/fig3_anatomy}

\paragraph{How repeated endings are counted.}
A sample ends in repetition if some non-empty line occurs at least five times in its last 2,000 characters. Samples that
mark many answers repeat their answer lines, which is why the students from the teacher-SFT start score high on this
measure.

%% file: tables/lean_cases.tex
\begin{table}[h]
\centering
\footnotesize
\caption{\textbf{The answer cases per benchmark.} Fraction of samples with no marked answer, with only wrong marked
answers, and with a correct marked answer (lenient p@1). Starting points in gray.}
\label{tab:app-cases}
\setlength{\tabcolsep}{4pt}
\begin{tabular}{@{}lccccccccc@{}}
\toprule
 & \multicolumn{3}{c}{AIME 2026} & \multicolumn{3}{c}{AMC23} & \multicolumn{3}{c}{GSM8K-200} \\
\cmidrule(lr){2-4}\cmidrule(lr){5-7}\cmidrule(lr){8-10}
Model & none & wrong & right & none & wrong & right & none & wrong & right \\
\midrule
Teacher (thinking) & 0.092 & 0.234 & 0.674 & 0.014 & 0.026 & 0.960 & 0.007 & 0.056 & 0.937 \\
\midrule
\textcolor{cMuted}{4B external-SFT start} & \textcolor{cMuted}{0.537} & \textcolor{cMuted}{0.315} & \textcolor{cMuted}{0.148} & \textcolor{cMuted}{0.242} & \textcolor{cMuted}{0.210} & \textcolor{cMuted}{0.549} & \textcolor{cMuted}{0.020} & \textcolor{cMuted}{0.081} & \textcolor{cMuted}{0.899} \\
\quad + distillation & 0.535 & 0.135 & 0.329 & 0.230 & 0.042 & 0.728 & 0.050 & 0.054 & 0.896 \\
\textcolor{cMuted}{4B teacher-SFT start} & \textcolor{cMuted}{0.819} & \textcolor{cMuted}{0.065} & \textcolor{cMuted}{0.116} & \textcolor{cMuted}{0.565} & \textcolor{cMuted}{0.023} & \textcolor{cMuted}{0.412} & \textcolor{cMuted}{0.516} & \textcolor{cMuted}{0.023} & \textcolor{cMuted}{0.461} \\
\quad + distillation & 0.531 & 0.153 & 0.316 & 0.228 & 0.044 & 0.728 & 0.038 & 0.059 & 0.903 \\
\midrule
\textcolor{cMuted}{1.7B external-SFT start} & \textcolor{cMuted}{0.640} & \textcolor{cMuted}{0.331} & \textcolor{cMuted}{0.029} & \textcolor{cMuted}{0.391} & \textcolor{cMuted}{0.310} & \textcolor{cMuted}{0.300} & \textcolor{cMuted}{0.089} & \textcolor{cMuted}{0.164} & \textcolor{cMuted}{0.747} \\
\quad + distillation & 0.842 & 0.101 & 0.057 & 0.540 & 0.097 & 0.363 & 0.172 & 0.071 & 0.757 \\
\textcolor{cMuted}{1.7B teacher-SFT start} & \textcolor{cMuted}{0.961} & \textcolor{cMuted}{0.026} & \textcolor{cMuted}{0.013} & \textcolor{cMuted}{0.852} & \textcolor{cMuted}{0.019} & \textcolor{cMuted}{0.129} & \textcolor{cMuted}{0.733} & \textcolor{cMuted}{0.025} & \textcolor{cMuted}{0.243} \\
\quad + distillation & 0.861 & 0.071 & 0.067 & 0.554 & 0.074 & 0.372 & 0.189 & 0.068 & 0.743 \\
\midrule
\textcolor{cMuted}{0.6B external-SFT start} & \textcolor{cMuted}{0.485} & \textcolor{cMuted}{0.509} & \textcolor{cMuted}{0.006} & \textcolor{cMuted}{0.331} & \textcolor{cMuted}{0.496} & \textcolor{cMuted}{0.173} & \textcolor{cMuted}{0.169} & \textcolor{cMuted}{0.296} & \textcolor{cMuted}{0.536} \\
\quad + distillation & 0.851 & 0.145 & 0.003 & 0.644 & 0.144 & 0.212 & 0.242 & 0.092 & 0.667 \\
\textcolor{cMuted}{0.6B teacher-SFT start} & \textcolor{cMuted}{0.968} & \textcolor{cMuted}{0.031} & \textcolor{cMuted}{0.001} & \textcolor{cMuted}{0.930} & \textcolor{cMuted}{0.041} & \textcolor{cMuted}{0.029} & \textcolor{cMuted}{0.835} & \textcolor{cMuted}{0.030} & \textcolor{cMuted}{0.135} \\
\quad + distillation & 0.933 & 0.065 & 0.003 & 0.748 & 0.087 & 0.165 & 0.319 & 0.061 & 0.620 \\
\bottomrule
\end{tabular}
\end{table}

%% file: figures/fig_cases.tex
\begin{figure}[t]
\centering
\begin{tikzpicture}
\begin{groupplot}[group style={group size=4 by 1, horizontal sep=0.28cm,
    yticklabels at=edge left, ylabels at=edge left, group name=cases},
  paperstyle, scale only axis, height=3.50cm, ymin=0, ymax=1, enlargelimits=false, clip=false,
  ytick={0,0.25,0.5,0.75,1}, yticklabels={0,,0.5,,1}, xmajorgrids=false, ymajorgrids=false,
  xtick style={draw=none}, xticklabel style={text=cMuted},
  axis x line*=bottom, axis y line*=left, ylabel={fraction of samples}, title style={yshift=-3pt}]
\nextgroupplot[width=0.895cm, xmin=0.45, xmax=1.55, xtick={1}, xticklabels={thinking}, title={\textbf{Teacher}}]
\fill[cFourB] (axis cs:0.640,0.0000) rectangle (axis cs:1.360,0.8482);
\fill[cFourB!58] (axis cs:0.640,0.8482) rectangle (axis cs:1.360,0.8545);
\draw[white, line width=.5pt] (axis cs:0.640,0.8482) -- (axis cs:1.360,0.8482);
\fill[cFourB!26] (axis cs:0.640,0.8545) rectangle (axis cs:1.360,0.8571);
\draw[white, line width=.5pt] (axis cs:0.640,0.8545) -- (axis cs:1.360,0.8545);
\fill[cOneSevenB] (axis cs:0.640,0.8571) rectangle (axis cs:1.360,0.9623);
\draw[white, line width=.5pt] (axis cs:0.640,0.8571) -- (axis cs:1.360,0.8571);
\node[font=\tiny, text=cTeacher, inner sep=.6pt] at (axis cs:1.000,0.9097) {0.105};
\fill[cNeutral!45] (axis cs:0.640,0.9623) rectangle (axis cs:1.360,0.9999);
\fill[pattern=north east lines, pattern color=cMuted!75] (axis cs:0.640,0.9623) rectangle (axis cs:1.360,0.9999);
\draw[white, line width=.5pt] (axis cs:0.640,0.9623) -- (axis cs:1.360,0.9623);
\nextgroupplot[width=3.66cm, xmin=0.45, xmax=4.95, xtick={1,2,3.4,4.4}, xticklabels={start,after,start,after}, title={\textbf{4B}}]
\fill[cFourB] (axis cs:0.640,0.0000) rectangle (axis cs:1.360,0.5297);
\fill[cFourB!58] (axis cs:0.640,0.5297) rectangle (axis cs:1.360,0.5312);
\draw[white, line width=.5pt] (axis cs:0.640,0.5297) -- (axis cs:1.360,0.5297);
\fill[cFourB!26] (axis cs:0.640,0.5312) rectangle (axis cs:1.360,0.5319);
\draw[white, line width=.5pt] (axis cs:0.640,0.5312) -- (axis cs:1.360,0.5312);
\fill[cOneSevenB] (axis cs:0.640,0.5319) rectangle (axis cs:1.360,0.7339);
\draw[white, line width=.5pt] (axis cs:0.640,0.5319) -- (axis cs:1.360,0.5319);
\node[font=\tiny, text=cTeacher, inner sep=.6pt] at (axis cs:1.000,0.6329) {0.202};
\fill[cNeutral!45] (axis cs:0.640,0.7339) rectangle (axis cs:1.360,1.0001);
\fill[pattern=north east lines, pattern color=cMuted!75] (axis cs:0.640,0.7339) rectangle (axis cs:1.360,1.0001);
\draw[white, line width=.5pt] (axis cs:0.640,0.7339) -- (axis cs:1.360,0.7339);
\node[fill=white, fill opacity=.85, text opacity=1, font=\tiny, text=cTeacher, inner sep=.6pt] at (axis cs:1.000,0.8670) {0.266};
\fill[cFourB] (axis cs:1.640,0.0000) rectangle (axis cs:2.360,0.6413);
\fill[cFourB!58] (axis cs:1.640,0.6413) rectangle (axis cs:2.360,0.6500);
\draw[white, line width=.5pt] (axis cs:1.640,0.6413) -- (axis cs:2.360,0.6413);
\fill[cFourB!26] (axis cs:1.640,0.6500) rectangle (axis cs:2.360,0.6511);
\draw[white, line width=.5pt] (axis cs:1.640,0.6500) -- (axis cs:2.360,0.6500);
\fill[cOneSevenB] (axis cs:1.640,0.6511) rectangle (axis cs:2.360,0.7283);
\draw[white, line width=.5pt] (axis cs:1.640,0.6511) -- (axis cs:2.360,0.6511);
\node[font=\tiny, text=cTeacher, inner sep=.6pt] at (axis cs:2.000,0.6897) {0.077};
\fill[cNeutral!45] (axis cs:1.640,0.7283) rectangle (axis cs:2.360,1.0000);
\fill[pattern=north east lines, pattern color=cMuted!75] (axis cs:1.640,0.7283) rectangle (axis cs:2.360,1.0000);
\draw[white, line width=.5pt] (axis cs:1.640,0.7283) -- (axis cs:2.360,0.7283);
\node[fill=white, fill opacity=.85, text opacity=1, font=\tiny, text=cTeacher, inner sep=.6pt] at (axis cs:2.000,0.8642) {0.272};
\fill[cFourB] (axis cs:3.040,0.0000) rectangle (axis cs:3.760,0.3228);
\fill[cFourB!58] (axis cs:3.040,0.3228) rectangle (axis cs:3.760,0.3276);
\draw[white, line width=.5pt] (axis cs:3.040,0.3228) -- (axis cs:3.760,0.3228);
\fill[cFourB!26] (axis cs:3.040,0.3276) rectangle (axis cs:3.760,0.3297);
\draw[white, line width=.5pt] (axis cs:3.040,0.3276) -- (axis cs:3.760,0.3276);
\fill[cOneSevenB] (axis cs:3.040,0.3297) rectangle (axis cs:3.760,0.3664);
\draw[white, line width=.5pt] (axis cs:3.040,0.3297) -- (axis cs:3.760,0.3297);
\fill[cNeutral!45] (axis cs:3.040,0.3664) rectangle (axis cs:3.760,1.0000);
\fill[pattern=north east lines, pattern color=cMuted!75] (axis cs:3.040,0.3664) rectangle (axis cs:3.760,1.0000);
\draw[white, line width=.5pt] (axis cs:3.040,0.3664) -- (axis cs:3.760,0.3664);
\node[fill=white, fill opacity=.85, text opacity=1, font=\tiny, text=cTeacher, inner sep=.6pt] at (axis cs:3.400,0.6832) {0.634};
\fill[cFourB] (axis cs:4.040,0.0000) rectangle (axis cs:4.760,0.6347);
\fill[cFourB!58] (axis cs:4.040,0.6347) rectangle (axis cs:4.760,0.6426);
\draw[white, line width=.5pt] (axis cs:4.040,0.6347) -- (axis cs:4.760,0.6347);
\fill[cFourB!26] (axis cs:4.040,0.6426) rectangle (axis cs:4.760,0.6490);
\draw[white, line width=.5pt] (axis cs:4.040,0.6426) -- (axis cs:4.760,0.6426);
\fill[cOneSevenB] (axis cs:4.040,0.6490) rectangle (axis cs:4.760,0.7344);
\draw[white, line width=.5pt] (axis cs:4.040,0.6490) -- (axis cs:4.760,0.6490);
\node[font=\tiny, text=cTeacher, inner sep=.6pt] at (axis cs:4.400,0.6917) {0.085};
\fill[cNeutral!45] (axis cs:4.040,0.7344) rectangle (axis cs:4.760,0.9999);
\fill[pattern=north east lines, pattern color=cMuted!75] (axis cs:4.040,0.7344) rectangle (axis cs:4.760,0.9999);
\draw[white, line width=.5pt] (axis cs:4.040,0.7344) -- (axis cs:4.760,0.7344);
\node[fill=white, fill opacity=.85, text opacity=1, font=\tiny, text=cTeacher, inner sep=.6pt] at (axis cs:4.400,0.8672) {0.266};
\draw[cMuted!70, line width=.4pt] ([yshift=-13pt]axis cs:0.64,0) -- ([yshift=-13pt]axis cs:2.36,0) node[midway, below, font=\scriptsize, text=cMuted, inner sep=1.5pt, align=center] {external\\SFT};
\draw[cMuted!70, line width=.4pt] ([yshift=-13pt]axis cs:3.04,0) -- ([yshift=-13pt]axis cs:4.76,0) node[midway, below, font=\scriptsize, text=cMuted, inner sep=1.5pt, align=center] {teacher\\SFT};
\nextgroupplot[width=3.66cm, xmin=0.45, xmax=4.95, xtick={1,2,3.4,4.4}, xticklabels={start,after,start,after}, title={\textbf{1.7B}}, legend style={at={(0.340,1)}, anchor=south, yshift=12pt, legend columns=5, /tikz/every even column/.append style={column sep=0.9em}}]
\addlegendimage{legend image code/.code={\fill[cFourB] (0cm,-0.7ex) rectangle (0.3cm,0.7ex);}}
\addlegendentry{one value, right}
\addlegendimage{legend image code/.code={\fill[cFourB!58] (0cm,-0.7ex) rectangle (0.3cm,0.7ex);}}
\addlegendentry{several values, last right}
\addlegendimage{legend image code/.code={\fill[cFourB!26] (0cm,-0.7ex) rectangle (0.3cm,0.7ex);}}
\addlegendentry{right, then overwritten}
\addlegendimage{legend image code/.code={\fill[cOneSevenB] (0cm,-0.7ex) rectangle (0.3cm,0.7ex);}}
\addlegendentry{only wrong}
\addlegendimage{legend image code/.code={\fill[cNeutral!45] (0cm,-0.7ex) rectangle (0.3cm,0.7ex);\fill[pattern=north east lines, pattern color=cMuted!75] (0cm,-0.7ex) rectangle (0.3cm,0.7ex);}}
\addlegendentry{no marked answer}
\fill[cFourB] (axis cs:0.640,0.0000) rectangle (axis cs:1.360,0.3459);
\fill[cFourB!58] (axis cs:0.640,0.3459) rectangle (axis cs:1.360,0.3535);
\draw[white, line width=.5pt] (axis cs:0.640,0.3459) -- (axis cs:1.360,0.3459);
\fill[cFourB!26] (axis cs:0.640,0.3535) rectangle (axis cs:1.360,0.3584);
\draw[white, line width=.5pt] (axis cs:0.640,0.3535) -- (axis cs:1.360,0.3535);
\fill[cOneSevenB] (axis cs:0.640,0.3584) rectangle (axis cs:1.360,0.6269);
\draw[white, line width=.5pt] (axis cs:0.640,0.3584) -- (axis cs:1.360,0.3584);
\node[font=\tiny, text=cTeacher, inner sep=.6pt] at (axis cs:1.000,0.4927) {0.269};
\fill[cNeutral!45] (axis cs:0.640,0.6269) rectangle (axis cs:1.360,0.9999);
\fill[pattern=north east lines, pattern color=cMuted!75] (axis cs:0.640,0.6269) rectangle (axis cs:1.360,0.9999);
\draw[white, line width=.5pt] (axis cs:0.640,0.6269) -- (axis cs:1.360,0.6269);
\node[fill=white, fill opacity=.85, text opacity=1, font=\tiny, text=cTeacher, inner sep=.6pt] at (axis cs:1.000,0.8134) {0.373};
\fill[cFourB] (axis cs:1.640,0.0000) rectangle (axis cs:2.360,0.3672);
\fill[cFourB!58] (axis cs:1.640,0.3672) rectangle (axis cs:2.360,0.3868);
\draw[white, line width=.5pt] (axis cs:1.640,0.3672) -- (axis cs:2.360,0.3672);
\fill[cFourB!26] (axis cs:1.640,0.3868) rectangle (axis cs:2.360,0.3922);
\draw[white, line width=.5pt] (axis cs:1.640,0.3868) -- (axis cs:2.360,0.3868);
\fill[cOneSevenB] (axis cs:1.640,0.3922) rectangle (axis cs:2.360,0.4820);
\draw[white, line width=.5pt] (axis cs:1.640,0.3922) -- (axis cs:2.360,0.3922);
\node[font=\tiny, text=cTeacher, inner sep=.6pt] at (axis cs:2.000,0.4371) {0.090};
\fill[cNeutral!45] (axis cs:1.640,0.4820) rectangle (axis cs:2.360,1.0000);
\fill[pattern=north east lines, pattern color=cMuted!75] (axis cs:1.640,0.4820) rectangle (axis cs:2.360,1.0000);
\draw[white, line width=.5pt] (axis cs:1.640,0.4820) -- (axis cs:2.360,0.4820);
\node[fill=white, fill opacity=.85, text opacity=1, font=\tiny, text=cTeacher, inner sep=.6pt] at (axis cs:2.000,0.7410) {0.518};
\fill[cFourB] (axis cs:3.040,0.0000) rectangle (axis cs:3.760,0.1187);
\fill[cFourB!58] (axis cs:3.040,0.1187) rectangle (axis cs:3.760,0.1252);
\draw[white, line width=.5pt] (axis cs:3.040,0.1187) -- (axis cs:3.760,0.1187);
\fill[cFourB!26] (axis cs:3.040,0.1252) rectangle (axis cs:3.760,0.1282);
\draw[white, line width=.5pt] (axis cs:3.040,0.1252) -- (axis cs:3.760,0.1252);
\fill[cOneSevenB] (axis cs:3.040,0.1282) rectangle (axis cs:3.760,0.1514);
\draw[white, line width=.5pt] (axis cs:3.040,0.1282) -- (axis cs:3.760,0.1282);
\fill[cNeutral!45] (axis cs:3.040,0.1514) rectangle (axis cs:3.760,1.0001);
\fill[pattern=north east lines, pattern color=cMuted!75] (axis cs:3.040,0.1514) rectangle (axis cs:3.760,1.0001);
\draw[white, line width=.5pt] (axis cs:3.040,0.1514) -- (axis cs:3.760,0.1514);
\node[fill=white, fill opacity=.85, text opacity=1, font=\tiny, text=cTeacher, inner sep=.6pt] at (axis cs:3.400,0.5757) {0.849};
\fill[cFourB] (axis cs:4.040,0.0000) rectangle (axis cs:4.760,0.2415);
\fill[cFourB!58] (axis cs:4.040,0.2415) rectangle (axis cs:4.760,0.2541);
\draw[white, line width=.5pt] (axis cs:4.040,0.2415) -- (axis cs:4.760,0.2415);
\fill[cFourB!26] (axis cs:4.040,0.2541) rectangle (axis cs:4.760,0.3940);
\draw[white, line width=.5pt] (axis cs:4.040,0.2541) -- (axis cs:4.760,0.2541);
\fill[cOneSevenB] (axis cs:4.040,0.3940) rectangle (axis cs:4.760,0.4654);
\draw[white, line width=.5pt] (axis cs:4.040,0.3940) -- (axis cs:4.760,0.3940);
\node[font=\tiny, text=cTeacher, inner sep=.6pt] at (axis cs:4.400,0.4297) {0.071};
\fill[cNeutral!45] (axis cs:4.040,0.4654) rectangle (axis cs:4.760,1.0000);
\fill[pattern=north east lines, pattern color=cMuted!75] (axis cs:4.040,0.4654) rectangle (axis cs:4.760,1.0000);
\draw[white, line width=.5pt] (axis cs:4.040,0.4654) -- (axis cs:4.760,0.4654);
\node[fill=white, fill opacity=.85, text opacity=1, font=\tiny, text=cTeacher, inner sep=.6pt] at (axis cs:4.400,0.7327) {0.535};
\draw[cMuted!70, line width=.4pt] ([yshift=-13pt]axis cs:0.64,0) -- ([yshift=-13pt]axis cs:2.36,0) node[midway, below, font=\scriptsize, text=cMuted, inner sep=1.5pt, align=center] {external\\SFT};
\draw[cMuted!70, line width=.4pt] ([yshift=-13pt]axis cs:3.04,0) -- ([yshift=-13pt]axis cs:4.76,0) node[midway, below, font=\scriptsize, text=cMuted, inner sep=1.5pt, align=center] {teacher\\SFT};
\nextgroupplot[width=3.66cm, xmin=0.45, xmax=4.95, xtick={1,2,3.4,4.4}, xticklabels={start,after,start,after}, title={\textbf{0.6B}}]
\fill[cFourB] (axis cs:0.640,0.0000) rectangle (axis cs:1.360,0.2138);
\fill[cFourB!58] (axis cs:0.640,0.2138) rectangle (axis cs:1.360,0.2316);
\draw[white, line width=.5pt] (axis cs:0.640,0.2138) -- (axis cs:1.360,0.2138);
\fill[cFourB!26] (axis cs:0.640,0.2316) rectangle (axis cs:1.360,0.2381);
\draw[white, line width=.5pt] (axis cs:0.640,0.2316) -- (axis cs:1.360,0.2316);
\fill[cOneSevenB] (axis cs:0.640,0.2381) rectangle (axis cs:1.360,0.6716);
\draw[white, line width=.5pt] (axis cs:0.640,0.2381) -- (axis cs:1.360,0.2381);
\node[font=\tiny, text=cTeacher, inner sep=.6pt] at (axis cs:1.000,0.4548) {0.433};
\fill[cNeutral!45] (axis cs:0.640,0.6716) rectangle (axis cs:1.360,1.0000);
\fill[pattern=north east lines, pattern color=cMuted!75] (axis cs:0.640,0.6716) rectangle (axis cs:1.360,1.0000);
\draw[white, line width=.5pt] (axis cs:0.640,0.6716) -- (axis cs:1.360,0.6716);
\node[fill=white, fill opacity=.85, text opacity=1, font=\tiny, text=cTeacher, inner sep=.6pt] at (axis cs:1.000,0.8358) {0.328};
\fill[cFourB] (axis cs:1.640,0.0000) rectangle (axis cs:2.360,0.2768);
\fill[cFourB!58] (axis cs:1.640,0.2768) rectangle (axis cs:2.360,0.2857);
\draw[white, line width=.5pt] (axis cs:1.640,0.2768) -- (axis cs:2.360,0.2768);
\fill[cFourB!26] (axis cs:1.640,0.2857) rectangle (axis cs:2.360,0.2941);
\draw[white, line width=.5pt] (axis cs:1.640,0.2857) -- (axis cs:2.360,0.2857);
\fill[cOneSevenB] (axis cs:1.640,0.2941) rectangle (axis cs:2.360,0.4211);
\draw[white, line width=.5pt] (axis cs:1.640,0.2941) -- (axis cs:2.360,0.2941);
\node[font=\tiny, text=cTeacher, inner sep=.6pt] at (axis cs:2.000,0.3576) {0.127};
\fill[cNeutral!45] (axis cs:1.640,0.4211) rectangle (axis cs:2.360,1.0000);
\fill[pattern=north east lines, pattern color=cMuted!75] (axis cs:1.640,0.4211) rectangle (axis cs:2.360,1.0000);
\draw[white, line width=.5pt] (axis cs:1.640,0.4211) -- (axis cs:2.360,0.4211);
\node[fill=white, fill opacity=.85, text opacity=1, font=\tiny, text=cTeacher, inner sep=.6pt] at (axis cs:2.000,0.7106) {0.579};
\fill[cFourB] (axis cs:3.040,0.0000) rectangle (axis cs:3.760,0.0477);
\fill[cFourB!58] (axis cs:3.040,0.0477) rectangle (axis cs:3.760,0.0505);
\draw[white, line width=.5pt] (axis cs:3.040,0.0477) -- (axis cs:3.760,0.0477);
\fill[cFourB!26] (axis cs:3.040,0.0505) rectangle (axis cs:3.760,0.0551);
\draw[white, line width=.5pt] (axis cs:3.040,0.0505) -- (axis cs:3.760,0.0505);
\fill[cOneSevenB] (axis cs:3.040,0.0551) rectangle (axis cs:3.760,0.0889);
\draw[white, line width=.5pt] (axis cs:3.040,0.0551) -- (axis cs:3.760,0.0551);
\fill[cNeutral!45] (axis cs:3.040,0.0889) rectangle (axis cs:3.760,1.0000);
\fill[pattern=north east lines, pattern color=cMuted!75] (axis cs:3.040,0.0889) rectangle (axis cs:3.760,1.0000);
\draw[white, line width=.5pt] (axis cs:3.040,0.0889) -- (axis cs:3.760,0.0889);
\node[fill=white, fill opacity=.85, text opacity=1, font=\tiny, text=cTeacher, inner sep=.6pt] at (axis cs:3.400,0.5444) {0.911};
\fill[cFourB] (axis cs:4.040,0.0000) rectangle (axis cs:4.760,0.2098);
\fill[cFourB!58] (axis cs:4.040,0.2098) rectangle (axis cs:4.760,0.2289);
\draw[white, line width=.5pt] (axis cs:4.040,0.2098) -- (axis cs:4.760,0.2098);
\fill[cFourB!26] (axis cs:4.040,0.2289) rectangle (axis cs:4.760,0.2624);
\draw[white, line width=.5pt] (axis cs:4.040,0.2289) -- (axis cs:4.760,0.2289);
\fill[cOneSevenB] (axis cs:4.040,0.2624) rectangle (axis cs:4.760,0.3334);
\draw[white, line width=.5pt] (axis cs:4.040,0.2624) -- (axis cs:4.760,0.2624);
\node[font=\tiny, text=cTeacher, inner sep=.6pt] at (axis cs:4.400,0.2979) {0.071};
\fill[cNeutral!45] (axis cs:4.040,0.3334) rectangle (axis cs:4.760,1.0000);
\fill[pattern=north east lines, pattern color=cMuted!75] (axis cs:4.040,0.3334) rectangle (axis cs:4.760,1.0000);
\draw[white, line width=.5pt] (axis cs:4.040,0.3334) -- (axis cs:4.760,0.3334);
\node[fill=white, fill opacity=.85, text opacity=1, font=\tiny, text=cTeacher, inner sep=.6pt] at (axis cs:4.400,0.6667) {0.667};
\draw[cMuted!70, line width=.4pt] ([yshift=-13pt]axis cs:0.64,0) -- ([yshift=-13pt]axis cs:2.36,0) node[midway, below, font=\scriptsize, text=cMuted, inner sep=1.5pt, align=center] {external\\SFT};
\draw[cMuted!70, line width=.4pt] ([yshift=-13pt]axis cs:3.04,0) -- ([yshift=-13pt]axis cs:4.76,0) node[midway, below, font=\scriptsize, text=cMuted, inner sep=1.5pt, align=center] {teacher\\SFT};
\end{groupplot}
\end{tikzpicture}
\caption{\textbf{Below 4B, distillation trades wrong answers for no answer at all.} From the external-SFT
start, the share of samples with only wrong answers shrinks at every size; at 4B right answers take its place, at 1.7B
and 0.6B mostly samples that mark no answer. From the teacher-SFT start,
whose students rarely marked an answer, it mostly adds right answers. Three-benchmark mean; the blue cases sum to
lenient $p@1$; after distillation every no-answer sample is truncated. Per benchmark: Table~\ref{tab:app-cases}.}
\label{fig:cases}
\end{figure}
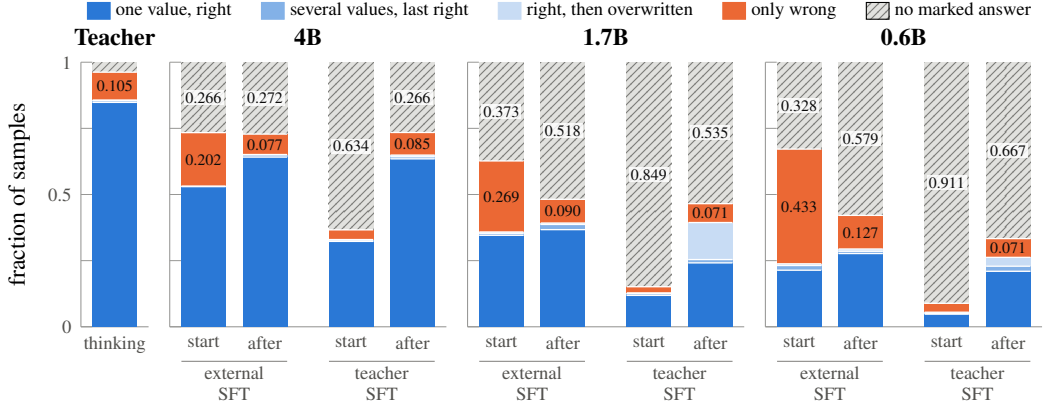

%% file: tables/lean_unfinished.tex
\begin{table}[h]
\centering
\footnotesize
\caption{\textbf{The unfinished samples per benchmark.} Samples that reach the evaluation limit. \emph{Correct value
present}: the correct value occurs anywhere in the text, on problems whose integer answer has at least three digits and
does not occur in the statement (\emph{eligible}); \emph{chance}: the answer of another such problem occurs.
\emph{Repeated ending}: some line occurs at least five times in the last 2{,}000 characters.}
\label{tab:app-unfinished-bench}
\scriptsize
\setlength{\tabcolsep}{3pt}
\begin{tabular}{@{}llcccccc@{}}
\toprule
Model & Benchmark & unfinished & \makecell{marked\\an answer} & eligible & \makecell{correct value\\present} & chance & \makecell{repeated\\ending} \\
\midrule
Teacher (thinking) & AIME 2026 & 154 & 0.143 & 143 & 0.147 & 0.070 & 0.058 \\
 & AMC23 & 36 & 0.250 & 1 & 0.000 & 0.000 & 0.111 \\
 & GSM8K-200 & 58 & 0.207 & 30 & 0.367 & 0.033 & 0.103 \\
\addlinespace[2pt]
4B, distilled & AIME 2026 & 798 & 0.034 & 666 & 0.141 & 0.051 & 0.060 \\
 & AMC23 & 460 & 0.039 & 61 & 0.197 & 0.000 & 0.041 \\
 & GSM8K-200 & 380 & 0.166 & 108 & 0.648 & 0.065 & 0.184 \\
\addlinespace[2pt]
1.7B, distilled & AIME 2026 & 1,242 & 0.024 & 938 & 0.137 & 0.036 & 0.179 \\
 & AMC23 & 1,090 & 0.049 & 127 & 0.386 & 0.039 & 0.149 \\
 & GSM8K-200 & 1,215 & 0.093 & 437 & 0.618 & 0.059 & 0.256 \\
\addlinespace[2pt]
0.6B, distilled & AIME 2026 & 1,273 & 0.037 & 931 & 0.044 & 0.016 & 0.306 \\
 & AMC23 & 1,355 & 0.088 & 165 & 0.145 & 0.012 & 0.450 \\
 & GSM8K-200 & 1,863 & 0.170 & 626 & 0.557 & 0.075 & 0.129 \\
\bottomrule
\end{tabular}
\end{table}

%% file: tables/lean_forced.tex
\begin{table}[h]
\centering
\footnotesize
\caption{\textbf{Forcing a stop recovers answers mainly where the model could already solve.} Every unfinished sample
of the officially post-trained 0.6B on 128 problems (4 samples each) is given back to the model that wrote it with
\think{} appended, and the model continues for up to 1{,}024 tokens. \emph{Recovered}: the continuation's answer is
correct. Problems are grouped by how many of the starting model's four samples solved them.}
\label{tab:app-forced}
\setlength{\tabcolsep}{4pt}
\begin{tabular}{@{}lcccccccc@{}}
\toprule
 & \multicolumn{2}{c}{all} & \multicolumn{2}{c}{solved 0/4} & \multicolumn{2}{c}{solved 1--2/4} & \multicolumn{2}{c}{solved 3--4/4} \\
\cmidrule(lr){2-3}\cmidrule(lr){4-5}\cmidrule(lr){6-7}\cmidrule(lr){8-9}
Model & $n$ & recovered & $n$ & recovered & $n$ & recovered & $n$ & recovered \\
\midrule
before training & 104 & 0.067 & 84 & 0.036 & 19 & 0.158 & 1 & 1.000 \\
after 20 thinking-mode steps & 288 & 0.101 & 206 & 0.049 & 60 & 0.183 & 22 & 0.364 \\
\bottomrule
\end{tabular}
\end{table}

%% file: figures/fig3_anatomy.tex
\begin{figure}[t]
\centering
\pgfdeclareplotmark{paperstar}{\pgfpathmoveto{\pgfqpointpolar{90}{1.00\pgfplotmarksize}}\pgfpathlineto{\pgfqpointpolar{126}{0.42\pgfplotmarksize}}\pgfpathlineto{\pgfqpointpolar{162}{1.00\pgfplotmarksize}}\pgfpathlineto{\pgfqpointpolar{198}{0.42\pgfplotmarksize}}\pgfpathlineto{\pgfqpointpolar{234}{1.00\pgfplotmarksize}}\pgfpathlineto{\pgfqpointpolar{270}{0.42\pgfplotmarksize}}\pgfpathlineto{\pgfqpointpolar{306}{1.00\pgfplotmarksize}}\pgfpathlineto{\pgfqpointpolar{342}{0.42\pgfplotmarksize}}\pgfpathlineto{\pgfqpointpolar{378}{1.00\pgfplotmarksize}}\pgfpathlineto{\pgfqpointpolar{414}{0.42\pgfplotmarksize}}\pgfpathclose\pgfusepathqfillstroke}
\begin{minipage}[c]{0.5\textwidth}
\centering
\begin{tikzpicture}
\begin{axis}[paperstyle, scale only axis, width=5.7cm, height=3.5cm,
  xmin=0.4, xmax=1.0, ymin=0, ymax=0.48, clip=false,
  xtick={0.4,0.5,0.6,0.7,0.8,0.9,1}, xticklabels={0.4,,0.6,,0.8,,1},
  ytick={0,0.1,0.2,0.3,0.4}, axis x line*=bottom, axis y line*=left,
  xlabel={first marked answer at (fraction of output)}, ylabel={overwrite rate},
  xlabel style={yshift=2pt}, ylabel style={yshift=-2pt},
  legend style={at={(0.44,-0.27)}, anchor=north, legend columns=2,
    /tikz/every even column/.append style={column sep=.9em}}, legend cell align=left]
\addlegendimage{only marks, mark=*, mark size=2.1pt, cMuted}
\addlegendentry{non-thinking route}
\addlegendimage{only marks, mark=paperstar, mark size=3.3pt, cTeacher}
\addlegendentry{thinking teacher}
\addlegendimage{only marks, mark=square*, mark size=1.9pt, cMuted}
\addlegendentry{thinking, external-SFT start}
\addlegendimage{only marks, mark=paperstar, mark size=3.3pt, cTeacher, mark options={fill=white}}
\addlegendentry{non-thinking teacher}
\addlegendimage{only marks, mark=triangle*, mark size=3.0pt, cMuted}
\addlegendentry{thinking, teacher-SFT start}
\addlegendimage{empty legend}
\addlegendentry{\textcolor{cFourB}{\rule{1.1ex}{1.1ex}}\,4B\enspace \textcolor{cOneSevenB}{\rule{1.1ex}{1.1ex}}\,1.7B\enspace \textcolor{cZeroSixB}{\rule{1.1ex}{1.1ex}}\,0.6B}
\addplot[only marks, mark=*, mark size=2.1pt, cFourB, mark options={fill=cFourB, draw=white, line width=.35pt}, forget plot] coordinates {(0.7270,0.0030)};
\addplot[only marks, mark=*, mark size=2.1pt, cOneSevenB, mark options={fill=cOneSevenB, draw=white, line width=.35pt}, forget plot] coordinates {(0.9003,0.0089)};
\addplot[only marks, mark=*, mark size=2.1pt, cZeroSixB, mark options={fill=cZeroSixB, draw=white, line width=.35pt}, forget plot] coordinates {(0.8970,0.0153)};
\addplot[only marks, mark=square*, mark size=1.9pt, cFourB, mark options={fill=cFourB, draw=white, line width=.35pt}, forget plot] coordinates {(0.8753,0.0026)};
\addplot[only marks, mark=square*, mark size=1.9pt, cOneSevenB, mark options={fill=cOneSevenB, draw=white, line width=.35pt}, forget plot] coordinates {(0.8877,0.0315)};
\addplot[only marks, mark=square*, mark size=1.9pt, cZeroSixB, mark options={fill=cZeroSixB, draw=white, line width=.35pt}, forget plot] coordinates {(0.8640,0.1574)};
\addplot[only marks, mark=triangle*, mark size=3.0pt, cFourB, mark options={fill=cFourB, draw=white, line width=.35pt}, forget plot] coordinates {(0.8720,0.0097)};
\addplot[only marks, mark=triangle*, mark size=3.0pt, cOneSevenB, mark options={fill=cOneSevenB, draw=white, line width=.35pt}, forget plot] coordinates {(0.4410,0.4369)};
\addplot[only marks, mark=triangle*, mark size=3.0pt, cZeroSixB, mark options={fill=cZeroSixB, draw=white, line width=.35pt}, forget plot] coordinates {(0.4980,0.1803)};
\addplot[only marks, mark=paperstar, mark size=3.3pt, cTeacher, mark options={fill=white, line width=.4pt}, forget plot] coordinates {(0.9500,0.0012)};
\addplot[only marks, mark=paperstar, mark size=3.3pt, cTeacher, mark options={fill=cTeacher, line width=.4pt}, forget plot] coordinates {(0.8523,0.0035)};
\node[anchor=west, font=\scriptsize, text=cMuted, inner sep=1pt, xshift=4pt] at (axis cs:0.4410,0.4369) {1.7B, teacher-SFT start};
\node[anchor=west, font=\scriptsize, text=cMuted, inner sep=1pt, xshift=4pt] at (axis cs:0.4980,0.1803) {0.6B, teacher-SFT start};
\draw[cMuted, line width=.3pt] (axis cs:0.8443,0.0135) -- (axis cs:0.7773,0.0885) node[anchor=east, font=\scriptsize, text=cMuted, inner sep=1pt] {thinking teacher};
\end{axis}
\end{tikzpicture}
\end{minipage}\hfill
\begin{minipage}[c]{0.46\textwidth}
\caption{\textbf{The two small students distilled from the teacher-SFT start answer early, then
overwrite the answer.} One point per evaluated model, three-benchmark mean. Overwrite rate: how often the last
marked answer is wrong although a correct one was marked. Every other student answers late; from the external-SFT
start, overwriting still rises as size falls.}
\label{fig:anatomy}
\end{minipage}
\end{figure}
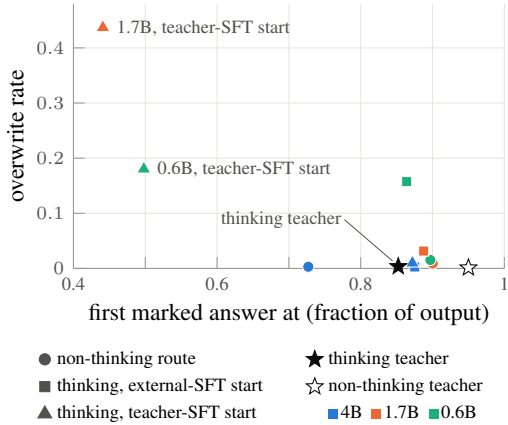

%% file: appendix/F_notes.tex
\section{Data notes}
\label{app:notes}

\paragraph{Data notes.}
856 of the 13,597 training problems (6.3\%) have statements cut short at a blank line by an upstream extraction step;
every run in this paper except the two easy-to-hard curriculum runs trains on the full set. GSM8K was not among the benchmarks we decontaminated against. AMC23 is
used both for validation during training and for evaluation, but every evaluated checkpoint is taken at a fixed step,
so no checkpoint was chosen on it. The two runs of one configuration differ only by being run twice over the same data
order.

%% file: main.bbl
\begin{thebibliography}{32}
\providecommand{\natexlab}[1]{#1}
\providecommand{\url}[1]{\texttt{#1}}
\expandafter\ifx\csname urlstyle\endcsname\relax
  \providecommand{\doi}[1]{doi: #1}\else
  \providecommand{\doi}{doi: \begingroup \urlstyle{rm}\Url}\fi

\bibitem[Agarwal et~al.(2024)Agarwal, Vieillard, Zhou, Stanczyk, Ramos, Geist,
  and Bachem]{agarwal2024gkd}
Rishabh Agarwal, Nino Vieillard, Yongchao Zhou, Piotr Stanczyk, Sabela Ramos,
  Matthieu Geist, and Olivier Bachem.
\newblock On-policy distillation of language models: Learning from
  self-generated mistakes.
\newblock In \emph{International Conference on Learning Representations
  (ICLR)}, 2024.

\bibitem[Chen et~al.(2025)Chen, Xu, Liang, He, Pang, Yu, Song, Liu, Zhou,
  Zhang, Wang, Tu, Mi, and Yu]{chen2024overthinking}
Xingyu Chen, Jiahao Xu, Tian Liang, Zhiwei He, Jianhui Pang, Dian Yu, Linfeng
  Song, Qiuzhi Liu, Mengfei Zhou, Zhuosheng Zhang, Rui Wang, Zhaopeng Tu,
  Haitao Mi, and Dong Yu.
\newblock Do {NOT} think that much for 2+3=? {On} the overthinking of long
  reasoning models.
\newblock In \emph{Proceedings of the 42nd International Conference on Machine
  Learning (ICML)}, volume 267 of \emph{Proceedings of Machine Learning
  Research}, pp.\  9487--9499, 2025.

\bibitem[Davidov et~al.(2026)Davidov, Cohen, Kalinsky, Fairstein, Kushilevitz,
  Yazdi, and Rebeschini]{davidov2026quit}
Hen Davidov, Nachshon Cohen, Oren Kalinsky, Yaron Fairstein, Guy Kushilevitz,
  Ram Yazdi, and Patrick Rebeschini.
\newblock Knowing when to quit: A principled framework for dynamic abstention
  in {LLM} reasoning.
\newblock In \emph{Proceedings of the 43rd International Conference on Machine
  Learning (ICML)}, volume 306 of \emph{Proceedings of Machine Learning
  Research}, 2026.

\bibitem[de~Haan et~al.(2019)de~Haan, Jayaraman, and Levine]{dehaan2019causal}
Pim de~Haan, Dinesh Jayaraman, and Sergey Levine.
\newblock Causal confusion in imitation learning.
\newblock In \emph{Advances in Neural Information Processing Systems
  (NeurIPS)}, volume~32, 2019.

\bibitem[Fu et~al.(2026)Fu, Huang, Jiang, Liu, Jiang, Zhu, and
  Zhao]{fu2026revisiting}
Yuqian Fu, Haohuan Huang, Kaiwen Jiang, Jiacai Liu, Zhuo Jiang, Yuanheng Zhu,
  and Dongbin Zhao.
\newblock Revisiting on-policy distillation: Empirical failure modes and simple
  fixes.
\newblock In \emph{Conference on Language Modeling (COLM)}, 2026.

\bibitem[Ge et~al.(2026)Ge, Zhang, Huang, Zhu, Yuan, Gu, Wu, Huang, Zhang, Han,
  Zhou, and Yao]{ge2026tts}
Xinmu Ge, Zizhuo Zhang, Yu~Huang, Jianing Zhu, Lin Yuan, Wanli Gu, Weichang Wu,
  Weiran Huang, Xiaolu Zhang, Bo~Han, Jun Zhou, and Jiangchao Yao.
\newblock Towards understanding on-policy distillation through the lens of
  test-time scaling.
\newblock \emph{arXiv preprint arXiv:2608.11829}, 2026.

\bibitem[Gu et~al.(2024)Gu, Dong, Wei, and Huang]{gu2024minillm}
Yuxian Gu, Li~Dong, Furu Wei, and Minlie Huang.
\newblock {MiniLLM}: Knowledge distillation of large language models.
\newblock In \emph{International Conference on Learning Representations
  (ICLR)}, 2024.

\bibitem[Guo et~al.(2025)Guo, Yang, Zhang, Song, Wang, Zhu, Xu, Zhang, Ma, Bi,
  et~al.]{guo2025deepseekr1}
Daya Guo, Dejian Yang, Haowei Zhang, Junxiao Song, Peiyi Wang, Qihao Zhu,
  Runxin Xu, Ruoyu Zhang, Shirong Ma, Xiao Bi, et~al.
\newblock {DeepSeek-R1} incentivizes reasoning in {LLMs} through reinforcement
  learning.
\newblock \emph{Nature}, 645\penalty0 (8081):\penalty0 633--638, 2025.

\bibitem[He et~al.(2026)He, Lei, Luo, Zhang, Zhang, Li, Tang, Wang, Zhan, Bai,
  Cui, Yu, Li, Ye, Ding, and Cheng]{he2026simpleopd}
Haonan He, Haodi Lei, Yun Luo, Haoran Zhang, Shunkai Zhang, Yizhuo Li, Shengji
  Tang, Zhilin Wang, Runzhe Zhan, Lei Bai, Ganqu Cui, Fangchen Yu, Yafu Li,
  Peng Ye, Ning Ding, and Yu~Cheng.
\newblock {SimpleOPD}: Simple tokenizer-agnostic on-policy distillation for
  long-context reasoning.
\newblock \emph{arXiv preprint arXiv:2608.14277}, 2026.

\bibitem[Hinton et~al.(2015)Hinton, Vinyals, and Dean]{hinton2015distilling}
Geoffrey Hinton, Oriol Vinyals, and Jeff Dean.
\newblock Distilling the knowledge in a neural network.
\newblock \emph{arXiv preprint arXiv:1503.02531}, 2015.
\newblock NIPS 2014 Deep Learning Workshop.

\bibitem[Jin et~al.(2026)Jin, Min, Yang, Wei, Zhou, Kadhe, Baracaldo, and
  Lee]{jin2026eopd}
Woogyeol Jin, Taywon Min, Yongjin Yang, Dennis Wei, Yi~Zhou, Swanand~Ravindra
  Kadhe, Nathalie Baracaldo, and Kimin Lee.
\newblock Entropy-aware on-policy distillation of language models.
\newblock In \emph{International Conference on Machine Learning (ICML)}, 2026.

\bibitem[Kaur et~al.(2026)Kaur, Ri, He, Fowl, and Arora]{kaur2026rethinking}
Simran Kaur, Narutatsu Ri, Yinghui He, Liam Fowl, and Sanjeev Arora.
\newblock Rethinking on-policy self-distillation for thinking models.
\newblock \emph{arXiv preprint arXiv:2607.05184}, 2026.

\bibitem[Kim \& Rush(2016)Kim and Rush]{kim2016sequence}
Yoon Kim and Alexander~M. Rush.
\newblock Sequence-level knowledge distillation.
\newblock In \emph{Proceedings of the 2016 Conference on Empirical Methods in
  Natural Language Processing (EMNLP)}, pp.\  1317--1327, 2016.

\bibitem[Li et~al.(2026)Li, Zuo, He, Zhang, Xiao, Qian, Yu, ang Gao, Yang, Liu,
  and Ding]{li2026rethinking}
Yaxuan Li, Yuxin Zuo, Bingxiang He, Jinqian Zhang, Chaojun Xiao, Cheng Qian,
  Tianyu Yu, Huan ang Gao, Wenkai Yang, Zhiyuan Liu, and Ning Ding.
\newblock Rethinking on-policy distillation of large language models:
  Phenomenology, mechanism, and recipe.
\newblock \emph{arXiv preprint arXiv:2604.13016}, 2026.

\bibitem[Li et~al.(2025)Li, Yue, Xu, Jiang, Niu, Lin, Ramasubramanian, and
  Poovendran]{li2025smallmodels}
Yuetai Li, Xiang Yue, Zhangchen Xu, Fengqing Jiang, Luyao Niu, Bill~Yuchen Lin,
  Bhaskar Ramasubramanian, and Radha Poovendran.
\newblock Small models struggle to learn from strong reasoners.
\newblock In \emph{Findings of the Association for Computational Linguistics:
  ACL 2025}, pp.\  25366--25394, 2025.

\bibitem[Liu et~al.(2026)Liu, He, Tao, Zhan, Gao, Zhang, Duan, Xiong, Sheng,
  Zhang, Guan, and Li]{liu2026beyond}
Shiqi Liu, Zeyu He, Letian Tao, Guojian Zhan, Jiaxin Gao, Feihong Zhang,
  Jingliang Duan, Wei Xiong, Kehua Sheng, Bo~Zhang, Yang Guan, and Shengbo~Eben
  Li.
\newblock Beyond token-local imitation: Reward-compatible temporal credit
  assignment for on-policy distillation.
\newblock \emph{arXiv preprint arXiv:2609.16937}, 2026.

\bibitem[Lu \& {Thinking Machines Lab}(2025)Lu and {Thinking Machines
  Lab}]{tml2025opd}
Kevin Lu and {Thinking Machines Lab}.
\newblock On-policy distillation.
\newblock Thinking Machines Lab: Connectionism,
  \url{https://thinkingmachines.ai/blog/on-policy-distillation/}, October 2025.
\newblock doi:10.64434/tml.20251026.

\bibitem[Ma(2026)]{ma2026outcome}
Guoqing Ma.
\newblock Outcome-confounded local supervision in on-policy distillation.
\newblock \emph{arXiv preprint arXiv:2607.23731}, 2026.

\bibitem[Muennighoff et~al.(2025)Muennighoff, Yang, Shi, Li, Fei-Fei,
  Hajishirzi, Zettlemoyer, Liang, Cand{\`e}s, and Hashimoto]{muennighoff2025s1}
Niklas Muennighoff, Zitong Yang, Weijia Shi, Xiang~Lisa Li, Li~Fei-Fei,
  Hannaneh Hajishirzi, Luke Zettlemoyer, Percy Liang, Emmanuel Cand{\`e}s, and
  Tatsunori Hashimoto.
\newblock s1: Simple test-time scaling.
\newblock In \emph{Proceedings of the 2025 Conference on Empirical Methods in
  Natural Language Processing (EMNLP)}, pp.\  20275--20321, 2025.

\bibitem[Ross et~al.(2011)Ross, Gordon, and Bagnell]{ross2011dagger}
Stephane Ross, Geoffrey~J. Gordon, and J.~Andrew Bagnell.
\newblock A reduction of imitation learning and structured prediction to
  no-regret online learning.
\newblock In \emph{Proceedings of the Fourteenth International Conference on
  Artificial Intelligence and Statistics (AISTATS)}, volume~15 of
  \emph{Proceedings of Machine Learning Research}, pp.\  627--635, 2011.

\bibitem[Swamy et~al.(2022)Swamy, Choudhury, Bagnell, and
  Wu]{swamy2022unobserved}
Gokul Swamy, Sanjiban Choudhury, J.~Andrew Bagnell, and Zhiwei~Steven Wu.
\newblock Sequence model imitation learning with unobserved contexts.
\newblock In \emph{Advances in Neural Information Processing Systems
  (NeurIPS)}, volume~35, pp.\  17665--17676, 2022.

\bibitem[Wang et~al.(2026)Wang, Wang, Chen, Xue, Fang, Yu, and
  Wong]{wang2026demystify}
Rui Wang, Hongru Wang, Yi~Chen, Boyang Xue, Tianqing Fang, Wenhao Yu, and
  Kam-Fai Wong.
\newblock Demystifying on-policy distillation: Roles, pathologies, and
  regulations.
\newblock \emph{arXiv preprint arXiv:2607.13399}, 2026.

\bibitem[Wang et~al.(2023)Wang, Wei, Schuurmans, Le, Chi, Narang, Chowdhery,
  and Zhou]{wang2023selfconsistency}
Xuezhi Wang, Jason Wei, Dale Schuurmans, Quoc Le, Ed~Chi, Sharan Narang,
  Aakanksha Chowdhery, and Denny Zhou.
\newblock Self-consistency improves chain of thought reasoning in language
  models.
\newblock In \emph{International Conference on Learning Representations
  (ICLR)}, 2023.

\bibitem[Wang et~al.(2025)Wang, Liu, Xu, Liang, Chen, He, Song, Yu, Li, Zhang,
  Wang, Tu, Mi, and Yu]{wang2025underthinking}
Yue Wang, Qiuzhi Liu, Jiahao Xu, Tian Liang, Xingyu Chen, Zhiwei He, Linfeng
  Song, Dian Yu, Juntao Li, Zhuosheng Zhang, Rui Wang, Zhaopeng Tu, Haitao Mi,
  and Dong Yu.
\newblock Thoughts are all over the place: On the underthinking of long
  reasoning models.
\newblock In \emph{Advances in Neural Information Processing Systems
  (NeurIPS)}, volume~38, 2025.

\bibitem[Weihs et~al.(2021)Weihs, Jain, Liu, Salvador, Lazebnik, Kembhavi, and
  Schwing]{weihs2021imitationgap}
Luca Weihs, Unnat Jain, Iou-Jen Liu, Jordi Salvador, Svetlana Lazebnik,
  Aniruddha Kembhavi, and Alexander Schwing.
\newblock Bridging the imitation gap by adaptive insubordination.
\newblock In \emph{Advances in Neural Information Processing Systems
  (NeurIPS)}, volume~34, pp.\  19134--19146, 2021.

\bibitem[Xin et~al.(2026)Xin, Zhao, Sun, Li, Shen, and Xiong]{xin2026kat}
Haoran Xin, Anhao Zhao, Ying Sun, Jin Li, Xiaoyu Shen, and Hui Xiong.
\newblock Escaping the {KL} agreement trap in on-policy distillation.
\newblock \emph{arXiv preprint arXiv:2606.09471}, 2026.

\bibitem[Yang et~al.(2025)Yang, Li, Yang, Zhang, Hui, Zheng, Yu, Gao, Huang,
  Lv, Zheng, Liu, Zhou, Huang, Hu, Ge, Wei, Lin, Tang, Yang, Tu, Zhang, Yang,
  Yang, Zhou, Zhou, Lin, Dang, Bao, Yang, Yu, Deng, Li, Xue, Li, Zhang, Wang,
  Zhu, Men, Gao, Liu, Luo, Li, Tang, Yin, Ren, Wang, Zhang, Ren, Fan, Su,
  Zhang, Zhang, Wan, Liu, Wang, Cui, Zhang, Zhou, and Qiu]{qwen3}
An~Yang, Anfeng Li, Baosong Yang, Beichen Zhang, Binyuan Hui, Bo~Zheng, Bowen
  Yu, Chang Gao, Chengen Huang, Chenxu Lv, Chujie Zheng, Dayiheng Liu, Fan
  Zhou, Fei Huang, Feng Hu, Hao Ge, Haoran Wei, Huan Lin, Jialong Tang, Jian
  Yang, Jianhong Tu, Jianwei Zhang, Jianxin Yang, Jiaxi Yang, Jing Zhou,
  Jingren Zhou, Junyang Lin, Kai Dang, Keqin Bao, Kexin Yang, Le~Yu, Lianghao
  Deng, Mei Li, Mingfeng Xue, Mingze Li, Pei Zhang, Peng Wang, Qin Zhu, Rui
  Men, Ruize Gao, Shixuan Liu, Shuang Luo, Tianhao Li, Tianyi Tang, Wenbiao
  Yin, Xingzhang Ren, Xinyu Wang, Xinyu Zhang, Xuancheng Ren, Yang Fan, Yang
  Su, Yichang Zhang, Yinger Zhang, Yu~Wan, Yuqiong Liu, Zekun Wang, Zeyu Cui,
  Zhenru Zhang, Zhipeng Zhou, and Zihan Qiu.
\newblock {Qwen3} technical report.
\newblock \emph{arXiv preprint arXiv:2505.09388}, 2025.

\bibitem[Yang et~al.(2026)Yang, Si, Duan, Zhu, Zhu, Li, Chen, Lin, and
  Wang]{yang2026deer}
Chenxu Yang, Qingyi Si, Yongjie Duan, Zheliang Zhu, Chenyu Zhu, Qiaowei Li,
  Minghui Chen, Zheng Lin, and Weiping Wang.
\newblock Dynamic early exit in reasoning models.
\newblock In \emph{International Conference on Learning Representations
  (ICLR)}, 2026.

\bibitem[Yu et~al.(2025)Yu, Zhang, Zhu, Yuan, Zuo, Yue, Dai, Fan, Liu, Liu,
  Liu, Liu, Lin, Lin, Ma, Sheng, Tong, Zhang, Zhang, Zhang, Zhang, Zhu, Zhu,
  Chen, Chen, Wang, Yu, Song, Wei, Zhou, Liu, Ma, Zhang, Yan, Wu, and
  Wang]{yu2025dapo}
Qiying Yu, Zheng Zhang, Ruofei Zhu, Yufeng Yuan, Xiaochen Zuo, Yu~Yue, Weinan
  Dai, Tiantian Fan, Gaohong Liu, Juncai Liu, Lingjun Liu, Xin Liu, Haibin Lin,
  Zhiqi Lin, Bole Ma, Guangming Sheng, Yuxuan Tong, Chi Zhang, Mofan Zhang,
  Ru~Zhang, Wang Zhang, Hang Zhu, Jinhua Zhu, Jiaze Chen, Jiangjie Chen,
  Chengyi Wang, Hongli Yu, Yuxuan Song, Xiangpeng Wei, Hao Zhou, Jingjing Liu,
  Wei-Ying Ma, Ya-Qin Zhang, Lin Yan, Yonghui Wu, and Mingxuan Wang.
\newblock {DAPO}: An open-source {LLM} reinforcement learning system at scale.
\newblock In \emph{Advances in Neural Information Processing Systems
  (NeurIPS)}, volume~38, pp.\  113222--113244, 2025.

\bibitem[Yue et~al.(2025)Yue, Chen, Lu, Zhao, Wang, Yue, Song, and
  Huang]{yue2025rlvr}
Yang Yue, Zhiqi Chen, Rui Lu, Andrew Zhao, Zhaokai Wang, Yang Yue, Shiji Song,
  and Gao Huang.
\newblock Does reinforcement learning really incentivize reasoning capacity in
  {LLMs} beyond the base model?
\newblock In \emph{Advances in Neural Information Processing Systems}, 2025.

\bibitem[Zhang et~al.(2025)Zhang, Chen, Pan, Zhao, Panda, Li, and
  He]{zhang2025probing}
Anqi Zhang, Yulin Chen, Jane Pan, Chen Zhao, Aurojit Panda, Jinyang Li, and
  He~He.
\newblock Reasoning models know when they're right: Probing hidden states for
  self-verification.
\newblock In \emph{Conference on Language Modeling (COLM)}, 2025.

\bibitem[Zhou et~al.(2026)Zhou, Li, Tang, Wu, and Terzopoulos]{ziheng2026esr}
Ziheng Zhou, Jiaqi Li, Huacong Tang, Ying~Nian Wu, and Demetri Terzopoulos.
\newblock Less is more: Early stopping rollout for on-policy distillation.
\newblock \emph{arXiv preprint arXiv:2605.27028}, 2026.

\end{thebibliography}
